\documentclass[journal]{IEEEtran}
\usepackage[numbers,sort&compress]{natbib}% our
\usepackage{footnote}
\usepackage{url}

\usepackage{soul}% Added by us
\soulregister\cite7
\soulregister\citep7
\soulregister\citet7
\soulregister\ref7
\soulregister\pageref7

\ifCLASSINFOpdf
   \usepackage[pdftex]{graphicx}
   \graphicspath{{myfigure}}
   \DeclareGraphicsExtensions{.pdf,.jpeg,.png,.jpg}
\else
   \usepackage[dvips]{graphicx}
   \graphicspath{{../eps/}}
   \DeclareGraphicsExtensions{.eps}
\fi
\usepackage{amsmath}
\usepackage{amsfonts}
\usepackage{algpseudocode}%Added by us
\usepackage{algorithm}%Added by us
\ifCLASSOPTIONcompsoc
  \usepackage[caption=false,font=normalsize,labelfont=sf,textfont=sf]{subfig}
\else
  \usepackage[caption=false,font=footnotesize]{subfig}
\fi
\usepackage{fixltx2e}
\usepackage{stfloats}
\fnbelowfloat
\begin{document}
	
%
% paper title
% Titles are generally capitalized except for words such as a, an, and, as,
% at, but, by, for, in, nor, of, on, or, the, to and up, which are usually
% not capitalized unless they are the first or last word of the title.
% Linebreaks \\ can be used within to get better formatting as desired.
% Do not put math or special symbols in the title.
%\title{HSI Restoration Via $L_{1-2}$ spatial-spectral Total Variation Regularized Local Low-rank Tensor Recovery}
\title{Enhanced nonconvex low-rank approximation of tensor multi-modes for tensor completion}
%
% author names and IEEE memberships
% note positions of commas and nonbreaking spaces ( ~ ) LaTeX will not break
% a structure at a ~ so this keeps an author's name from being broken across
% two lines.
% use \thanks{} to gain access to the first footnote area
% a separate \thanks must be used for each paragraph as LaTeX2e's \thanks
% was not built to handle multiple paragraphs
%

\author{Haijin~Zeng,
		Yongyong Chen,
        Xiaozhen~Xie,
        and~Jifeng~Ning% <-this % stops a space
%\thanks{This work was supported by the Fundamental Research Funds for the Central Universities under Grant No. 2452019073 and
%the National Natural Science Foundation of China under Grant No. 61876153.(Corresponding author: Xiaozhen Xie.)}
\thanks{Haijin~Zeng,  and Xiaozhen~Xie are with the College of Science, Northwest A\&F University, Yangling 712100, China (e-mail: zeng\_navy@163.com; xiexzh@nwafu.edu.cn).}
\thanks{Yongyong Chen is with the Department of Computer and
	Information Science, University of Macau, Macau 999078, China (e-mail:
	yongyongchen.cn@gmail.com).}
\thanks{Jifeng~Ning is with the College of Information Engineering, Northwest A\&F University, Yangling 712100, China (e-mail:  njf@nwsuaf.edu.cn).}% <-this % stops a space
\thanks{Manuscript received April 19, 2005; revised August 26, 2015.}}

\maketitle

% As a general rule, do not put math, special symbols or citations
% in the abstract or keywords.
\begin{abstract}
Higher-order low-rank tensor arises in many data processing applications and has attracted great interests.
Inspired by low-rank approximation theory, researchers have proposed a series of effective tensor completion methods.
However, most of these methods directly consider the global low-rankness of underlying tensors, which is not sufficient for a low sampling rate;
in addition, the single nuclear norm or its relaxation is usually adopted to approximate the rank function, which would lead to suboptimal solution deviated from the original one.
To alleviate the above problems,
in this paper,
we propose a novel low-rank approximation of tensor multi-modes (LRATM),
in which a double nonconvex $L_{\gamma}$ norm is designed to represent the underlying joint-manifold drawn from the modal factorization factors of the underlying tensor.
A block successive upper-bound minimization method-based algorithm is designed to efficiently solve the proposed model,
and it can be demonstrated that our numerical scheme converges to the coordinatewise minimizers.
Numerical results on three types of public multi-dimensional datasets have tested and
shown that our algorithm can recover a variety of low-rank tensors with significantly fewer samples than the compared methods.
\end{abstract}

% Note that keywords are not normally used for peerreview papers.

% For peer review papers, you can put extra information on the cover
% page as needed:
% \ifCLASSOPTIONpeerreview
% \begin{center} \bfseries EDICS Category: 3-BBND \end{center}
% \fi
%
% For peerreview papers, this IEEEtran command inserts a page break and
% creates the second title. It will be ignored for other modes.
\IEEEpeerreviewmaketitle

\section{Introduction}
% The very first letter is a 2 line initial drop letter followed
% by the rest of the first word in caps.
%
% form to use if the first word consists of a single letter:
% \IEEEPARstart{A}{demo} file is ....
%
% form to use if you need the single drop letter followed by
% normal text (unknown if ever used by the IEEE):
% \IEEEPARstart{A}{}demo file is ....
%
% Some journals put the first two words in caps:
% \IEEEPARstart{T}{his demo} file is ....
%
% Here we have the typical use of a "T" for an initial drop letter
% and "HIS" in caps to complete the first word.
%\IEEEPARstart{T}{his}
% You must have at least 2 lines in the paragraph with the drop letter
% (should never be an issue)

%begin test

Tensors, as a generalization of vectors and matrices, arise in many data processing applications and have attracted great interests.
For instance, video inpainting \cite{tensor_video},
magnetic resonance imaging (MRI) data recovery \cite{tensor_MRI, MRI_tensor},
3D image reconstruction \cite{tensor_3dimage, tensor_3DReconstruction},
high-order web link analysis [16],
hyperspectral image (HSI) or multispectral image recovery \cite{tensor_HSI, chen2018tensor},
personalized web search \cite{tensor_web}, and seismic data reconstruction \cite{tensor_seismic_data}.

%low-rank tensor decomposition based methods are considered state-of-the-art methods with promising performance.
%The key problem of these methods is to decompose the target tensor into a combination of several sub-tensors and matrixes which usually describes the intrinsic subspace structures.
%To this end, various low-rank tensor decomposition methods are proposed.
%The most representative methods are weighted low-rank tensor decomposition method \cite{LRTD_1},
%Bayes-based framework \cite{bayes_1, bayes_2} and multi-linear graph embedding \cite{muti_linear_1, Tensor_factor_prior}.
%These methods can effectively recover tensors, however they are usually sensitive to a given rank which is usually estimated based on the raw data. However there may exist
%noise and outlier in the raw data, e.g. video variations caused by illumination and color.
%These factors would reduce the ability to represent the correlation among data.

%Therefore, they have been universally utilized in tensor completion problems.
%Usually, they can be solved by replacing the rank function with its convex or nonconvex relaxations in the minimization problem.

Tensor completion tries to recover a low-rank tensor from its partially observed entries.
A large number of tensor completion methods have been proposed and successfully used in many applications.
Among them, the tensor rank minimization based methods are considered as state-of-the-art methods with promising performance,
and their robustness to noisy and missing data has also been proven \cite{lu2019TRPCA}.
However, due to the nonunique definitions of the tensor rank,
it is extremely hard to directly solve the tensor rank minimization problem.
To overcome this issue, many researchers have been devoted to defining the tensor rank based on the different decomposition methods,
such as the matrix factorization \cite{LMaFit}, CANDECOMP/PARAFAC (CP) decomposition \cite{CP}, Tucker decomposition \cite{tucker_1, ZENG_HSI_tensor},
tensor singular value decomposition (t-SVD) \cite{t_SVD}, tensor train \cite{tensor_train_TingzhuHuang} and tensor ring \cite{tensor_ring_YipengLiu}.

The commonly used definitions of tensor rank are CP-rank, Tucker-rank, multi-rank and tubal-rank based on t-SVD.
However, it is NP-hard to solve the minimization problem of CP-rank which has no relaxation and certain limitations in applications.
Although the Tucker-rank, relying on matrix ranks, is relatively simple,
it is also NP-hard to directly minimizing the Tucker-rank problem.
To tackle this difficulty, the sum of nuclear norm (SNN) \cite{HaLRTC} is introduced as a convex relaxation of the Tucker-rank.
Specifically, SNN is defined by the sum of nuclear norms of the unfolding matrices along all dimensions in tensors.
Due to the similar approximation to matrix case and convenient calculation algorithm, SNN is widely used in the tensor completion task \cite{tensor_Qrank}.
Besides, the tensor multi-rank and tubal-rank induced from t-SVD are computable.
As a tightest convex surrogate for tensor multi-rank, the tensor nuclear norm (TNN) \cite{t_SVD} is defined as
the sum of the matrix nuclear norms of each frontal slice in the Fourier transformed tensor.
TNN has shown its effectiveness to keep the intrinsic structure of tensors and attracted extensive attention for tensor completion problems in recent years \cite{lu2019TRPCA}.

Due to the convexity of matrix nuclear norms, SNN and TNN have limitation in the accuracy of approximation to the tensor rank function.
Recently, a number of studies \cite{list_nonconvex,Original_BFMN}, both practically and theoretically,
have shown that the nonconvex approximation of rank function can provide better estimation accuracy and variable selection consistency than the nuclear norm.
For example,  a partial sum of the tensor nuclear norm (PSTNN) is proposed as a nonconvex surrogate of the tensor rank by Jiang et al. \cite{PSTNN};
Xue et al. \cite{xue2019nonconvex} unfold the underlying tensor along its all modes and use a nonconvex logarithmic surrogate function to refine the rank approximation.
Actually, except for the logarithmic function used by Xue et al.,
a series of nonconvex surrogate are proposed for approximating to the rank function better,
such as the minimax concave function \cite{Folded_concave}, log-sum function \cite{logsum_penalty}, log-determinant function \cite{ji_nonlogDet}, $L_p$ norm for $p\in (0, 1)$ \cite{lysaker_noise_2004_10_27,burger_nonlinear_2006_10_28}, $L_{1/2}$ norm \cite{lou_L1L2} and $\gamma$ norm \cite{Non_LRMA}.
In addition, TNN and PSTNN involve the singular value decompositions (SVDs) of many matrices, which are time-consuming.
To cope with this issue, Xu et al. \cite{Tmac} propose a parallel matrix factorization low-rank tensor completion model (TMac), which obtain promising results with less running time than TNN and PSTNN.
Further, combined with the total variation (TV) regularization, Ji et al. \cite{MFTV} propose TV regularized low-rank matrix factorization method (MF-TV) for low-rank tensor completion problems.

\begin{figure*}[!t]	
	\centering
	\subfloat[Original image]{\includegraphics[width=0.23\linewidth]{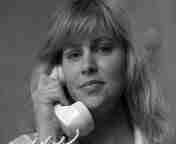}}%%
	\hfil	
	\subfloat[Our model]{\includegraphics[width=0.23\linewidth]{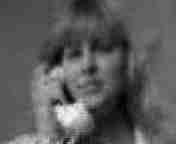}}%
	\hfil										
	\subfloat[TMac]{\includegraphics[width=0.23\linewidth]{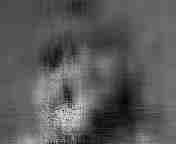}}%
	\hfil
	\subfloat[TNN]{\includegraphics[width=0.23\linewidth]{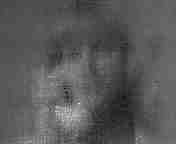}}%	
	\caption{The completed results of Suzie with 95\% missing entries by different methods. From (a) to (d), the original image, the result by our model, Tmac, and TNN, respectively.}
	\label{TNN_Tamc_our-model_figure_video_sr0.05}
\end{figure*}

Although the above-mentioned low-rank tensor completion methods show great success in dealing with various issues, three major open questions have yet to be addressed.
Firstly, in the tensor rank approximations based on tensor decomposition or matrix decomposition,
the low-rank priors of underlying tensor are only explored by the convex or nonconvex relaxations of original tensor rank function,
while the low-rank priors of factors obtained by the decomposition are not investigated further.
Secondly, TNN or PSTNN based methods \cite{TNN, PSTNN} need to compute lots of SVDs, which become very slow or even not applicable for large-scale problems \cite{Tmac}.
Thirdly, the aforementioned methods adopt single surrogate of tensor rank function,
which would cause suboptimal solution of the low-rank tensor completion problems \cite{T_Sp} and can not fully explore the low-rank priors in all modes,
especially when the tensor data is heavily contaminated or the sampling rate is very low.
One can see an example in Fig. \ref{TNN_Tamc_our-model_figure_video_sr0.05}.

%In recent works, researchers usually adopt nonconvex penalties instead of the traditional nuclear norm for low-rank based problems \cite{lou_L1_L2, Sp_Lp, Original_BFMN, DBNN_Grassmann}.
%Nonconvex penalty has decomposable approach and it could construct a more accurate low rank matrix than the traditional nuclear norm \cite{Sp_performance}.

In this paper, motivated and convinced by the much better performance of models that utilize the low-ranknesses in all mode in tensors \cite{HaLRTC, Tmac},
instead of using the single surrogate of tensor rank function to represent the low-rank prior in underlying tensor directly,
we first apply parallel matrix factorization to all modes of underlying tensor.
Further, the novel double $L_{\gamma}$ norm, a kind of nonconvex penalty, is designed to represent the underlying joint-manifold drawn from the mode factorization factors.
By exploiting this auxiliary information,
our method leverages low-rank decomposition and low-rank approximation, which help to accurately estimate the mode factors and missing entries.
An block successive upper-bound minimization method-based algorithm is designed to efficiently solve the proposed model,
and it can be demonstrated that our numerical scheme converge to the coordinatewise minimizers.
The proposed model has been evaluated on three types of public tensor datasets,
which show that our algorithm can recover a variety of low-rank tensors with significantly fewer samples than the compared methods.

The rest of this paper is organized as follows.
Section \ref{notation} introduces some notations about tensors and the operations.
Section \ref{related works} reviews the related works.
In Section \ref{the proposed model}, the proposed model is presented and its optimization is deduced in detail.
In Section \ref{Numerical experiments}, the proposed model is evaluated on several public tensor datasets.
Section \ref{conclusion} gives the conclusions.

\section{Preliminary}
\label{notation}

%Before introducing our model and algorithm, we review some notations, tensor operations, regularizers with physical meaning and operators.

\subsection{Notations}

In this paper, following \cite{Tmac}, vector, matrix and tensor are denoted as bold lower-case letter $\mathbf{x}$, bold upper-case letter $\mathbf{X}$ and caligraphic letter $\mathcal{X}$, respectively.
Let $x_{i_{1} \ldots i_{N}}$ represent the $\left(i_{1}, \ldots, i_{N}\right)$-th component of an $N$-way tensor $\mathcal{X}$.
Then, for $\mathcal{X}, \mathcal{Y} \in \mathbb{R}^{I_{1} \times \ldots \times I_{N}},$ their inner product is defined as
\begin{equation}
\label{equation:inner_product}
\langle\mathcal{X}, \mathcal{Y}\rangle=\sum_{i_{1}=1}^{I_{1}} \cdots \sum_{i_{N}=1}^{I_{N}} x_{i_{1} \cdots i_{N}} y_{i_{1} \cdots i_{N}}.
\end{equation}
Based on (\ref{equation:inner_product}), the \textbf{Frobenius norm} of a tensor $\mathcal{X}$ is defined as $\|\mathcal{X}\|_{\text{F}}=\sqrt{\langle\mathcal{X}, \mathcal{X}\rangle}$.
\textbf{Fibers} of tensor $\mathcal{X}$ are defined as a vector obtained by fixing all indices of $\mathcal{X}$ except one, and \textbf{slices}
of $\mathcal{X}$ are defined as a matrix by fixing all indices of $\mathcal{X}$ except two.
The \textbf{mode-$n$ matricization}/\textbf{unfolding} of $\mathcal{X} \in \mathbb{R}^{I_{1} \times \ldots \times I_{N}}$ is denoted
as  a matrix $\mathbf{X}_{(n)} \in \mathbb{R}^{I_{n} \times \Pi_{j \neq n} I_{j}}$ with columns being the mode-$n$ fibers of $\mathcal{X}$ in the lexicographical order.

Furthermore, to clearly represent the matricization process, we define \textbf{unfold}$_{n}(\mathcal{X})=\mathbf{X}_{(n)}$,
and \textbf{fold}$_{n}$ is the inverse of \textbf{unfold}$_{n}$, i.e., \textbf{fold} $_{n}\left(\text { \textbf{unfold} }_{n}(\mathcal{X})\right)=\mathcal{X}$.
The $n$-rank of an $N$-way tensor
$\mathcal{X},$ denoted as $\operatorname{rank}_{n}(\mathcal{X}),$ is the rank of $\mathbf{X}_{(n)},$ and the rank of $\mathcal{X}$ is defined as an array:
\begin{equation}
\operatorname{rank}(\mathcal{X})=\left(\operatorname{rank}\left(\mathbf{X}_{(1)}\right), \cdots, \operatorname{rank}\left(\mathbf{X}_{(N)}\right)\right).
\end{equation}

\subsection{Operators} \label{operators}
The \textbf{Proximal Operator} of is defined as follows:
\begin{equation}
\label{equation:PPA_0}
\operatorname{prox}_{f}(v):=\arg \min _{u} f(u)+\frac{\rho}{2}\|u-v\|^{2},
\end{equation}
where $f(u)$ is convex; $\rho$ is the proximal parameter.
Then, the minimization of $\{f(u)\}$ is equivalent to
\begin{equation}
\arg \min _{u} f(u)+\frac{\rho}{2}\|u-u^k\|^{2}, k=1,2,\cdots,
\end{equation}
where $u^k$ is the last update of $u$.

We define the \textbf{Projection Operator} as follows:
\begin{equation}
\left(\mathcal{P}_{\Omega}(\mathcal{Y})\right)_{i_{1} \cdots i_{N}}=\left\{\begin{array}{ll}
{y_{i_{1}, \cdots, i_N,}} & {\left(i_{1}, \cdots, i_{N}\right) \in \Omega} \\
{0,} & {\text { otherwise }}
\end{array}\right.
\end{equation}
where $\Omega$ is the index set of observed entries.
The function of $\mathcal{P}_{\Omega}$ is to keep the entries in $\Omega$ and zeros out others.

\section{Related Works}
\label{related works}

We first introduce related tensor completing methods based on the tensor rank minimization.
Given a partial observed tensor $\mathcal{F} = \mathcal{P}_{\Omega}(\mathcal{Y}) \in \mathbb{R}^{I_{1} \times I_{2} \times \cdots  \times I_{N}}$,
tensor completion task is to recover a low-rank tensor $\mathcal{Y}$ from $\mathcal{F}$, according to the priors of underlying tensor $\mathcal{Y}$.

In the past decade, TNN induced by t-SVD \cite{t_SVD} has been widely used for 3-order low-rank tensor completion \cite{TNN}.
The TNN based method aims to recover a low-rank tensor by penalizing the nuclear norm of each front slice in the Fourier transformed domain,
\begin{equation}
\label{TNN}
\arg\min_{\mathcal{Y}} \frac{1}{I_{3}} \sum_{i=1}^{I_{3}}\left\|\bar{\mathbf{Y}}^{(i)}\right\|_{*} , s.t.~  \mathcal{P}_{\Omega}(\mathcal{Y})=\mathcal{F},
\end{equation}
where $\bar{\mathbf{Y}}^{(i)} \in \mathbb{C}^{I_{1} \times I_{2}}$ denotes the $i$-th frontal slice of $\bar{\mathcal{Y}}$,
$\bar{\mathcal{Y}}=\operatorname{fft}(\mathcal{Y},[],3)$ denotes the fast Fourier transform of $\mathcal{Y}$ along the third dimension.

\begin{figure*}
	\centering
	\includegraphics[width=1\linewidth]{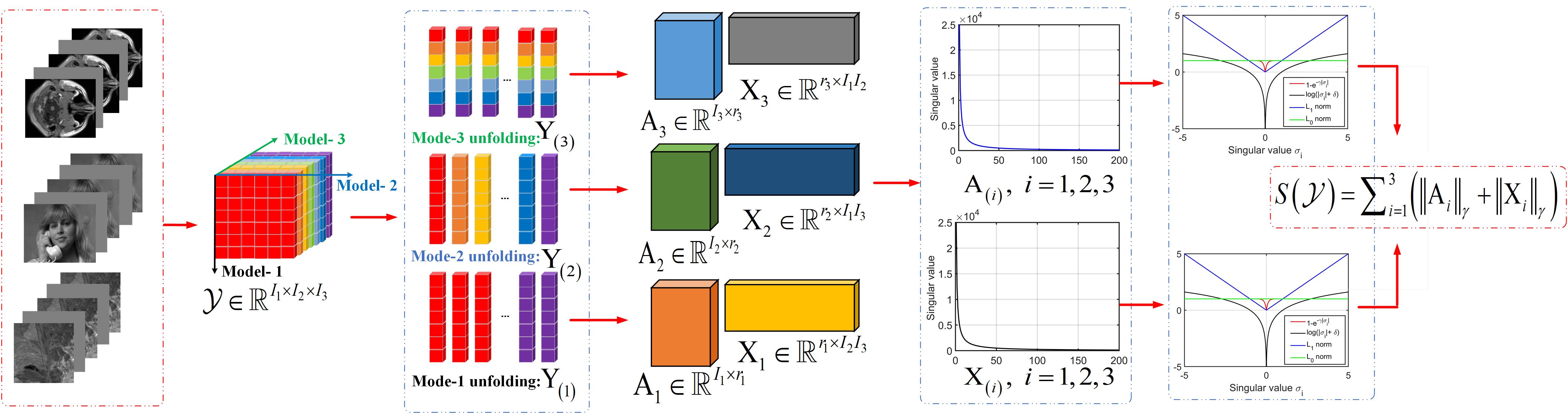}
	\caption{Flowchart of the proposed low-rank tensor approximation: nonconvex tensor $L_{\gamma}$ norm.}
	\label{fig:3-model-lowrank-lx-norm}
\end{figure*}

Then, to alleviate bias phenomenons of the TNN minimization in tensor completion tasks,
Jiang et al. \cite{PSTNN} represent the low-rank prior of underlying tensor by using the PSTNN.
The PSTNN regularized tensor completion model is formulated as follows:
\begin{equation}
\label{PSTNN}
\arg\min_{\mathcal{Y}} \frac{1}{I_{3}} \sum_{i=1}^{I_{3}}\left\|\bar{\mathbf{Y}}^{(i)}\right\|_{p=M}, s.t.~  \mathcal{P}_{\Omega}(\mathcal{Y})=\mathcal{F},
\end{equation}
where $$\|\bar{\mathbf{Y}}^{(i)}\|_{p=M} := \sum_{j=M+1}^{\min (I_1, I_2)} \sigma_{j}(\bar{\mathbf{Y}}^{(i)}),$$
and $\sigma_{j}(\bar{\mathbf{Y}}^{(i)})$ denotes the $j$-th largest singular value of $\bar{\mathbf{Y}}^{(i)}$.

To reduce the burden of calculating SVDs in TNN and PSTNN,
Liu et al. \cite{HaLRTC} unfold the $N$-order tensor into multiple modal matrices along the direction of each mode,
and then use the sum of the nuclear norms of these modal matrices, i.e., SNN, to describe the low-rank structure of the underlying tensor.
With that definition, the completion model is formulated as follows:
\begin{equation}
\label{}
\arg\min_{\mathcal{Y}} \sum_{n=1}^{N} \alpha_i \left\|\mathbf{Y}_{(n)}\right\|_{*}, s.t.,  \mathcal{P}_{\Omega}(\mathcal{Y})=\mathcal{F}.
\end{equation}

Furthermore,
Xu et al. \cite{Tmac} propose a Tmac model by using parallel matrix factorization,
which obtain promising results with less computational complexity than TNN and PSTNN,
\begin{equation}
\label{Tmac}
\begin{aligned}
&\arg\min _{\mathcal{Y}, \mathbf{X}, \mathbf{A}} \sum_{n=1}^{N} \frac{\alpha_{n}}{2}\left\|\mathbf{Y}_{(n)}-\mathbf{A}_{n} \mathbf{X}_{n}\right\|_{F}^{2},\\
&\text { s.t. } \quad \mathcal{P}_{\Omega}(\mathcal{Y})=\mathcal{F},
\end{aligned}
\end{equation}
where $\alpha_n$ are weights and satisfy $\sum_{n=1}^{N}\alpha_n=1$.

Although the above-mentioned low-rank tensor completion methods report success on dealing with a large variety of tasks, there are several open issues which have yet to be addressed.
Firstly, the above approaches
either explore the low-rank prior lying in only one mode of the underlying tensor
or directly represent the low-rank prior of original tensor by using low-rank decomposition.
They do not further explore the prior of the factors (e.g., $\mathbf{A}_n, \mathbf{X}_n$ in (\ref{Tmac})) obtained by low-rank decomposition.
Secondly, TNN based methods \cite{TNN, PSTNN, lu2019TRPCA} need to compute lots of SVDs,
which is time-consuming or even not applicable for large-scale problems \cite{Tmac}.
Thirdly, these methods adopt single surrogate of tensor rank function,
which would cause suboptimal solution of the low-rank tensor completion problems \cite{T_Sp} and can not fully explore the low-rank priors in all modes,
especially when the tensor data is heavily contaminated or the sampling rate is very low.
One can see an example in Fig. \ref{TNN_Tamc_our-model_figure_video_sr0.05}, in which the video "suzie" with 95\% missing entries, it can be seen that our model restores most of the structural information of the video,
while the videos restored by the methods adopt single surrogate contain only the outline of the images.

\section{Double nonconvex $L_{\gamma}$ norm based low-rank approximation for tensor completion }
\label{the proposed model}
In the following, a novel double nonconvex $L_{\gamma}$ norm based low-rank approximation of tensor multi-modes (LRATM) is introduced firstly.
Then, the optimization of the proposed LRATM model is deduced in detail.

\subsection{LRATM Model}
For a tensor $\mathcal{Y}\in \mathbb{R}^{I_{1} \times I_{2} \times \cdots \times I_{N}}$,
to enhance the flexibility for handling different correlations along different modes in the underlying tensor,
while to effectively explore the underlying joint-manifold drawn from the mode factorization factors,
we first formulate a nonconvex novel tensor $L_{\gamma}$ norm,
\begin{equation}
\begin{aligned}
&\left\|\mathcal{Y}\right\|_{\gamma}= \sum_{n=1}^{N} (\tau_n \left\|\mathbf{X}_{n}\right\|_{\gamma}+\lambda_n \left\|\mathbf{A}_{n}\right\|_{\gamma}),
\end{aligned}
\end{equation}
where
$\mathbf{Y}_{(n)}=\mathbf{A}_{n} \mathbf{X}_{n}$, and
%$$\left\|\mathbf{X}\right\|_{\gamma}:=\sum_{t=1}^{\min \{p, q\}}\left(1-e^{-\sigma_{t}(\mathbf{X}) / \gamma}\right)$$
%is a nonconvex approximation of $rank(\mathbf{X})$,
%and $\sigma_{t}(\mathbf{X})$ is the $t$-th biggest singular value of $\mathbf{X} \in \mathbb{R}^{p \times q}$.
$$\left\|\mathbf{X}\right\|_{\gamma}:=\sum_{t=1}^{\min \{p, q\}}\left(1-e^{-\sigma_{t}(\mathbf{X}) / \gamma}\right)$$
is a nonconvex approximation of $rank(\mathbf{X})$,
and $\sigma_{t}(\mathbf{X})$ is the $t$-th biggest singular value of $\mathbf{X} \in \mathbb{R}^{p \times q}$. $\tau_n$ and $\lambda_n$ are non-negative constants that balance the two terms.

Then, our tensor $L_{\gamma}$ norm based low-rank approximation model for low-rank tensor completion,
i.e., the proposed LRATM model is written as
\begin{equation} \label{BMF_LRTC}
\begin{aligned}
\arg\min{\left\|\mathcal{Y}\right\|_{\gamma}}=&\arg\min_{\mathcal{Y},\mathbf{X},\mathbf{A}} \sum_{n=1}^{N} (\tau_n \left\|\mathbf{X}_{n}\right\|_{\gamma}+\lambda_n \left\|\mathbf{A}_{n}\right\|_{\gamma}), \\
&s.t., ~ \mathbf{Y}_{(n)}=\mathbf{A}_{n} \mathbf{X}_{n}, \mathcal{P}_{\Omega}(\mathcal{Y})=\mathcal{F}.
\end{aligned}
\end{equation}

To better understand the proposed LRATM model, we plot the flowchart of the proposed low-rank tensor approximation in Fig. \ref{fig:3-model-lowrank-lx-norm}.
It can be seen that the video, MRI and HSI in the first column essentially can be viewed as 3-order tensors in the second column.
When we unfold the 3-order tensor in three directions,
the low-rank structure of all the $N$ modes can be explored by using parallel matrix decomposition, i.e., $\mathbf{Y}_{(n)}=\mathbf{A}_{n} \mathbf{X}_{n}, n=1,2,\cdots, N$,
which is computationally much cheaper than SVD.
The decomposed factors have practical physical meaning,
$\mathbf{A}_n$ represents a library (each column contains a signature of the $n$-th mode direction),
and $\mathbf{X}_n$ is called an encoding.
For example, in the unmixing problem for HSI \cite{HSI_unmixing},
each column of $\mathbf{A}_3$ contains a spectral signature,
and each row of $\mathbf{X}_3$ contains the fractional abundances of a given endmembers.
This interpretation is also valid for the mode-3 unfolding of video and MRI.
The above parallel matrix decomposition can effectively explore the low-rank structure of underlying tensor,
but the prior information contained in the factor matrices ($\mathbf{A}_{n}, \mathbf{X}_{n}$) is not explored at all.
Therefore, to further enhance the potential capacity of tensor completion models, it is necessary to design new and reasonable formulas to explore the priors in the factor matrices.
Here, we propose the novel nonconvex double $L_{\gamma}$ norm to formulate the underlying joint-manifold drawn from the mode factorization factors $\mathbf{A}_n$ and $\mathbf{X}_n$.
The superiority of $L_{\gamma}$ norm over nuclear norm and other nonconvex penalties is shown in the fifth column of Fig. \ref{fig:3-model-lowrank-lx-norm}.
It is obvious that the red curve of $L_{\gamma}$ norm is closer to the green curve of $L_0$ norm (rank function) than other nonconvex penalties.

\subsection{Optimization Procedure of LRATM}

In this section, the proposed model is solved by using the block successive upper-bound minimization (BSUM) \cite{BSUM} method.
The objective function of the proposed LRATM model(\ref{BMF_LRTC}) can be formulated as follows:
\begin{equation}
\begin{aligned}
f(\mathbf{X}, \mathbf{A}, \mathcal{Y})=\sum_{n=1}^{N}& \left(\frac{\alpha_{n}}{2}\left\|\mathbf{Y}_{(n)}-\mathbf{A}_{n} \mathbf{X}_{n}\right\|_{\text{F}}^{2}+\tau_n \left\|\mathbf{X}_{n}\right\|_{\gamma}\right.\\
&\left.+\lambda_n \left\|\mathbf{A}_{n}\right\|_{\gamma}\right).
\end{aligned}
\end{equation}
According to the proximal operator (\ref{equation:PPA_0}), the update can be written as:
\begin{equation} \label{equation_original_PPA}
p(\mathcal{S}, \mathcal{S}^k)=f\left(\mathcal{S}, \mathcal{S}^k\right)+\frac{\rho}{2}\left\|\mathcal{S}-\mathcal{S}^{k}\right\|_{\text{F}}^{2},
\end{equation}
where $\rho>0$ is a positive constant; $\mathcal{S}=(\mathbf{X}, \mathbf{A}, \mathcal{Y})$ and $\mathcal{S}^{k}=\left(\mathbf{X}^{k}, \mathbf{A}^{k}, \mathcal{Y}^{k}\right)$.
Let
\begin{equation}
\left\{\begin{array}{l}
g_{1}\left(\mathbf{X}, \mathcal{S}_{1}^{k}\right)=f\left(\mathbf{X}, \mathbf{A}^{k}, \mathcal{Y}^{k}\right)+\frac{\rho}{2}\left\|\mathbf{X}-\mathbf{X}^{k}\right\|_{\text{F}}^{2}, \\
g_{2}\left(\mathbf{A}, \mathcal{S}_{2}^{k}\right)=f\left(\mathbf{X}^{k+1}, \mathbf{A}, \mathcal{Y}^{k}\right)+\frac{\rho}{2}\left\|\mathbf{A}-\mathbf{A}^{k}\right\|_{\text{F}}^{2,} \\
g_{3}\left(\mathcal{Y}, \mathcal{S}_{3}^{k}\right)=f\left(\mathbf{X}^{k+1}, \mathbf{A}^{k+1}, \mathcal{Y}\right)+\frac{\rho}{2}\left\|\mathcal{Y}-\mathcal{Y}^{k}\right\|_{\text{F}}^{2},
\end{array}\right.	
\end{equation}
where
\begin{equation}
\left\{\begin{array}{l}
S_{1}^{k}=\left(\mathbf{X}^{k}, \mathbf{A}^{k}, \mathcal{Y}^{k}\right),\\
S_{2}^{k}=\left(\mathbf{X}^{k+1}, \mathbf{A}^{k}, \mathcal{Y}^{k}\right),\\
S_{3}^{k}=\left(\mathbf{X}^{k+1}, \mathbf{A}^{k+1}, \mathcal{Y}^{k}\right).
\end{array}\right.	
\end{equation}
%$$S_{1}^{k}=\left(\mathbf{X}^{k}, \mathbf{A}^{k}, \mathcal{Y}^{k}\right), S_{2}^{k}=\left(\mathbf{X}^{k+1}, \mathbf{A}^{k}, \mathcal{Y}^{k}\right), S_{3}^{k}=\left(\mathbf{X}^{k+1}, \mathbf{A}^{k+1}, \mathcal{Y}^{k}\right) .$$
Then, problem (\ref{equation_original_PPA}) can be rewritten as follows:
\begin{equation} \label{equation:XAY}
\left\{\begin{array}{l}
\displaystyle \mathbf{X}^{k+1}=\arg\min_{\mathbf{X}} g_{1}\left(\mathbf{X}, \mathcal{S}_{1}^{k}\right), \\
\displaystyle \mathbf{A}^{k+1}=\arg\min_{\mathbf{A}} g_{2}\left(\mathbf{A}, \mathcal{S}_{2}^{k}\right), \\
\displaystyle \mathcal{Y}^{k+1}=\arg\min_{\mathcal{P}_{\Omega}(\mathcal{Y})=\mathcal{F}} g_{3}\left(\mathcal{Y}, \mathcal{S}_{3}^{k}\right).
\end{array}\right.
\end{equation}

\subsubsection{Update $\mathbf{X}_n$}

With fixing other variables, the optimization subproblem with respect to $\mathbf{X}_n, n=1,2,\cdots,N$, in (\ref{equation:XAY}) can be written as follows:
\begin{equation}
\label{equation_X}
\begin{aligned}
%\mathbf{X}_n^{k+1}=
\arg\min_{\{\mathbf{X}_n\}_{n=1}^{N}} \sum_{n=1}^{N}& \left(\frac{\alpha_{n}}{2}\left\|\mathbf{Y}_{(n)}-\mathbf{A}_{n} \mathbf{X}_{n}\right\|_{\text{F}}^{2}+\tau_n \left\|\mathbf{X}_{n}\right\|_{\gamma}\right.\\
&\left.+\frac{\rho_n}{2}\left\|\mathbf{X}_n-\mathbf{X}_n^{k}\right\|_{\text{F}}^{2}\right).
\end{aligned}
\end{equation}
To efficiently solve the above optimization,
we first introduce auxiliary variables $\mathbf{X}_{n}=\mathbf{Z}_{n}, n=1,2,\cdots,N$.
Then, by the augmented Lagrangian multiplier (ALM) method,
the optimization subproblem (\ref{equation_X}) can be rewritten as:
\begin{equation}
\label{equation:X_alm}
\begin{aligned}
&\arg\min_{\{\mathbf{X}_n, \{\mathbf{Z}_n\}_{n=1}^{N}} \sum_{n=1}^{N} \left(\frac{\alpha_{n}}{2}\left\|\mathbf{Y}_{(n)}-\mathbf{A}_{n} \mathbf{X}_{n}\right\|_{\text{F}}^{2}+\tau_n \left\|\mathbf{Z}_{n}\right\|_{\gamma}\right.\\
&\left.+\frac{\rho_n}{2}\left\|\mathbf{X}_n-\mathbf{X}_n^{k}\right\|_{\text{F}}^{2}+\left\langle\Gamma_{n}^\mathbf{X}, \mathbf{X}_n-\mathbf{Z}_n\right\rangle
+\frac{\rho_{n}}{2}\left\|\mathbf{X}_{n}-\mathbf{Z}_{n}\right\|_{\text{F}}^{2}\right).
\end{aligned}
\end{equation}
where $\Gamma_{n}^\mathbf{X}$ are Lagrangian multipliers.

With other variables fixed, the minimization subproblem for $\mathbf{Z}_n$ can be deduced from (\ref{equation:X_alm}) as follows:
\begin{equation}
\label{equation:Zn}
\mathbf{Z}_n^{k+1}= \arg \min_{\mathbf{Z}_n}  \left\|\mathbf{Z}_{n}\right\|_{\gamma}+\frac{\hat{\rho}_{n}}{2}\left\|\mathbf{Z}_{n}-\mathbf{P}_n^k\right\|_{\text{F}}^{2},
\end{equation}
where $\hat{\rho}_{n}=\rho_{n}/\tau_n$, $\mathbf{P}_n^k=\mathbf{X}_{n}^{k}+\Gamma_{n}^{\mathbf{X}}/\rho_n$.
Let $\sigma_{1}^{k} \geq \sigma_{2}^{k} \geq \cdots \geq \sigma_{t_n}^{k}$ represent the singular values of $\mathbf{Z}_n^{k} \in \mathbb{R}^{r_n \times s_n}$ with $t_n=\min \left\{r_n, s_n\right\}$ and $\nabla \phi\left(\sigma_{n}^{k}\right)$ denote the gradient of $\phi(x) = 1-e^{-x/\gamma}$ at $\sigma_{n}^{k} .$ Let
$$f(\mathbf{Z}_n)=\frac{1}{2}\left\|\mathbf{Z}_n-\mathbf{P}_n^k\right\|_{\text{F}}^{2}.$$
It is easy to prove that the gradient of $f(\mathbf{Z}_n)$ is Lipschitz continuous by setting the Lipschitz constant being $1.$ As stated in \cite{Non_LRMA}, considering the nonascending order of singular values and according to the antimonotone property of gradient of our nonconvex function, we have
\begin{equation}
\begin{aligned}
& 0 \leq \nabla \phi\left(\sigma_{1}^{k}\right) \leq \nabla \phi\left(\sigma_{2}^{k}\right) \leq \cdots \leq \nabla \phi\left(\sigma_{t_n}^{k}\right), \\
& \phi\left(\sigma_{i}(\mathbf{Z}_n)\right) \leq \phi\left(\sigma_{i}^{k}\right)+\nabla \phi\left(\sigma_{i}^{k}\right)\left(\sigma_{i}(\mathbf{Z}_n)-\sigma_{i}^{k}\right),
\end{aligned}
\label{equ:gradient}	
\end{equation}
where $i=1,2,\cdots,t_n$.
Following (\ref{equ:gradient}), the subproblem with respect to $\mathbf{Z}_n$ (\ref{equation:Zn}) can be written as following relaxation:
\begin{equation}
\begin{split}
&\arg \min _{\mathbf{Z}_n} \frac{1}{\hat{\rho}_{n}} \sum_{n=1}^{t_n} \phi\left(\sigma_{n}^{k}\right)+\nabla \phi\left(\sigma_{n}^{k}\right)\left(\sigma_{n}(\mathbf{Z}_n)-\sigma_{n}^{k}\right)+f(\mathbf{Z}_n) \\
&=\arg \min _{\mathbf{Z}_n} \frac{1}{\hat{\rho}_{n}} \sum_{n=1}^{t_n} \nabla \phi\left(\sigma_{n}^{k}\right) \sigma_{n}(\mathbf{Z}_n)+\frac{1}{2}\left\|\mathbf{Z}_n-\mathbf{P}_{n}^{k}\right\|_{\text{F}}^{2}.	
\end{split}
\label{equ:relaxation}
\end{equation}

Then, based on \cite{lu_nonconvex, Non_LRMA},
the solution of (\ref{equ:relaxation}) can be efficiently obtained by generalized weight singular value thresholding (WSVT) \cite{WSVT} , as shown in Lemma 1.

\noindent\textbf{Lemma 1}: For any $1 / \hat{\rho}_{n}>0$, the given data $\mathbf{P}_n^k=\mathbf{X}_{n}^{k}+\Gamma_{n}^{\mathbf{X}}/\rho_n$, and $0 \leq \nabla \phi\left(\sigma_{1}^{k}\right) \leq \nabla \phi\left(\sigma_{2}^{k}\right) \leq \cdots \leq \nabla \phi\left(\sigma_{t_n}^{k}\right)$,
a globally optimal solution $\mathbf{Z}_n^{*}$ to (\ref{equ:relaxation}) is given as follows:
\begin{equation}
\label{equation_solution_Zn}
\quad \mathbf{Z}_n^{k+1}= \operatorname{WSVT}\left(\mathbf{P}_n^{k}, \frac{\nabla \phi}{\hat{\rho}_{n}}\right) =\mathbf{U} \mathbf{S}_{\frac{\nabla \phi}{\hat{\rho}_{n}}}(\mathbf{\Sigma}) \mathbf{V}^{T},
\end{equation}
where $\mathbf{P}_n^{k}=\mathbf{U} \mathbf{\Sigma} \mathbf{V}^{T}$ is the SVD of $\mathbf{P}_n^{k}$;
$$\mathbf{S}_{\frac{\nabla \phi}{\hat{\rho}_{n}}}(\mathbf{\Sigma})=\operatorname{Diag}\left\{\max \left(\mathbf{\Sigma}_{ii}-\frac{\nabla \phi \left( \mathbf{\sigma}_{i}^{k}\right)}{\hat{\rho}_{n}}, 0\right)\right\},$$
and $i=1,2,\cdots,t_n$.

With other variables fixed, the minimization subproblem for $\mathbf{X}_n$, $n=1,2,\cdots, N$, can be deduced from (\ref{equation:X_alm}) as follows:
\begin{equation}
\label{equationforX}
\begin{aligned}
\mathbf{X}_n^{k+1}&= \arg\min_{\mathbf{X}_n}\frac{\alpha_{n}}{2}\left\|\mathbf{Y}_{(n)}-\mathbf{A}_{n}^{k} \mathbf{X}_{n}\right\|_{\text{F}}^{2}\\
&+\frac{\rho_{n}}{2}\left\|\mathbf{X}_{n}-\frac{\mathbf{Z}_{n}^{k+1}-\Gamma_{n}^k/\mu_n+\mathbf{X}_n^{k}}{2}\right\|_{\text{F}}^{2}.
\end{aligned}
\end{equation}
They are convex and have the following closed-form solutions
\begin{equation}
\label{equation:solution_Xn}
\begin{aligned}
\mathbf{X}_n^{k+1}=&\frac{1}{2}(\alpha_{n}\mathbf{A}_n^T\mathbf{A}_n+2\rho_n \mathbf{I}_n)^{-1}\left[2\alpha_{n}\mathbf{A}_n^T\mathbf{Y}_{(n)}\right.\\
&\left. +\mu_n \left(\mathbf{Z}_{n}^{k+1}-\Gamma_{n}^k/\mu_n+\mathbf{X}_n^{k}\right)\right].
\end{aligned}
\end{equation}

The Lagrangian multipliers $\Gamma_{n}^{\mathbf{X}}$ can be updated by the following equation
\begin{equation}
\label{equation:Lambda_2}
\Gamma_{n}^{\mathbf{X}} = \Gamma_{n}^{\mathbf{X}} + \mathbf{X}_n-\mathbf{Z}_n.
\end{equation}

\subsubsection{Update $\mathbf{A}_n$}
With fixing other variables,
the optimization subproblem with respect to $\mathbf{A}_n$, $n=1,2,\cdots,N$, in (\ref{equation:XAY}) can be written as follows:
\begin{equation}
\label{equation_A_aux}
\begin{aligned}
%\mathbf{A}_n^{k+1}=
\arg\min_{\{\mathbf{A}_n\}_{n=1}^{N}} \sum_{n=1}^{N} & \left(\frac{\alpha_{n}}{2}\left\|\mathbf{Y}_{(n)}-\mathbf{A}_{n} \mathbf{X}_{n}\right\|_{\text{F}}^{2}+\lambda_n \left\|\mathbf{A}_{n}\right\|_{\gamma}\right.\\
&\left.+\frac{\rho_n}{2}\left\|\mathbf{A}_n-\mathbf{A}_n^{k}\right\|_{\text{F}}^{2}\right).
\end{aligned}
\end{equation}
To efficiently solve the above optimization,
we first introduce auxiliary variables $\mathbf{A}_{n}=\mathbf{J}_{n}, n=1,2,\cdots,N$.
%(\ref{equation_A_aux}) can be rewritten as
%\begin{equation}
%\label{equation_A_alm}
%\begin{aligned}
%\arg\min_{\mathbf{A}_n} \sum_{n=1}^{N}& (\frac{\alpha_{n}}{2}\left\|\mathbf{Y}_{(n)}-\mathbf{A}_{n} \mathbf{X}_{n}\right\|_{\text{F}}^{2}+\lambda_n \left\|\mathbf{J}_{n}\right\|_{\gamma}\\
%&+\frac{\rho_n}{2}\left\|\mathbf{A}_n-\mathbf{A}_n^{k}\right\|_{\text{F}}^{2}),
%s.t., \mathbf{A}_{n}=\mathbf{J}_{n}.
%\end{aligned}
%\end{equation}
Then, by the ALM method, the problem (\ref{equation_A_aux}) can also be reformulated as
\begin{equation}
\label{equation:A_alm}
\begin{aligned}
& \arg\min_{\{\mathbf{A}_n,\mathbf{J}_n\}_{n=1}^{N}}  \sum_{n=1}^{N} \left(\frac{\alpha_{n}}{2}\left\|\mathbf{Y}_{(n)}-\mathbf{A}_{n} \mathbf{X}_{n}\right\|_{\text{F}}^{2}+\lambda_n \left\|\mathbf{J}_{n}\right\|_{\gamma}\right.\\
& \left. +\frac{\rho_n}{2}\left\|\mathbf{A}_n-\mathbf{A}_n^{k}\right\|_{\text{F}}^{2} +\left\langle\Gamma_{n}^\mathbf{A},\mathbf{A}_n-\mathbf{J}_n\right\rangle +\frac{\rho_{n}}{2}\left\|\mathbf{A}_{n}-\mathbf{J}_{n}\right\|_{\text{F}}^{2}\right),
\end{aligned}
\end{equation}
where $\Gamma_{n}^\mathbf{A}$ are the Lagrangian multipliers.

With other variables fixed, the minimization subproblem for $\mathbf{J}_n$ can be deduced from (\ref{equation:A_alm}) as follows:
\begin{equation}
\displaystyle \mathbf{J}_n^{k+1}= \arg \min_{\mathbf{J}_n} \lambda_n \left\|\mathbf{J}_{n}\right\|_{\gamma}+\frac{\rho_{n}}{2}\left\|\mathbf{J}_{n}-\mathbf{Q}_n^k\right\|_{\text{F}}^{2}.
\end{equation}
where $\tilde{\rho}_{n}=\rho_{n}/\lambda_n$;
$\mathbf{Q}_n^k=\mathbf{A}_{n}^{k}+\Gamma_{n}^{\mathbf{A}}/\rho_n$.
Its solution can also be obtained by \textbf{Lemma 1} as follows:
\begin{equation}
\label{equation_solution_Jn}
\quad \mathbf{J}_n^{k+1}=\operatorname{WSVT}\left(\mathbf{Q}_n^{k}, \frac{\nabla \phi}{\tilde{\rho}_{n}}\right).
\end{equation}

With other variables fixed, the minimization subproblem for $\mathbf{A}_n$, $n=1,2,\cdots, N$, can be deduced from (\ref{equation:A_alm}) as follows:
\begin{equation}
\label{equationforA}
\begin{aligned}
\mathbf{A}_n^{k+1}= &\arg\min_{\mathbf{A}_n} \frac{\alpha_{n}}{2}\left\|\mathbf{Y}_{(n)}-\mathbf{A}_{n} \mathbf{X}_{n}\right\|_{\text{F}}^{2}\\
&+\rho_{n}\left\|\mathbf{A}_{n}-\frac{\mathbf{J}_{n}^{k+1}-\Gamma_{n}^\mathbf{A}/\rho_n+\mathbf{A}_n^{k}}{2}\right\|_{\text{F}}^{2}.
\end{aligned}
\end{equation}
They are also convex and have the following closed-form solutions
\begin{equation}
\label{equation_solution_An}
\begin{aligned}
\mathbf{A}_{n}^{k+1}=&\left(\mathbf{Y}_{(n)}^{k}\left(\mathbf{X}_{n}^{k+1}\right)^{T}+\rho_n (\mathbf{J}_{n}^{k+1}-\Gamma_{n}^\mathbf{A}/\rho_n+\mathbf{A}_n^{k})\right)\\
&\left(\mathbf{X}_{n}^{k+1}\left(\mathbf{X}_{n}^{k+1}\right)^{T}+2\rho_{n} \mathbf{I}_{n}\right)^{\dagger}.
\end{aligned}
\end{equation}

Finally, the Lagrangian multipliers $\Gamma_{n}^{\mathbf{A}}$ can be updated by the following equation
\begin{equation}
\label{equation:Lambda_1}
\Gamma_{n}^{\mathbf{A}} = \Gamma_{n}^{\mathbf{A}} + \mathbf{A}_n-\mathbf{J}_n.
\end{equation}

\subsubsection{Update $\mathcal{Y}$}
With other variables fixed, the minimization subproblem with respect to $\mathcal{Y}$ in (\ref{equation:XAY}) can be written as
\begin{equation}
\begin{aligned}
& \arg\min_{\{\mathbf{Y}_{(n)}\}_{n=1}^{N}} \sum_{n=1}^{N}  \frac{\alpha_{n}}{2}\left\|\mathbf{Y}_{(n)}-\mathbf{A}_{n} \mathbf{X}_{n}\right\|_{\text{F}}^{2}+\frac{\rho}{2}\left\|\mathcal{Y}-\mathcal{Y}^{k}\right\|_{\text{F}}^{2} \\
& s.t., \mathcal{P}_{\Omega}(\mathcal{Y})=\mathcal{F}.
\end{aligned}
\end{equation}
Then, the update of $\mathcal{Y}^{k+1}$ can be written explicitly as
\begin{equation}
\label{equation_solution_Y}
\begin{array}{l}
\displaystyle \mathcal{Y}^{k+1}=\mathcal{P}_{{\Omega}^c}\left(\sum_{n=1}^{N} \alpha_{n} \text { fold }_{n}\left(\frac{\mathbf{A}_{n}^{k+1} \mathbf{X}_{n}^{k+1}+\rho_n \mathbf{Y}_{(n)}^{k}}{1+\rho_n}\right)\right)+\mathcal{F},
\end{array}
\end{equation}
where $\Omega^{C}$ is the complementary set of $\Omega$.

%$\mathcal{F}$ is the observed data;
%$P_{{\Omega}}$ is an operator defined in subsection \ref{operators}.

\begin{algorithm}[!t]
	\caption{:Algorithm for the LRATM model.} \label{algorithm:A1}
	\begin{algorithmic}[1]
		\Require
		The observed tensor $\mathcal{F}$; The set of index of observed entries $\Omega$; Stopping criterion $\varepsilon$, the given $n$-rank, $r=\left(r_{1}, r_{2}, r_{3}\right)$.
		%The given $n$-rank, $r = (r_1, r_2, r_3)$; 		
		%		regularization parameters.
		%		$\lambda, \tau$, and $\mu$.
		\Ensure		
		The completed tensor.		
		\State Initialize:  $\mathbf{X}_n=\mathbf{Z}_n=0, \mathbf{A}_n=\mathbf{J}_n=0,\Gamma_{n}^\mathbf{X}=0, \Gamma_{n}^\mathbf{A}=0,  n=1,2, \cdots, N$; $k=0$.
		\State Repeat until convergence:		
		\State Update $\mathbf{X}, \mathbf{Z}, \mathbf{A}, \mathbf{J}, \mathcal{Y},  \Gamma^{\mathbf{X}}, \Gamma^{\mathbf{A}}$ via
		
		1st step: Update $\mathbf{Z}_n$ via	(\ref{equation_solution_Zn})
		
		2nd step: Update $\mathbf{X}_n$ via	(\ref{equation:solution_Xn})
		
		3rd step: Update $\mathbf{A}_n$ via (\ref{equation_solution_An})
		
		4th step: Update $\mathbf{J}_n$ via	(\ref{equation_solution_Jn})
		
		5th step: Update $\mathcal{Y}$ via (\ref{equation_solution_Y})
		
		6th step: Update the parameter via 	(\ref{equation:Lambda_2}), (\ref{equation:Lambda_1})
		
		\State Check the convergence condition.
	\end{algorithmic}
\end{algorithm}

\subsection{ Complexity and Convergence Analysis}

The proposed algorithm for our LRATM model is summarized in Algorithm \ref{algorithm:A1}.
Further, we discuss the complexity and convergence of the proposed algorithm.

\subsubsection{Complexity analysis}

The cost of computing $\mathbf{X}_{n}$ is $O\left(I_{n} r_{n}^{2}+I_{n} r_{n} s_{n}+r_{n}^{2} s_{n}\right)$,
calculating $\mathbf{Z_n}$ has a complexity of $O\left( \Pi_{j \neq n} I_{j} \times r_{n}^2 \right)$,
the complexity of updating $\mathbf{J}_n$ is $O\left(I_{n} r_{n}^2\right)$,
calculating $\mathbf{A}_{n}$ has a complexity of $O\left(I_{n} r_{n}^{2}+I_{n} r_{n} s_{n}+r_{n}^{2} s_{n}\right)$,
calculating $\mathcal{Y}$ has a complexity of
$O\left(r_{1} I_{1} s_{1}+\cdots+r_{N} I_{N} s_{N}\right)$. Therefore, the total complexity of the proposed algorithm can be obtained by counting the complexity of the above variables, i.e.,
\begin{equation} \label{equation:complexity_model1}
O\left(\sum_{n}(3I_{n} r_{n}^2+\Pi_{j \neq n} I_{j} \times r_{n}^2+3 I_{n} r_{n} S_{n}+2 r_{n}^{2} s_{n})\right).
\end{equation}

\subsubsection{Convergence analysis}

In this section, we theoretically analyze the convergence of the proposed
algorithm by using the BSUM method \cite{BSUM}.

\noindent \textbf{Lemma 2} \cite{BSUM}:
Let us assume that the feasible set $\mathcal{X}$ is the cartesian product of $n$ closed convex sets: $\mathcal{X}=\mathcal{X}_1 \times \mathcal{X}_2 \times \cdots \times \mathcal{X}_{n}$.
Given the problem
\begin{equation}
\min f(x), s.t. ~ x \in \mathcal{X},
\end{equation}
assume $h\left(x, x^{k-1}\right)$ is an approximation of $f(x)$ at the $(k-1)$-th iteration,
which satisfies the following conditions:
\begin{equation}
\begin{array}{l}
1) \quad h_{i}\left(y_{i}, y\right)=f(y), \forall y \in \mathcal{X}, \forall i; \\
2) \quad h_{i}\left(x_{i}, y\right) \geq f\left(y_{1}, \ldots, y_{i-1}, x_{i}, y_{i+1}, \ldots, y_{n}\right), \\
\quad \quad \forall x_{i} \in \mathcal{X}_{i}, \forall y \in \mathcal{X}, \forall i;\\
3) \quad \left.h_{i}^{\prime}\left(x_{i}, y ; d_{i}\right)\right|_{x_i=y_i}=f^{\prime}(y ; d), \forall d=\left(0, \cdots, d_{i} \cdots 0\right) \\
\quad \quad \text { s.t. } y_{i}+d_{i} \in \mathcal{X}_{i}, \forall i;\\
4) \quad 	h_{i}\left(x_{i}, y\right) \text{is continuous in} \left(x_{i}, y\right), \forall i;
\end{array}
\end{equation}
where $h_{i}\left(x_{i}, y\right)$ is the sub-problem with respect to the $i$-th block and $f^{\prime}(y ; d)$ is the direction derivative of fat the point $y$ in direction $d$.
Suppose $h_{i}\left(x_{i}, y\right)$ is quasi-convex in $x_{i}$ for $i=1,2, \cdots, n$.
Furthermore, assume that each sub-problem $\arg\min h_i\left(x_{i}, x^{k-1}\right), s.t. ~ x \in \mathcal{X}_{i}$ has a unique solution for any point $x^{k-1} \in \mathcal{X} .$
Then, the iterates generated by the BSUM algorithm converge to the set of coordinatewise minimum of $f$.

\noindent \textbf{Theorem 1.} The iterates generated by (\ref{equation_original_PPA}) converge to the set of coordinatewise minimizers.

\textbf{Proof.} It is easy to verify that $g\left(\mathcal{S}, \mathcal{S}^{k}\right)$ is an approximation and a global upper bound of $f(\mathcal{S})$ at the $k$-th iteration, which satisfies the following conditions:
\begin{equation}
\begin{array}{l}
1) \quad g_{i}\left(\mathcal{S}_{i}, \mathcal{S}\right)=f(\mathcal{S}), \forall \mathcal{S}, i=1,2,3; \\
2) \quad g_{i}\left(\bar{\mathcal{S}}_{i}, \mathcal{S}\right) \geq f\left(\mathcal{S}_{1}, \cdots, \bar{\mathcal{S}_{i}}, \cdots, \mathcal{S}_{3}\right), \\
\quad\quad \forall \bar{\mathcal{S}}_{i}, \forall \mathcal{S}, i=1,2,3; \\
3) \quad \left.g_{i}^{\prime}\left(\bar{\mathcal{S}}_{i}, \mathcal{S} ; \mathbf{M}_{i}\right)\right|_{\bar{\mathcal{S}}_{i}=\mathcal{S}_{i}}=f^{\prime}\left(\mathcal{S} ; \mathbf{M}^i\right), \\
\quad \quad \forall \mathbf{M}^{i}=\left(0, \ldots, \mathbf{M}_{i},\ldots, 0\right); \\
4) \quad 	g_{i}\left(\bar{\mathcal{S}}_{i}, \mathcal{S}\right) \text{is continuous in} \left(\bar{\mathcal{S}}_{i}, \mathcal{S} \right), i=1,2,3;
%	\left.g_{2}^{\prime}\left(\bar{\mathcal{Z}}_{2}, \mathcal{Z} ; M_{2}\right)\right|_{\bar{\mathcal{Z}}_{2}=\mathcal{Z}_{2}}=f^{\prime}\left(\mathcal{Z} ; M^{2}\right), \forall M^{2}=\left(0, M_{2}, 0\right) \\
%	g_{3}^{\prime}\left(\bar{\mathcal{Z}}_{3}, \mathcal{Z} ; M_{3}\right) | _{\bar{\mathcal{Z}}_{3}= \mathcal{Z}_{3}}=f^{\prime}\left(\mathcal{Z} ; M^{3}\right), \forall M^{3}=\left(0,0, M_{3}\right)\\
%	g_{i}\left(\bar{\mathcal{Z}}_{i}, \mathcal{Z}\right) \text{is continuous in} \left(\bar{\mathcal{Z}}_{i}, \mathcal{Z}\right) i=1,2,3,
\end{array}	
\end{equation}
where $\mathcal{S}=(\mathbf{X}, \mathbf{A}, \mathcal{Y}),$ and $\mathcal{S}_{i}$ equal $\mathbf{X}, \mathbf{A}, \mathcal{Y}$ for $i=1,2,3,$ respectively.
In addition, the subproblem $g_{i}(i=1,2,3)$ is quasi-convex with respect to $\mathbf{X}, \mathbf{A}$ and $\mathcal{Y}$ respectively and each sub-problem of $g_{i}$ has a unique solution.
Therefore, all assumptions in \textbf{Lemma 1} are satisfied. According to the conclusion of \textbf{Lemma 1}, the \textbf{Theorem 1} is valid, and the proposed algorithm is theoretically convergent.

\section{Numerical experiments}
\label{Numerical experiments}

In this section, the proposed model is evaluated on three types of public tensor datasets,
i.e., video datasets, MRI dataset and HSI dataset
which have been frequently used to interpret the tensor completion performance of different models.
To demonstrate its effectiveness,
we compare the proposed model with TMac method \cite{Tmac}, MF-TV method \cite{MFTV}, single TNN based method \cite{lu2019TRPCA} and PSTNN based method \cite{PSTNN}.

\begin{table}[t]
	\caption{The averaged PSNR, SSIM, FSIM, ERGA and SAM of the completed results on video "suzie" by Tmac, MF-TV, TNN, PSTNN and our model with different sampling rates. The best values are highlighted in bolder fonts.}
	\centering
	\label{table_video_suzie}
	\setlength{\tabcolsep}{1mm}{
	\begin{tabular}{cccccccc}
		\hline \hline 				
		&& &SR =0.05 &&&&	\\
		PQI &	nosiy&	our model&		MF-TV&	TMac&	PSTNN&	TNN\\		
		PSNR	&	7.259	&	\textbf{29.464}	&		13.801	&	23.385	&	17.447	&	22.005	\\
		SSIM	&	0.009	&	\textbf{0.807}	&		0.094	&	0.622	&	0.192	&	0.563	\\
		FSIM	&	0.454	&	\textbf{0.885}	&		0.42	&	0.792	&	0.59	&	0.776	\\
		ERGA	&	1057.282	&	\textbf{83.571}	&		501.117	&	167.927	&	327.678	&	194.844	\\
		MSAM	&	77.324	&	\textbf{3.622}	&		24.095	&	6.927	&	13.775	&	7.797	\\
		\hline
		&&&SR = 0.1&&&&\\
		PQI &	nosiy&	our model&		MF-TV&	TMac&	PSTNN&	TNN\\
		PSNR&	7.493&	\textbf{32.056}&		22.356&	26.189&	26.647&	26.032\\
		SSIM&	0.014&	\textbf{0.878}&		0.605&	0.74&	0.68&	0.692\\
		FSIM&	0.426&	\textbf{0.924}&		0.758&	0.838&	0.843&	0.846\\
		ERGA&	1029.096&	\textbf{62.314}&		196.059&	124.369	&117.104&	124.923\\
		MSAM&	71.725&	\textbf{2.764}&		6.99&	5.423&	5.171&	5.405	\\	
		\hline	
		&&&SR = 0.2&&&&\\
		PQI &	nosiy&	our model&		MF-TV&	TMac&	PSTNN&	TNN\\
		PSNR&	8.005&	\textbf{34.378}&		32.064&	27.274&	30.566&	30.561\\
		SSIM&	0.02&	\textbf{0.918}&		0.872&	0.782&	0.829&	0.831\\
		FSIM&	0.391&	\textbf{0.948}&		0.916&	0.853&	0.91&	0.911\\
		ERGA&	970.285&	\textbf{47.877}&		66.692&	109.627&	75.472&	75.598\\
		MSAM&	63.522&	\textbf{2.183}&		2.81&	4.812&	3.399&	3.395\\		
		\hline \hline
	\end{tabular}}
\end{table}

To accurately evaluate the performance of the test models,
two types of standards are employed to quantitatively evaluate the quality of the completed tensors.
The first one is the visual evaluation of the completed data, which is a qualitative evaluation standard.
The second one is the five quantitative picture quality indices (PQIs),
including the peak signal-to-noise ratio (PSNR) \cite{PSNR},
structural similarity index (SSIM) \cite{SSIM},
feature similarity (FSIM) \cite{FSIM},
erreur relative globale adimensionnelle de synth\`ese (ERGAS) \cite{EGRAS},
the mean the spectral angle mapper (SAM) \cite{SAM}.
Larger PSNR, SSIM, FSIM and smaller ERGAS, SAM are, the better the completion performance of the corresponding model is.
Since the experimental datasets are all third-order tensors,
the PQIs for each frontal slice in the completed tensor are first calculated,
and then the mean of these PQIs are finally used to evaluate the performance of the models.
All experiments are performed on MATLAB 2019b, the CPU of the computer is Inter core i7@2.2GHz and the memory is 64GB. The code will be posted on the following URL: https://github.com/NavyZeng/LRATM.

%For a tensor $\mathcal{Y} \in \mathbb{R}^{I_1 \times \ldots \times I_N}$,
%let $S_{\text{number}}$ denote the number of sampled entries in its index set $\Omega$.
%Then the sampling ratio (SR) can be defined as:
%\begin{equation}
%\mathrm{SR}=\frac{S_{\text{number}}}{\prod_{n=1}^{N} I_{n}},
%\end{equation}where the sampled entries are chosen randomly from a tensor $\mathcal{Y}$ by a uniform distribution.
%In the proposed algorithms for model-1 and model-2, the inputs include the observed tensor $\mathcal{F} \in \mathbb{R}^{I_1 \times I_2 \times I_3}$,
%the stopping criteria $\epsilon$, the regularized parameters $\alpha, \beta, \lambda, \tau,$ and the penalty parameter $\beta$.
%All parameters are empirically.
%Specifically, the stopping criterion $\epsilon$ and the weights $\alpha_{i}(i=1,2,3)$ of the proposed model-1 and model-2 are set to be $10^{-5}$ and $1/3$ for all experiments;
%the regularization parameter $\mu$ and the penalty parameter $\beta$ for model-2 are set as 0.5 and 10, respectively;
%finally, the proximal parameter $\rho$ and regularized parameters $\lambda, \tau$ are all set as 0.1 for all experiments of model-1 and model-2.

\subsection{Video Data}

\begin{figure*}[!t]	
	\centering			
	\subfloat[Original]{\includegraphics[width=0.13\linewidth]{Image_video_b94}}%
	\hfil	
	\subfloat[95\% Masked]{\includegraphics[width=0.13\linewidth]{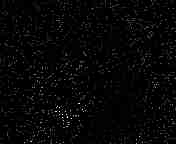}}%
	\hfil					
	\subfloat[our model]{\includegraphics[width=0.13\linewidth]{Image_video_b94_sr005LxLxNN}}%
	\hfil
	%	\subfloat[our model-1]{\includegraphics[width=0.23\linewidth]{myfigure/Image_result/Image_video_b94_sr0.05/DBNN}}%
	%	\hfil
	\subfloat[MF-TV]{\includegraphics[width=0.13\linewidth]{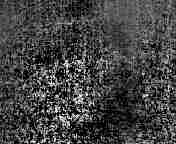}}%
	\hfil					
	\subfloat[Tmac]{\includegraphics[width=0.13\linewidth]{Image_video_b94_sr005Tmac}}%
	\hfil
	\subfloat[PSTNN]{\includegraphics[width=0.13\linewidth]{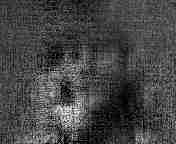}}%
	\hfil
	\subfloat[TNN]{\includegraphics[width=0.13\linewidth]{Image_video_b94_sr005TNN}}%	
	\caption{One slice of the recovered video for “suzie” by our model, MF-TV, Tmac, PSTNN and TNN.  The sampling rate is 5\%.}
	\label{figure_video_sr0.05}
\end{figure*}

\begin{figure*}[!t]	
	\centering			
	\subfloat[Original]{\includegraphics[width=0.13\linewidth]{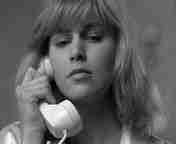}}%
	\hfil	
	\subfloat[90\% Masked]{\includegraphics[width=0.13\linewidth]{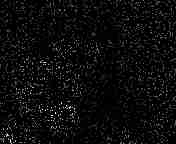}}%
	\hfil					
	\subfloat[our model]{\includegraphics[width=0.13\linewidth]{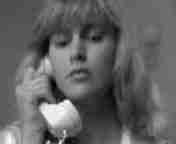}}%
	\hfil
	%	\subfloat[our model-1]{\includegraphics[width=0.23\linewidth]{myfigure/Image_result/Image_video_b10_sr0.1/DBNN}}%
	%	\hfil
	\subfloat[MF-TV]{\includegraphics[width=0.13\linewidth]{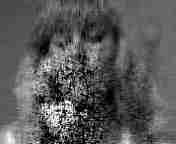}}%
	\hfil					
	\subfloat[Tmac]{\includegraphics[width=0.13\linewidth]{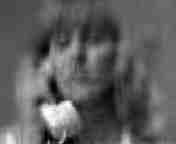}}%
	\hfil
	\subfloat[PSTNN]{\includegraphics[width=0.13\linewidth]{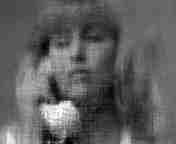}}%
	\hfil
	\subfloat[TNN]{\includegraphics[width=0.13\linewidth]{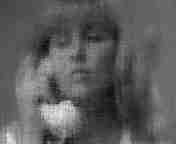}}%	
	\caption{One slice of the recovered video for “suzie” by our model, MF-TV, Tmac, PSTNN and TNN. The sampling rate is 10\%.}
	\label{figure_video_sr0.1}
\end{figure*}

\begin{figure*}[!t]	
	\centering			
	\subfloat[Original]{\includegraphics[width=0.13\linewidth]{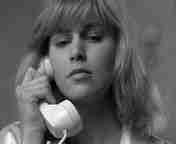}}%
	\hfil	
	\subfloat[80\% Masked]{\includegraphics[width=0.13\linewidth]{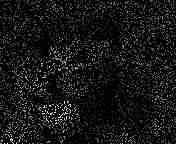}}%
	\hfil					
	\subfloat[our model]{\includegraphics[width=0.13\linewidth]{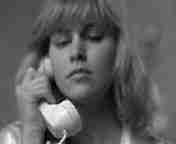}}%
	\hfil
	%	\subfloat[our model-1]{\includegraphics[width=0.23\linewidth]{myfigure/Image_result/Image_video_b10_sr0.2/DBNN}}%
	%	\hfil
	\subfloat[MF-TV]{\includegraphics[width=0.13\linewidth]{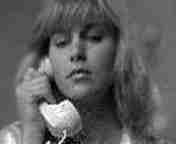}}%
	\hfil					
	\subfloat[Tmac]{\includegraphics[width=0.13\linewidth]{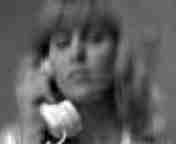}}%
	\hfil
	\subfloat[PSTNN]{\includegraphics[width=0.13\linewidth]{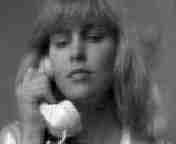}}%
	\hfil
	\subfloat[TNN]{\includegraphics[width=0.13\linewidth]{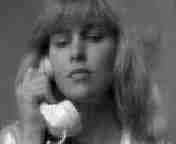}}%	
	\caption{One slice of the recovered video for “suzie” by our model, MF-TV, Tmac, PSTNN and TNN.  The sampling rate is 20\%.}
	\label{figure_video_sr0.2}
\end{figure*}

\begin{figure*}[!t]	
	\centering			
	\subfloat[Original]{\includegraphics[width=0.13\linewidth]{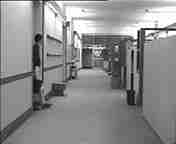}}%
	\hfil	
	\subfloat[95\% Masked]{\includegraphics[width=0.13\linewidth]{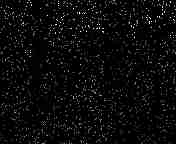}}%
	\hfil					
	\subfloat[our model]{\includegraphics[width=0.13\linewidth]{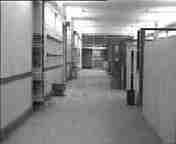}}%
	\hfil
	%	\subfloat[our model-1]{\includegraphics[width=0.23\linewidth]{myfigure/Image_result/Image_hall_b21_sr0.05/DBNN}}%
	%	\hfil
	\subfloat[MF-TV]{\includegraphics[width=0.13\linewidth]{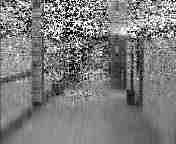}}%
	\hfil					
	\subfloat[Tmac]{\includegraphics[width=0.13\linewidth]{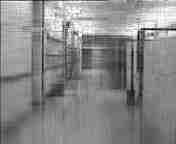}}%
	\hfil
	\subfloat[PSTNN]{\includegraphics[width=0.13\linewidth]{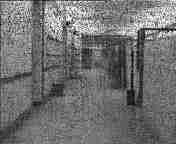}}%
	\hfil
	\subfloat[TNN]{\includegraphics[width=0.13\linewidth]{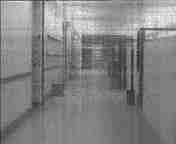}}%	
	\caption{One slice of the recovered video “hall” by our model, MF-TV, Tmac, PSTNN and TNN.  The sampling rate is 5\%.}
	\label{figure_hall_sr0.05}
\end{figure*}

In this subsection, we compare our model with MF-TV, Tmac, TNN and PSTNN on two video datasets: "suzie" and “hall”\footnote{http://trace.eas.asu.edu/yuv/},
both of which are colored using YUV format,
and two slices of them are shown in Fig. \ref{figure_video_sr0.05} and Fig. \ref{figure_hall_sr0.05}, respectively.
Their sizes both are 144 $\times$ 176 $\times$ 150.
We test all five methods on a series of sampling rates (SR): 5\%, 10\% and 20\%,
and all the test models are evaluated in terms of quantitative comparison and visual evaluation. In addition, the $n$-rank is approximated by using the number of the largest 0.5\% singular values. 

For quantitative comparison, Table \ref{table_video_suzie} and Table \ref{table_video_hall} report the PQIs of the results completed by different methods.
The best result for each PQI are marked in bold.
From Table \ref{table_video_suzie} and Table \ref{table_video_hall},
it can be found that our model obtains the highest indices among the five tested models in all SR cases;
Tmac obtains the second best PQIs, when the SR is set to 5\% or 10\%;
While MF-TV obtains the second best PQIs, when SR is set to 20\%.
The margins between the results by our model and the second best results are more than 5dB considering the PSNR.

\begin{figure*}[!t]	
	\centering			
	\subfloat[Original]{\includegraphics[width=0.13\linewidth]{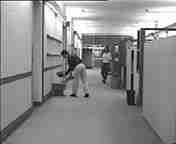}}%
	\hfil	
	\subfloat[90\% Masked]{\includegraphics[width=0.13\linewidth]{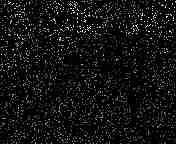}}%
	\hfil					
	\subfloat[our model]{\includegraphics[width=0.13\linewidth]{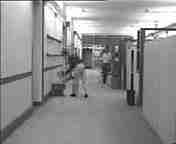}}%
	\hfil
	%	\subfloat[our model-1]{\includegraphics[width=0.23\linewidth]{myfigure/Image_result/Image_hall_b104_sr0.1/DBNN}}%
	%	\hfil
	\subfloat[MF-TV]{\includegraphics[width=0.13\linewidth]{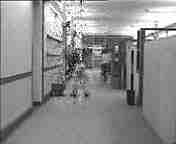}}%
	\hfil					
	\subfloat[Tmac]{\includegraphics[width=0.13\linewidth]{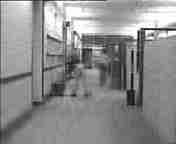}}%
	\hfil
	\subfloat[PSTNN]{\includegraphics[width=0.13\linewidth]{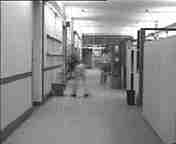}}%
	\hfil
	\subfloat[TNN]{\includegraphics[width=0.13\linewidth]{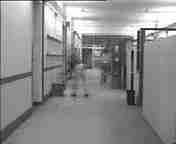}}%	
	\caption{One slice of the recovered video “hall” by our model, MF-TV, Tmac, PSTNN and TNN.  The sampling rate is 10\%.}
	\label{figure_hall_sr0.1}
\end{figure*}

For visual evaluation, we illustrate one frontal slice of the completed results with different sampling rates
in Fig. \ref{figure_video_sr0.05},
Fig. \ref{figure_video_sr0.1},
Fig. \ref{figure_video_sr0.2},
Fig. \ref{figure_hall_sr0.05} and Fig. \ref{figure_hall_sr0.1}.
It is clear from the figures that the results of our model are closest to the ground truth than other tested models, especially at low sampling rates.
Specifically, as shown in Fig. \ref{figure_hall_sr0.1}, Fig. \ref{figure_video_sr0.05} and Fig. \ref{figure_video_sr0.1},
when the sampling rates are 0.05 and 0.1, the advantages of our model are most obvious.
Our model restores most of the structural information of the videos,
while the videos restored by the competitive models contain only the outline of the images.
At a higher sampling rate, as shown in Fig. \ref{figure_video_sr0.2} and Fig. \ref{figure_hall_sr0.1},
our model and competitive models both complete the main structural information of the images,
but our model recovers more texture and detail information.

\begin{table} [t]
	\centering	
	\caption{The averaged PSNR, SSIM, FSIM, ERGA and SAM of the recovered results on video "hall" by Tmac, MF-TV, TNN, PSTNN and our model with different sampling rates. The best values are highlighted in bolder fonts.}
	\label{table_video_hall}
	\setlength{\tabcolsep}{1mm}{
	\begin{tabular}{cccccccc}
		\hline \hline 				
		&&&SR =0.05 &&&&	\\
		PQI&	nosiy&	our model&		MF-TV&	TMac&	PSTNN&	TNN\\
		PSNR	&	4.82	&	\textbf{28.347}	&		13.539	&	22.101	&	16.075	&	20.78	\\
		SSIM	&	0.007	&	\textbf{0.894}	&		0.412	&	0.675	&	0.36	&	0.636	\\
		FSIM	&	0.387	&	\textbf{0.920}	&		0.612	&	0.789	&	0.672	&	0.792	\\
		ERGA	&	1225.779	&	\textbf{83.146}	&	452.351	&	168.866	&	335.52	&	195.315	\\
		MSAM	&	77.299	&	\textbf{2.360}		&	12.865	&	3.818	&	8.64	&	4.299	\\
		\hline	
		&&&SR = 0.1&&&&\\		
		PQI	&	nosiy	&	our model		&	MF-TV	&	TMac	&	PSTNN	&	TNN	\\
		PSNR	&	5.055	&	\textbf{31.804}		&	24.855	&	26.936	&	29.014	&	28.433	\\
		SSIM	&	0.013	&	\textbf{0.935}		&	0.829	&	0.854	&	0.892	&	0.905	\\
		FSIM	&	0.393	&	\textbf{0.950}		&	0.873	&	0.888	&	0.934	&	0.936	\\
		ERGA	&	1193.075	&	\textbf{56.998}		&	131.422	&	97.185	&	77.395	&	82.259	\\
		MSAM	&	71.7	&\textbf{	1.904}		&	3.669	&	2.404	&	2.417	&	2.46	\\	
		\hline
		&&&SR = 0.2&&&&\\
		PQI	&	nosiy	&	our model		&	MF-TV	&	TMac	&	PSTNN	&	TNN	\\
		PSNR	&	5.567	&	\textbf{33.941}		&	33.006	&	27.648	&	33.629	&	33.691	\\
		SSIM	&	0.025	&	\textbf{0.952}		&	0.94	&	0.869	&	0.961	&	0.962	\\
		FSIM	&	0.403	&	\textbf{0.964}		&	0.954	&	0.897	&	0.973	&	0.974	\\
		ERGA	&	1124.737	&	\textbf{44.581}		&	50.971	&	89.271	&	46.123	&	45.851	\\
		MSAM	&	63.507	&	\textbf{1.574}		&	1.779	&	2.226	&	1.584	&	1.565	\\		
		\hline \hline
	\end{tabular}}
\end{table}

\subsection{Magnetic resonance imaging data}

\begin{figure*}[!t]	
	\centering
	\subfloat[]{\includegraphics[width=0.2\linewidth]{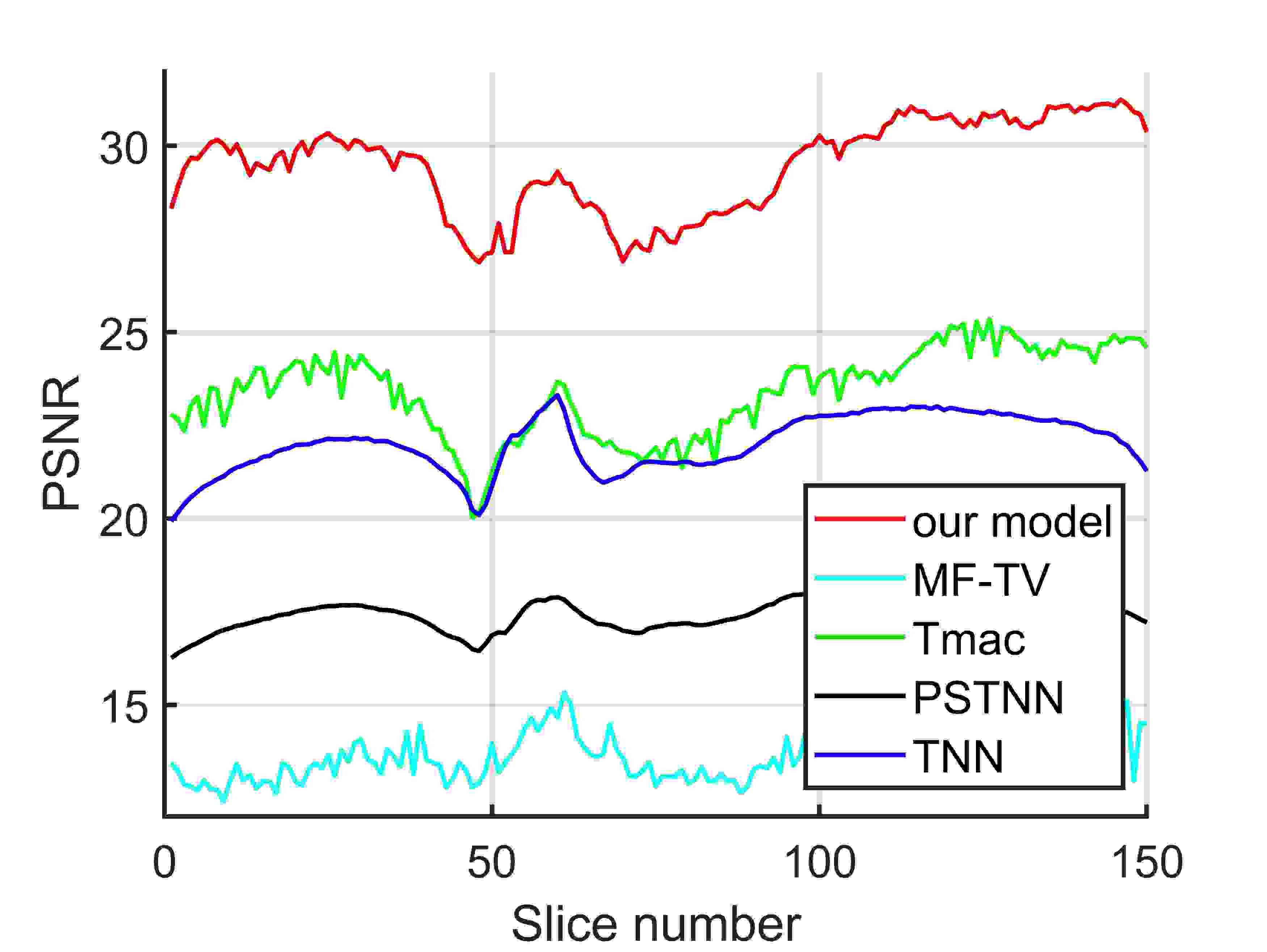}}
	\hfil
	\subfloat[]{\includegraphics[width=0.2\linewidth]{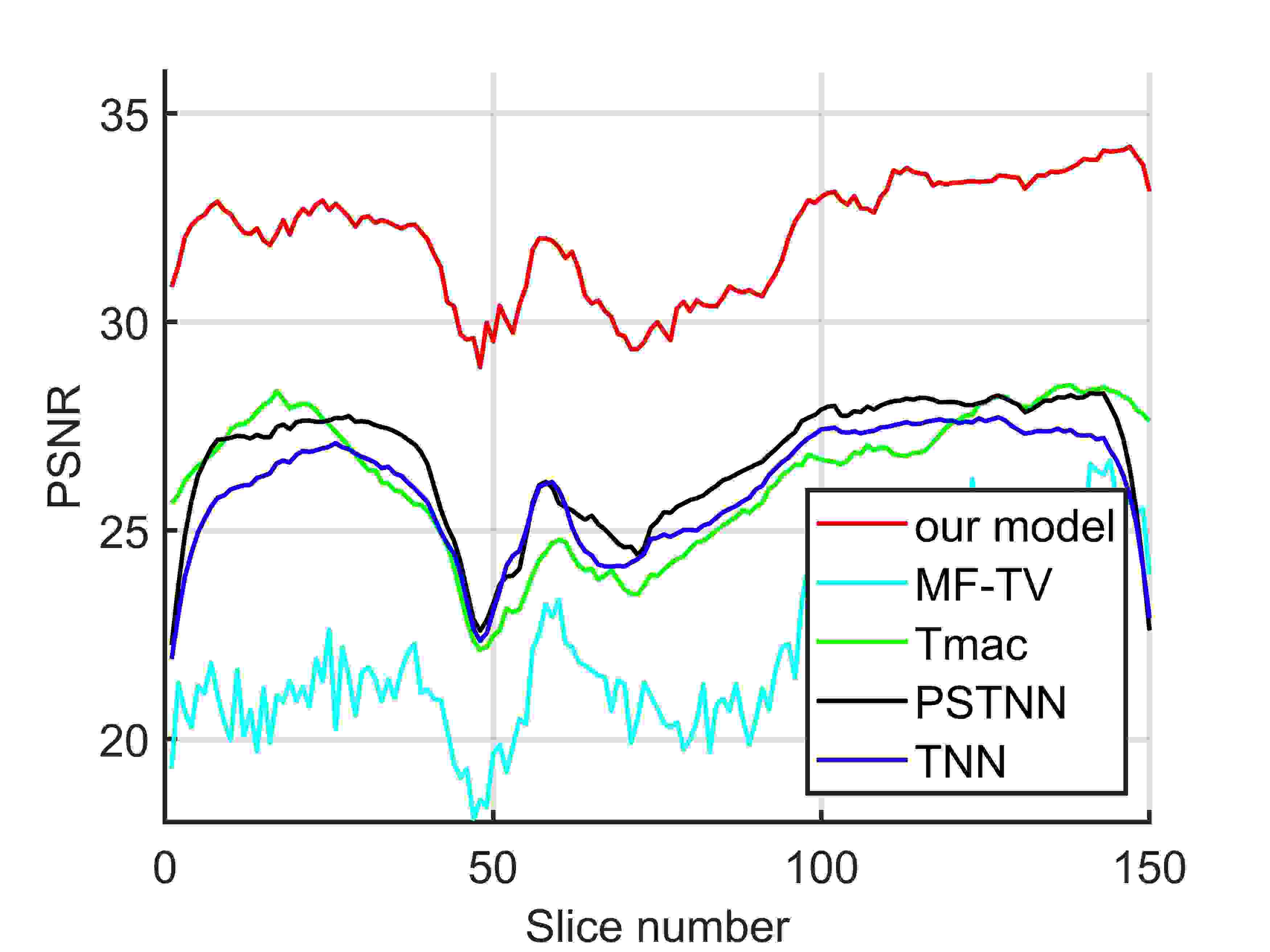}}
	\hfil
	\subfloat[]{\includegraphics[width=0.2\linewidth]{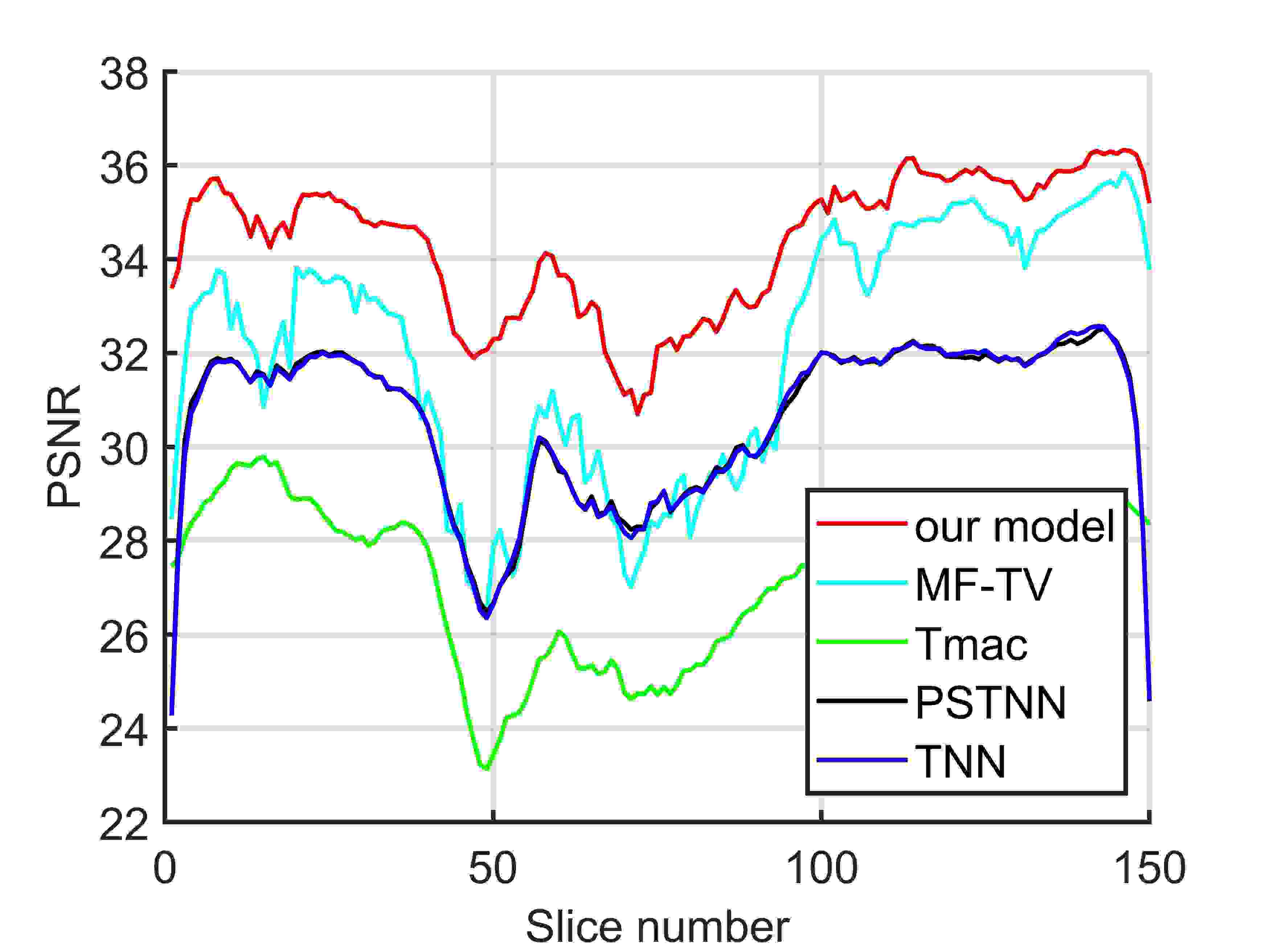}}
	\hfil
	\subfloat[]{\includegraphics[width=0.2\linewidth]{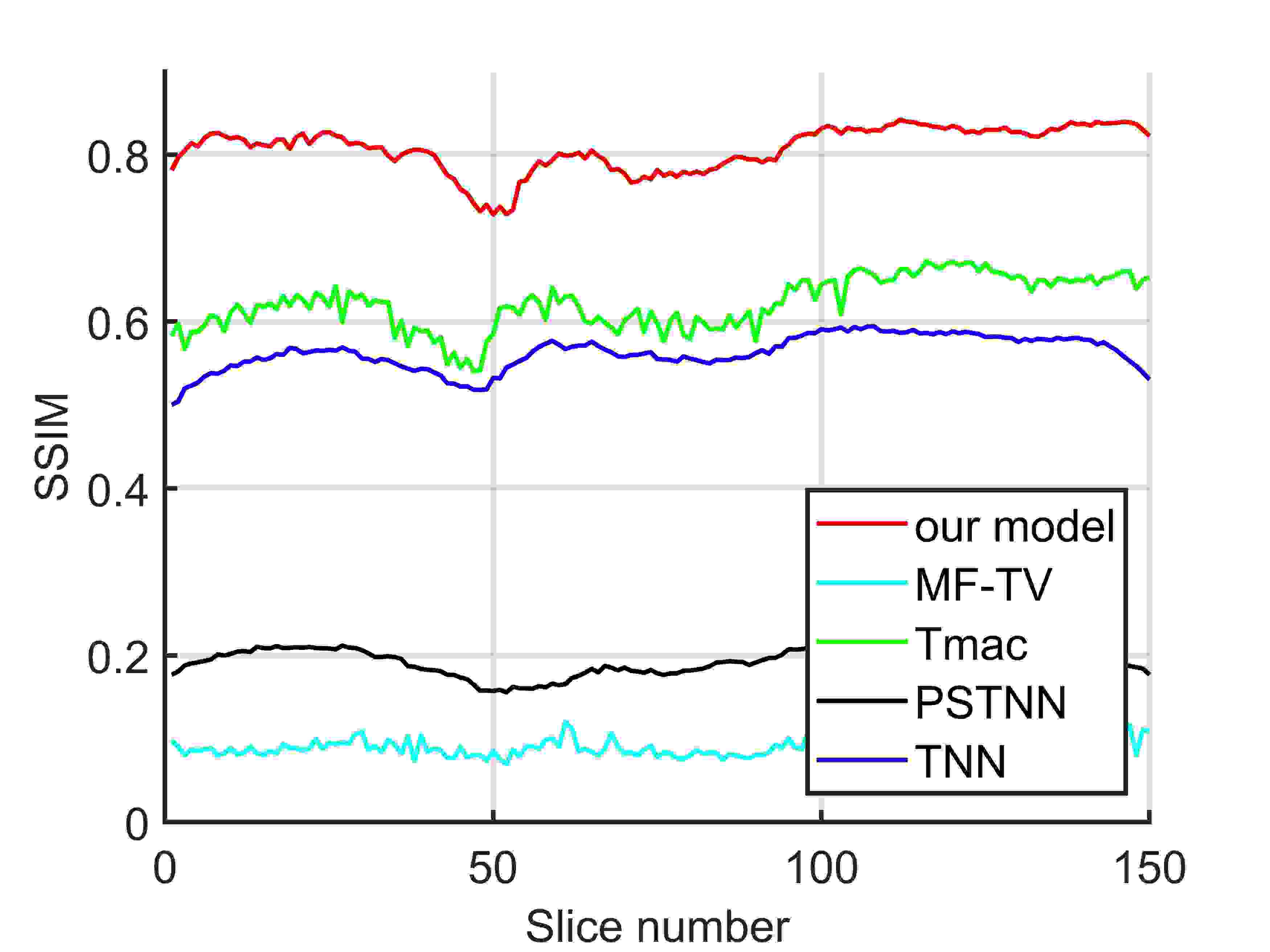}}
	\hfil
	\subfloat[]{\includegraphics[width=0.2\linewidth]{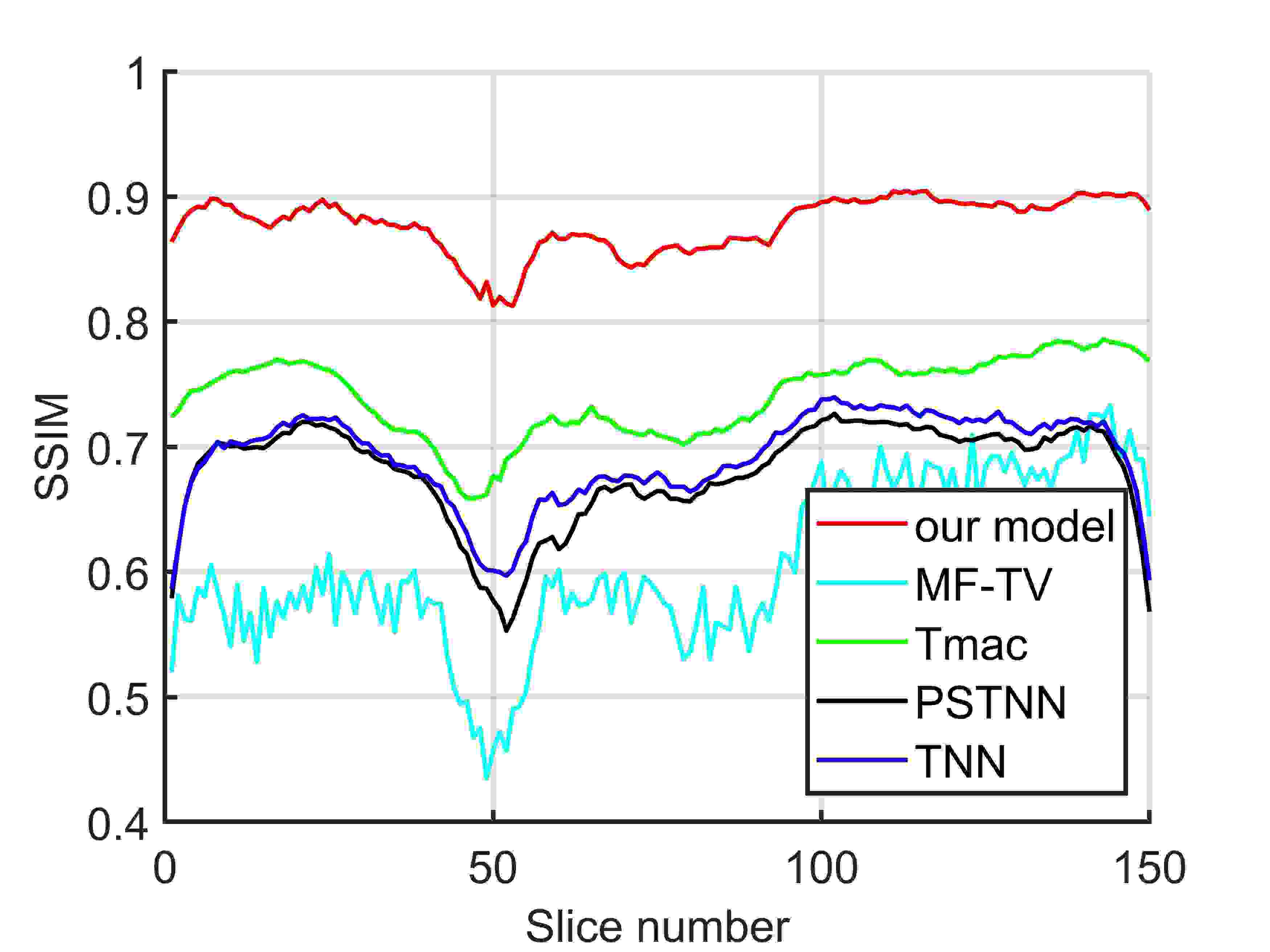}}
	\hfil
	\subfloat[]{\includegraphics[width=0.2\linewidth]{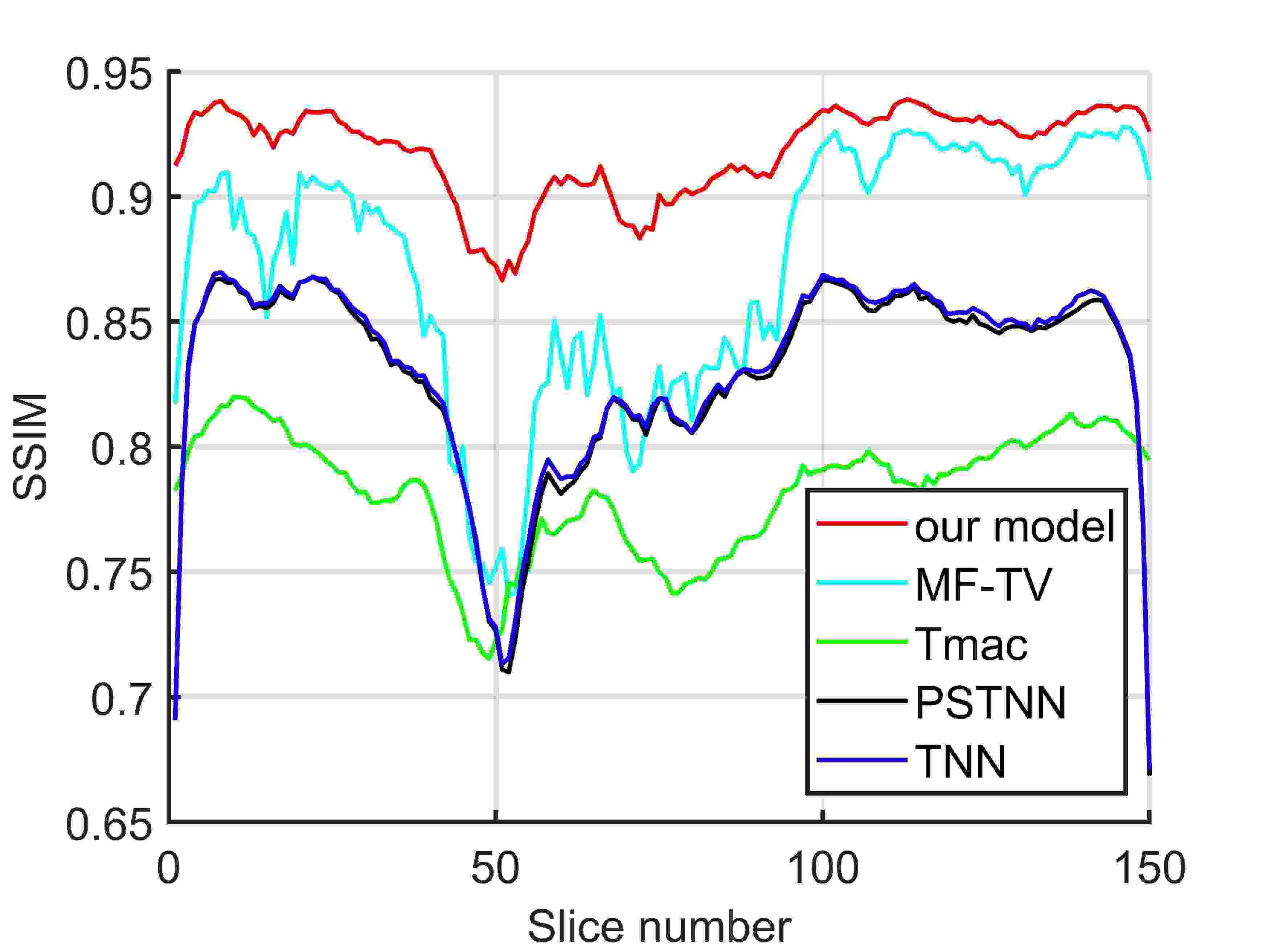}}	
	\hfil				
	\subfloat[]{\includegraphics[width=0.2\linewidth]{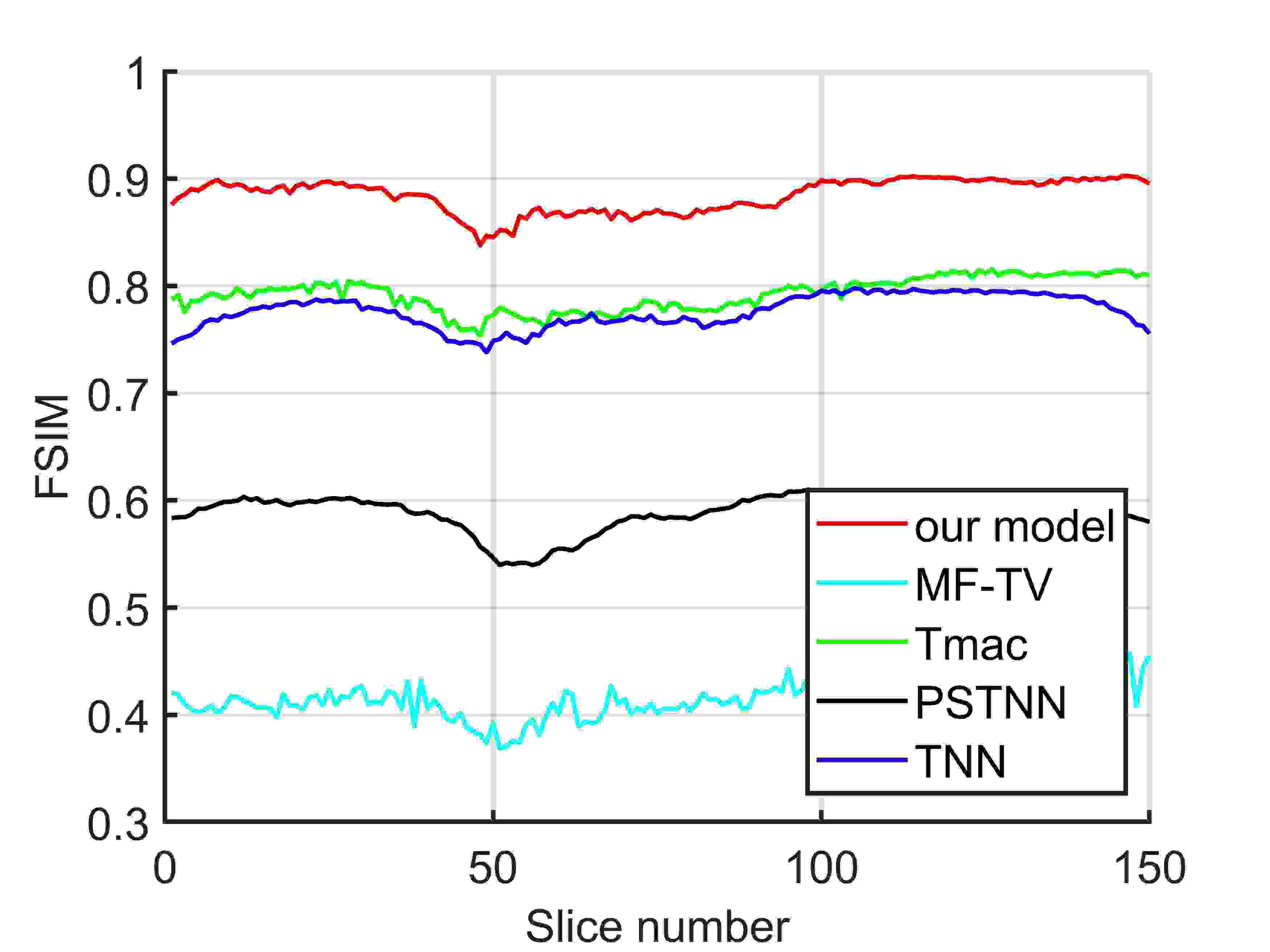}}%
	\hfil
	\subfloat[]{\includegraphics[width=0.2\linewidth]{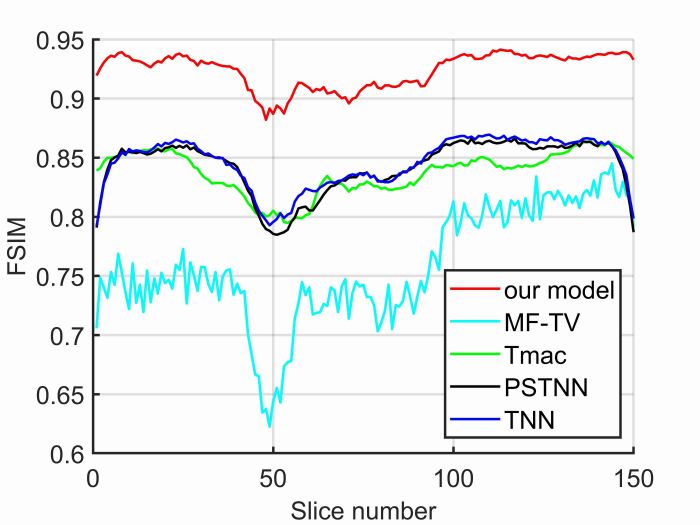}}%
	\hfil
	\subfloat[]{\includegraphics[width=0.2\linewidth]{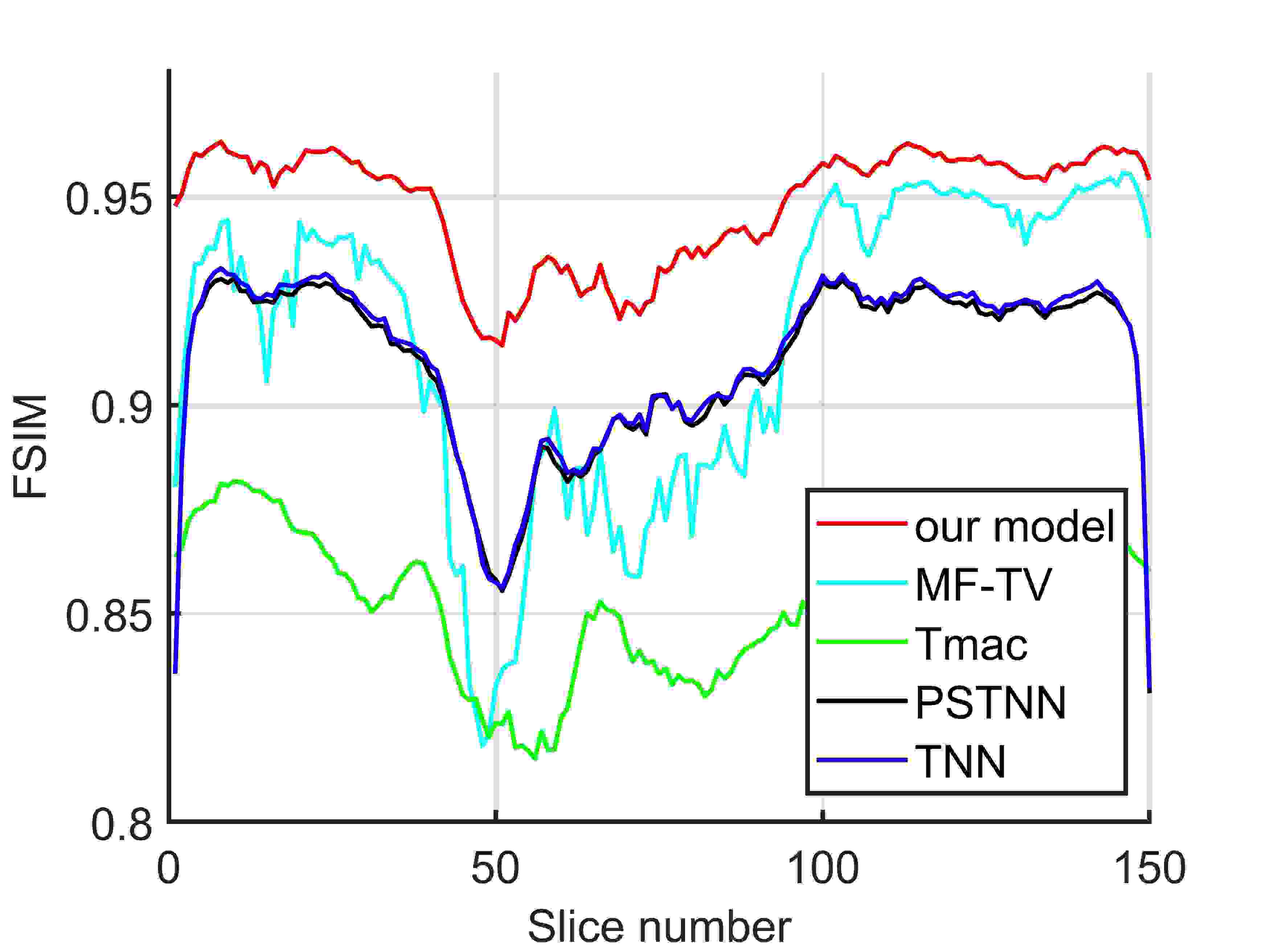}}%
	\caption{The PSNR, SSIM and FSIM of the recovered video "suzie" by MF-TV, Tmac, TNN, PSTNN and our model for all slices, respectively.}
	\label{PSNR and SSIM of video}
\end{figure*}

\begin{figure*}[!t]	
	\centering			
	\subfloat[Original]{\includegraphics[width=0.13\linewidth]{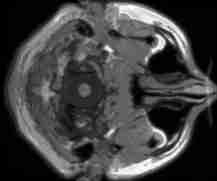}}%
	\hfil	
	\subfloat[90\% Masked]{\includegraphics[width=0.13\linewidth]{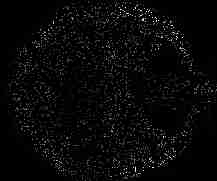}}%
	\hfil					
	\subfloat[our model]{\includegraphics[width=0.13\linewidth]{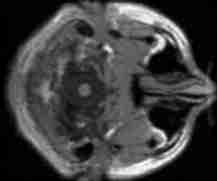}}%
	\hfil
	%	\subfloat[our model-1]{\includegraphics[width=0.23\linewidth]{myfigure/Image_result/Image_MRI_b7_sr0.1/DBNN}}%
	%	\hfil
	\subfloat[MF-TV]{\includegraphics[width=0.13\linewidth]{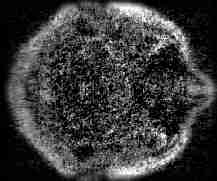}}%
	\hfil					
	\subfloat[Tmac]{\includegraphics[width=0.13\linewidth]{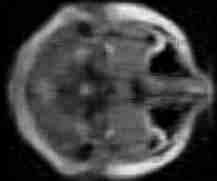}}%
	\hfil
	\subfloat[PSTNN]{\includegraphics[width=0.13\linewidth]{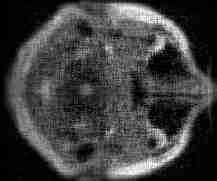}}%
	\hfil
	\subfloat[TNN]{\includegraphics[width=0.13\linewidth]{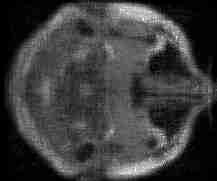}}%	
	\caption{One slice of the recovered MRI by our model, MF-TV, Tmac, PSTNN and TNN.  The sampling rate is 10\%.}
	\label{figure_MR_sr0.1_1}
\end{figure*}

\begin{figure*}[!t]		
	\centering			
	\subfloat[Original]{\includegraphics[width=0.13\linewidth]{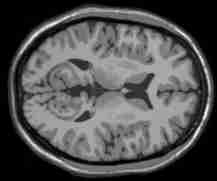}}%
	\hfil	
	\subfloat[90\% Masked]{\includegraphics[width=0.13\linewidth]{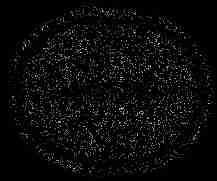}}%
	\hfil					
	\subfloat[our model]{\includegraphics[width=0.13\linewidth]{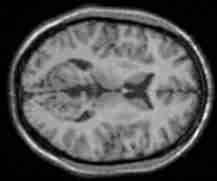}}%
	\hfil
	%	\subfloat[our model-1]{\includegraphics[width=0.23\linewidth]{myfigure/Image_result/Image_MRI_b83_sr0.1/DBNN}}%
	%	\hfil
	\subfloat[MF-TV]{\includegraphics[width=0.13\linewidth]{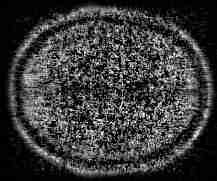}}%
	\hfil					
	\subfloat[Tmac]{\includegraphics[width=0.13\linewidth]{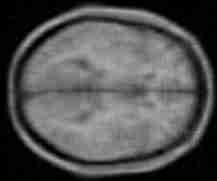}}%
	\hfil
	\subfloat[PSTNN]{\includegraphics[width=0.13\linewidth]{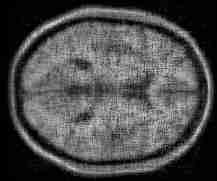}}%
	\hfil
	\subfloat[TNN]{\includegraphics[width=0.13\linewidth]{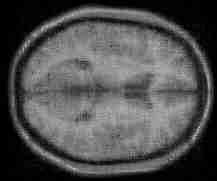}}%	
	\caption{One slice of the recovered MRI by our model, MF-TV, Tmac, PSTNN and TNN.  The sampling rate is 10\%.}
	\label{figure_MR_sr0.1_2}
\end{figure*}

\begin{figure*}[!t]	
	\centering			
	\subfloat[Original]{\includegraphics[width=0.13\linewidth]{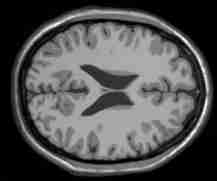}}%
	\hfil	
	\subfloat[90\% Masked]{\includegraphics[width=0.13\linewidth]{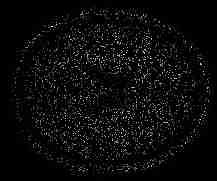}}%
	\hfil					
	\subfloat[our model]{\includegraphics[width=0.13\linewidth]{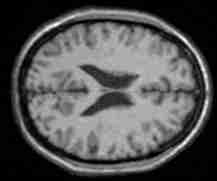}}%
	\hfil
	%	\subfloat[our model-1]{\includegraphics[width=0.23\linewidth]{myfigure/Image_result/Image_MRI_b95_sr0.1/DBNN}}%
	%	\hfil
	\subfloat[MF-TV]{\includegraphics[width=0.13\linewidth]{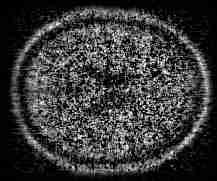}}%
	\hfil					
	\subfloat[Tmac]{\includegraphics[width=0.13\linewidth]{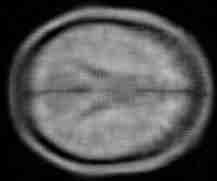}}%
	\hfil
	\subfloat[PSTNN]{\includegraphics[width=0.13\linewidth]{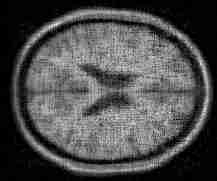}}%
	\hfil
	\subfloat[TNN]{\includegraphics[width=0.13\linewidth]{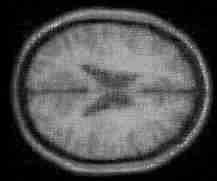}}%	
	\caption{One slice of the recovered MRI by our model, MF-TV, Tmac, PSTNN and TNN.  The sampling rate is 10\%.}
	\label{figure_MR_sr0.1_3}
\end{figure*}

\begin{figure*}[!t]	
	\centering			
	\subfloat[Original]{\includegraphics[width=0.13\linewidth]{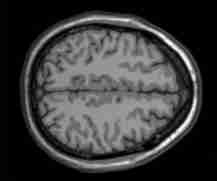}}%
	\hfil	
	\subfloat[90\% Masked]{\includegraphics[width=0.13\linewidth]{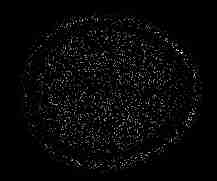}}%
	\hfil					
	\subfloat[our model]{\includegraphics[width=0.13\linewidth]{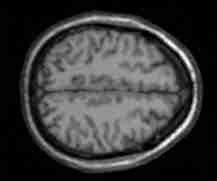}}%
	\hfil
	%	\subfloat[our model-1]{\includegraphics[width=0.23\linewidth]{myfigure/Image_result/Image_MRI_b118_sr0.1/DBNN}}%
	%	\hfil
	\subfloat[MF-TV]{\includegraphics[width=0.13\linewidth]{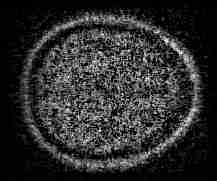}}%
	\hfil					
	\subfloat[Tmac]{\includegraphics[width=0.13\linewidth]{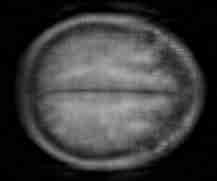}}%
	\hfil
	\subfloat[PSTNN]{\includegraphics[width=0.13\linewidth]{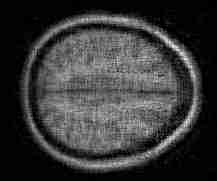}}%
	\hfil
	\subfloat[TNN]{\includegraphics[width=0.13\linewidth]{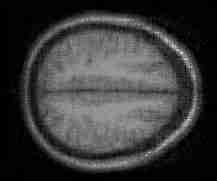}}%	
	\caption{One slice of the recovered MRI by our model, MF-TV, Tmac, PSTNN and TNN.  The sampling rate is 10\%.}
	\label{figure_MR_sr0.1_4}
\end{figure*}

\begin{figure*}[!t]	
	\centering
	\subfloat[]{\includegraphics[width=0.2\linewidth]{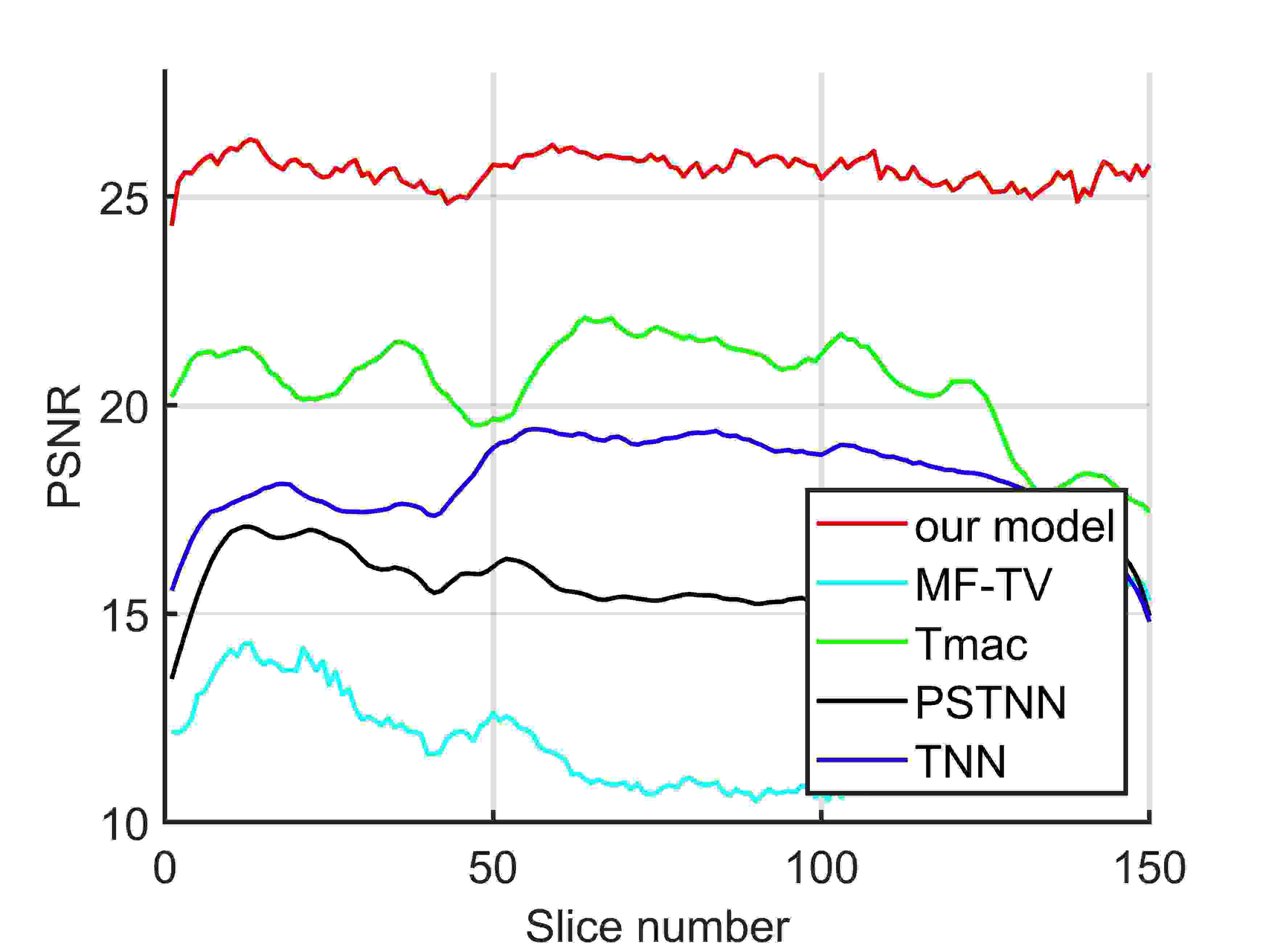}}
	\hfil	
	\subfloat[]{\includegraphics[width=0.2\linewidth]{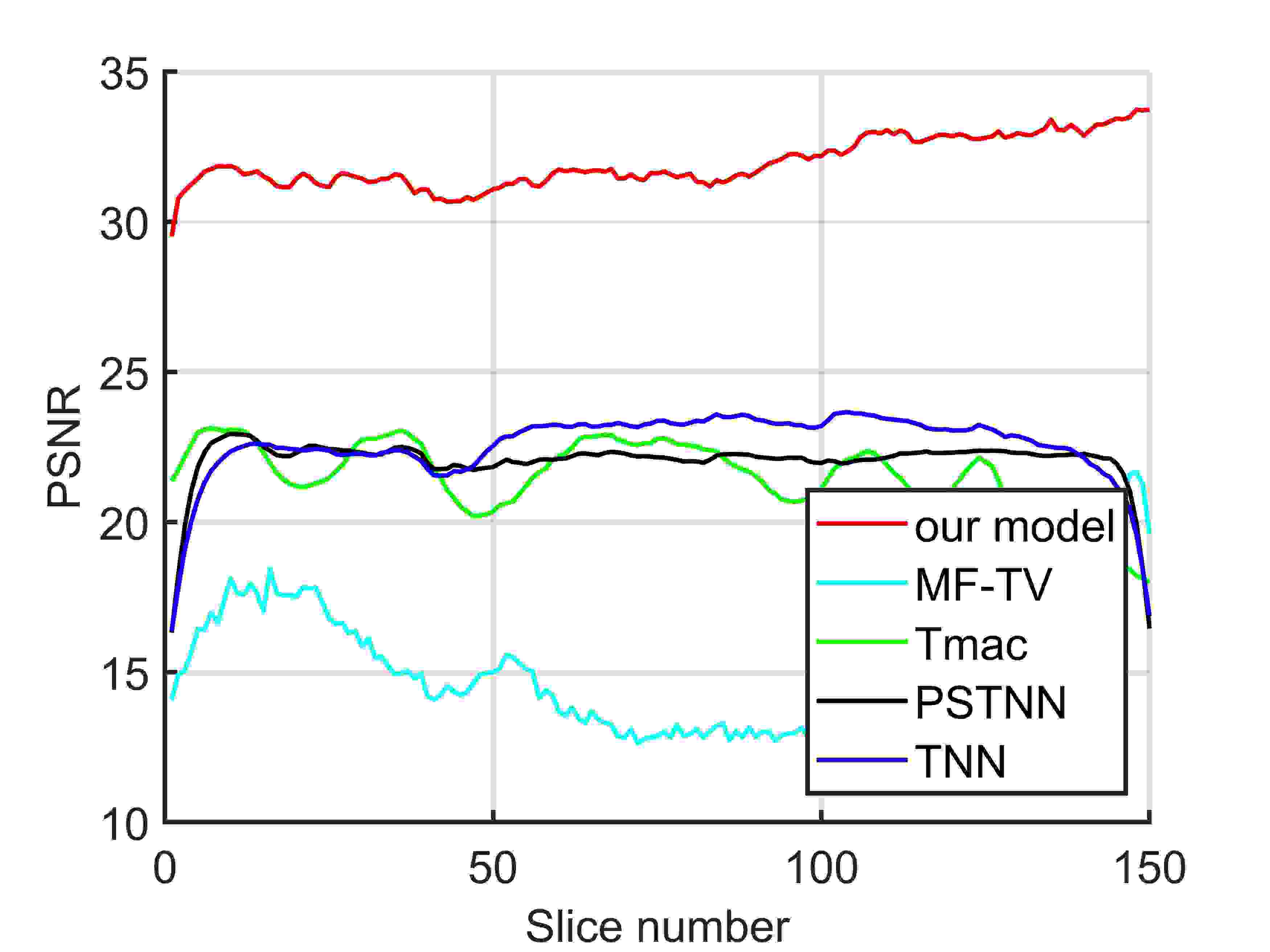}}
	\hfil
	\subfloat[]{\includegraphics[width=0.2\linewidth]{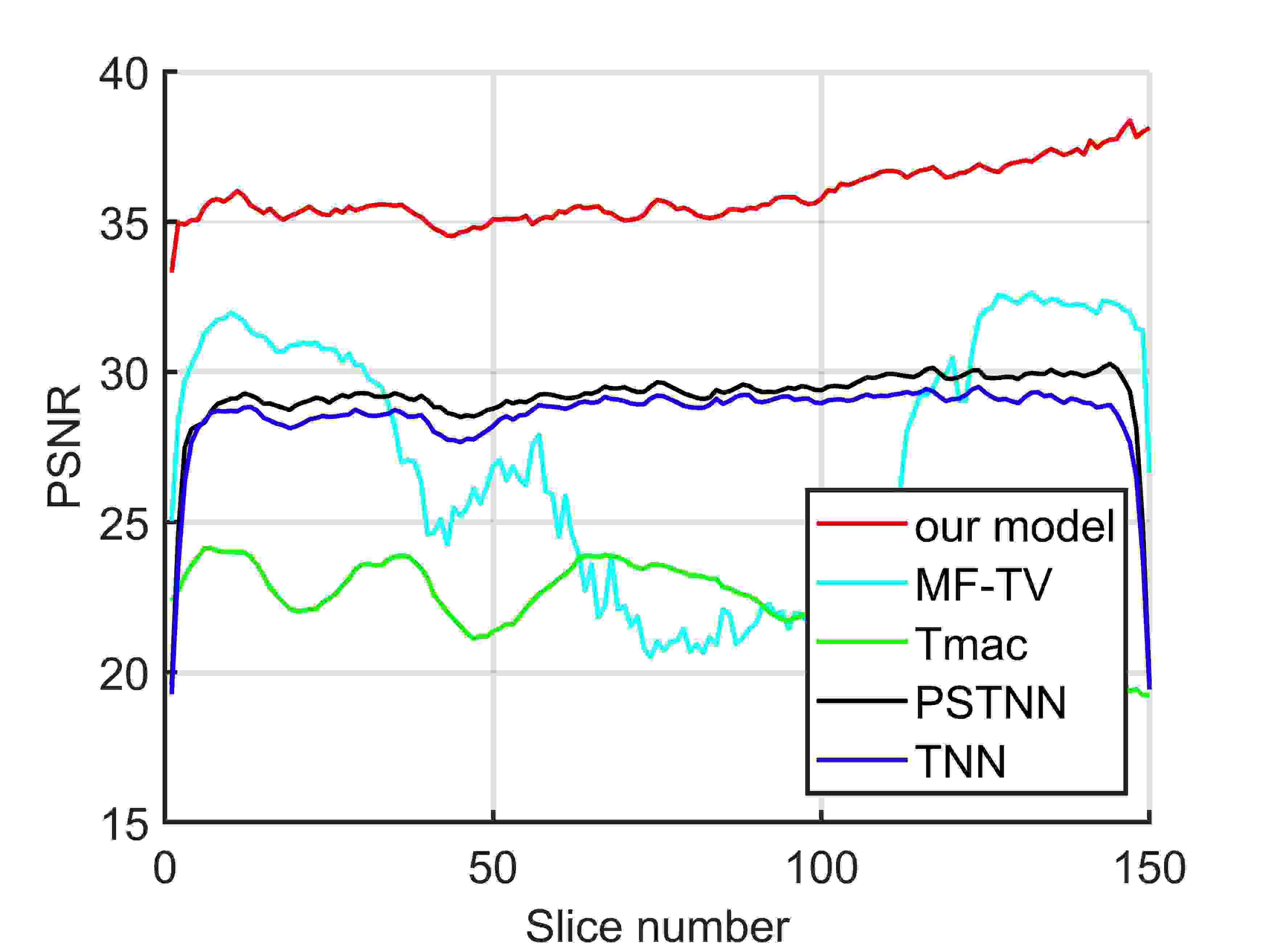}}
	\hfil
	\subfloat[]{\includegraphics[width=0.2\linewidth]{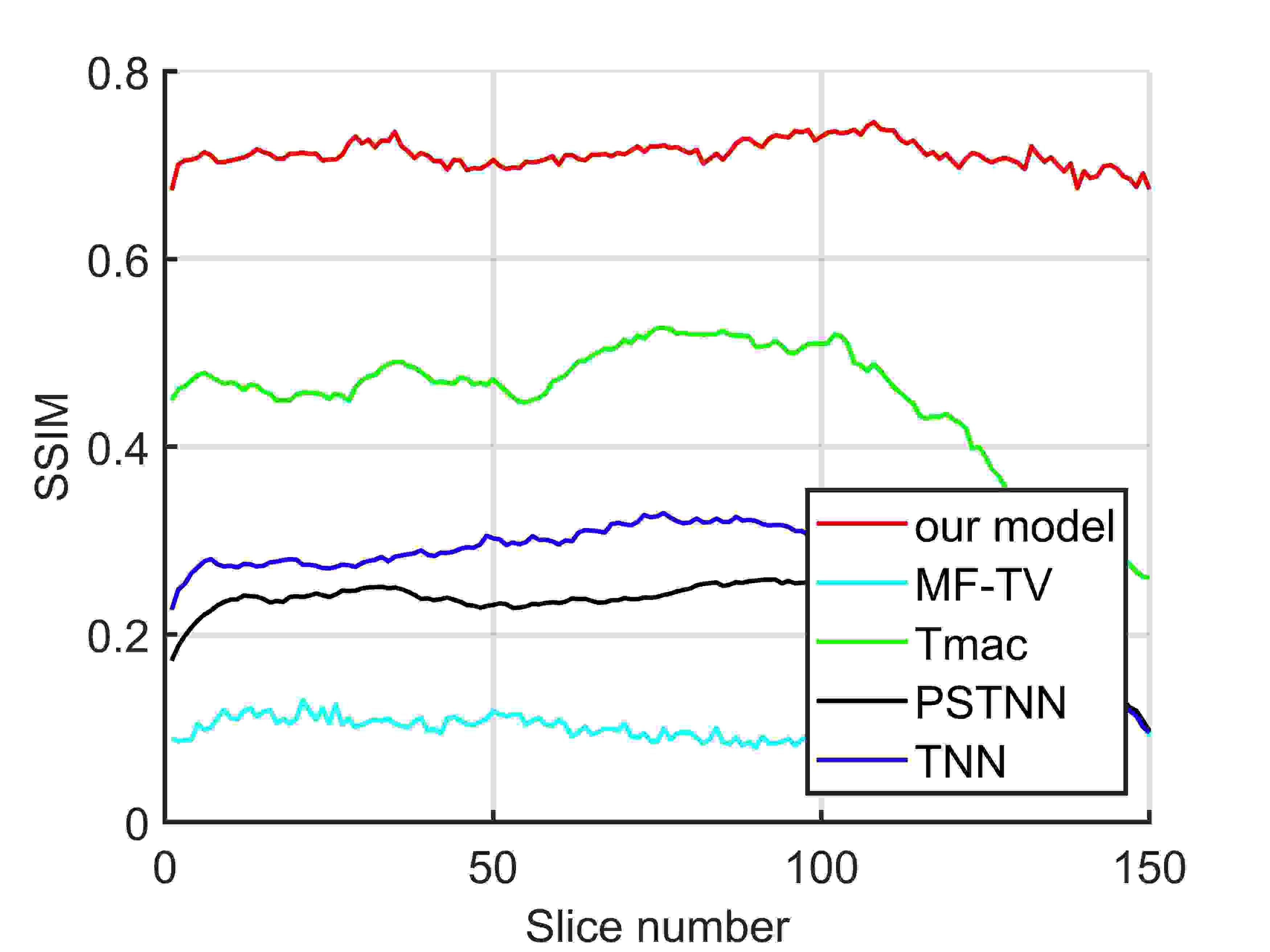}}
	\hfil
	\subfloat[]{\includegraphics[width=0.2\linewidth]{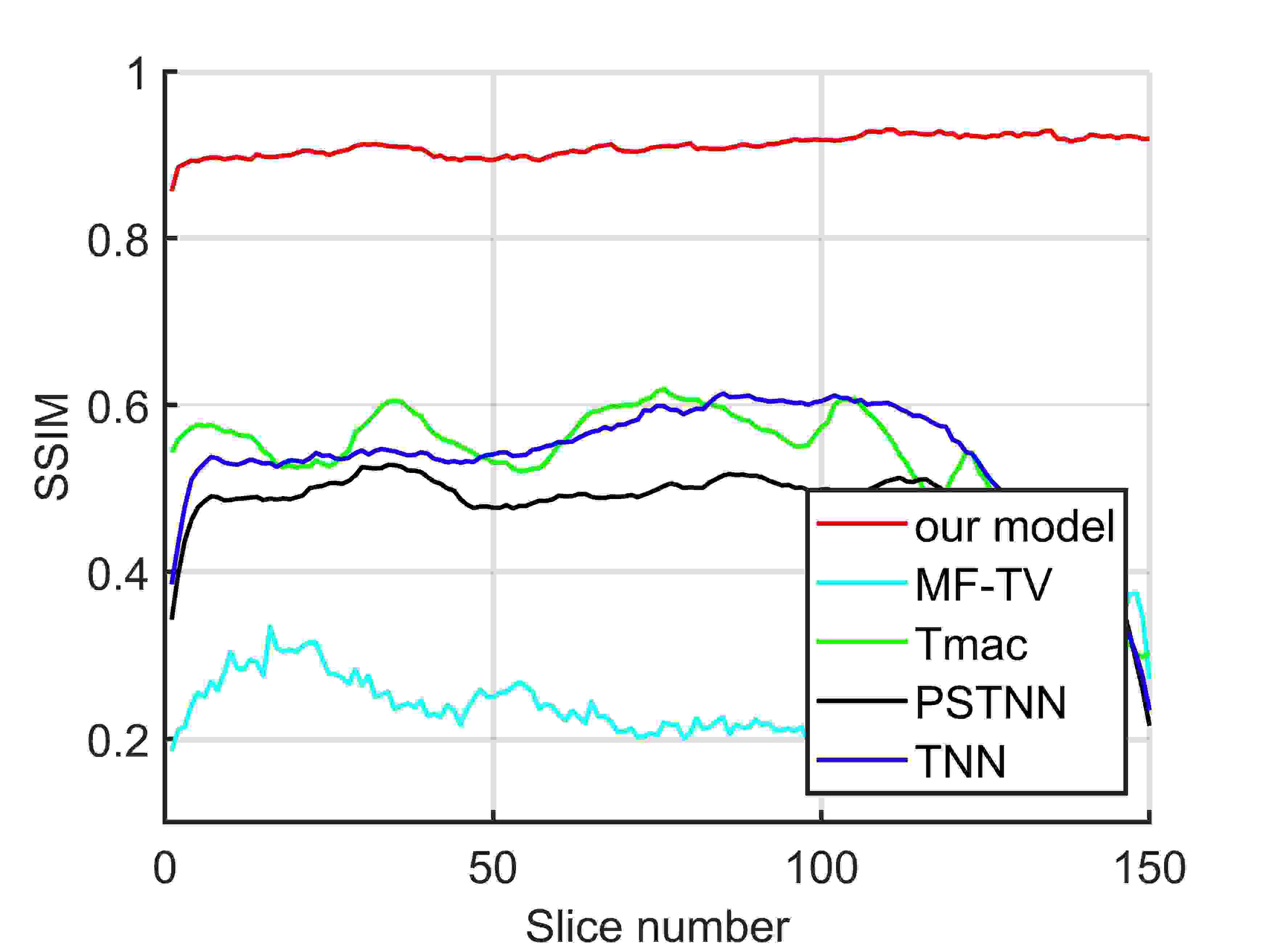}}
	\hfil
	\subfloat[]{\includegraphics[width=0.2\linewidth]{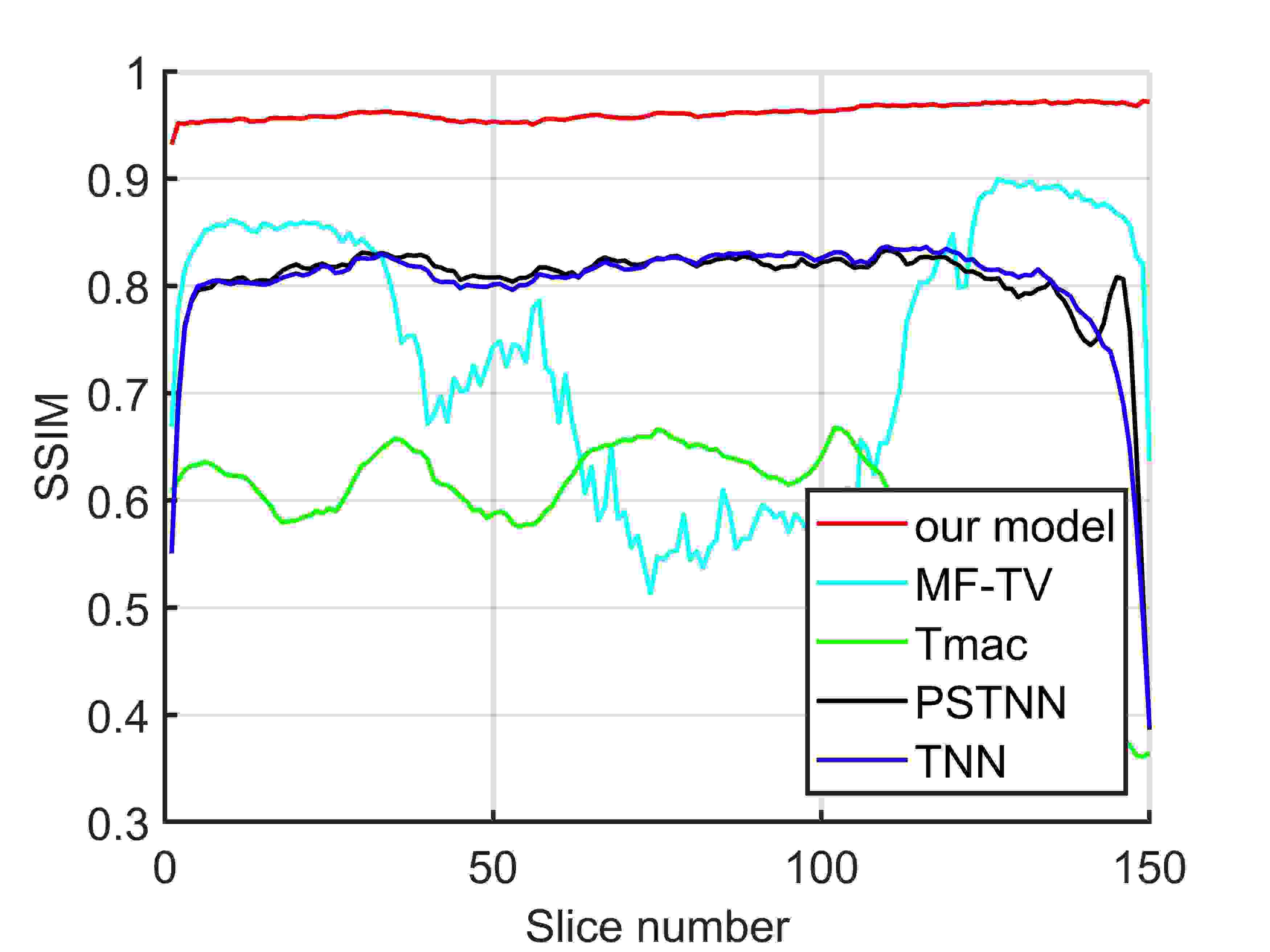}}
	\hfil
	\subfloat[]{\includegraphics[width=0.2\linewidth]{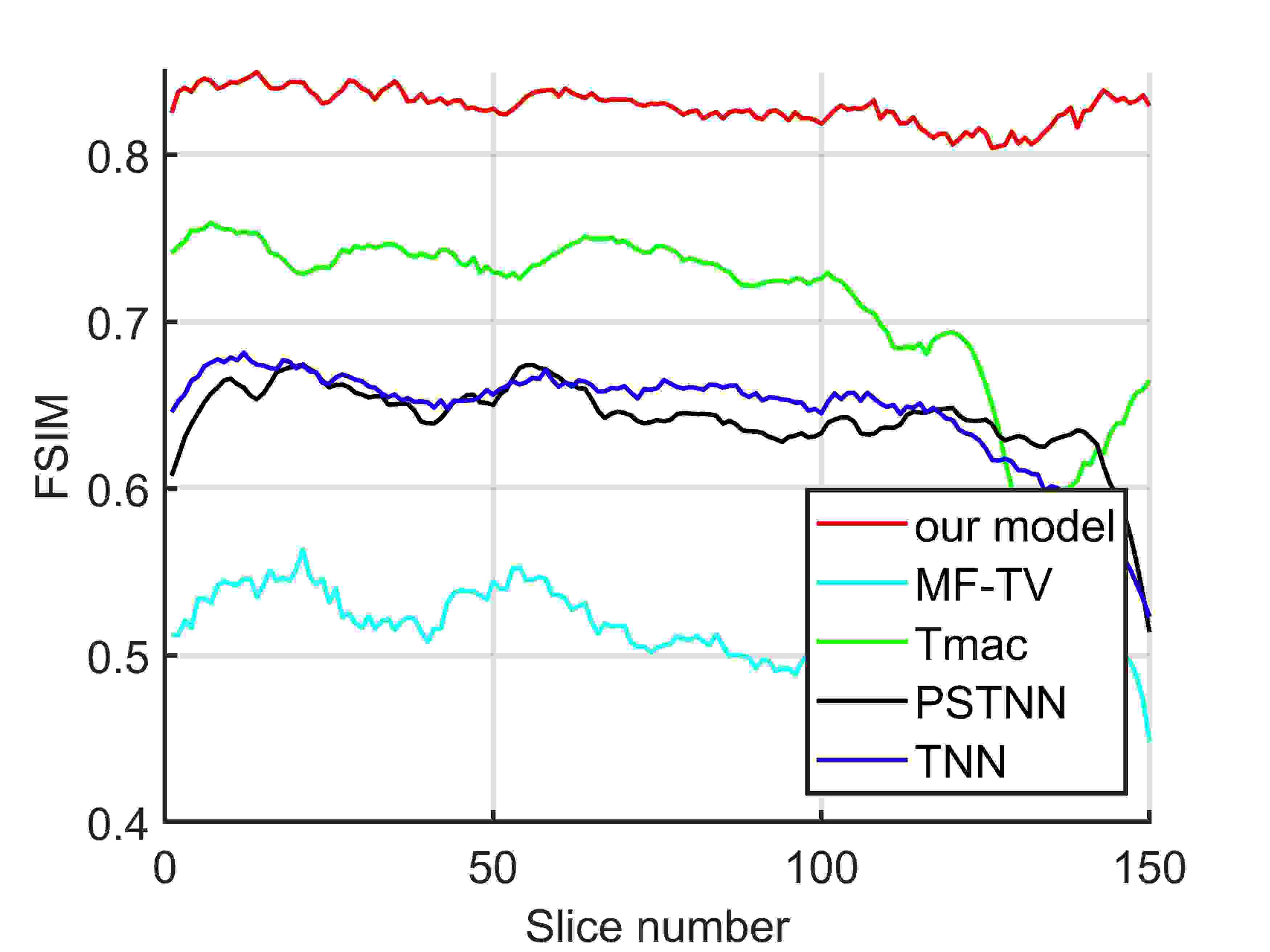}}%
	\hfil					
	\subfloat[]{\includegraphics[width=0.2\linewidth]{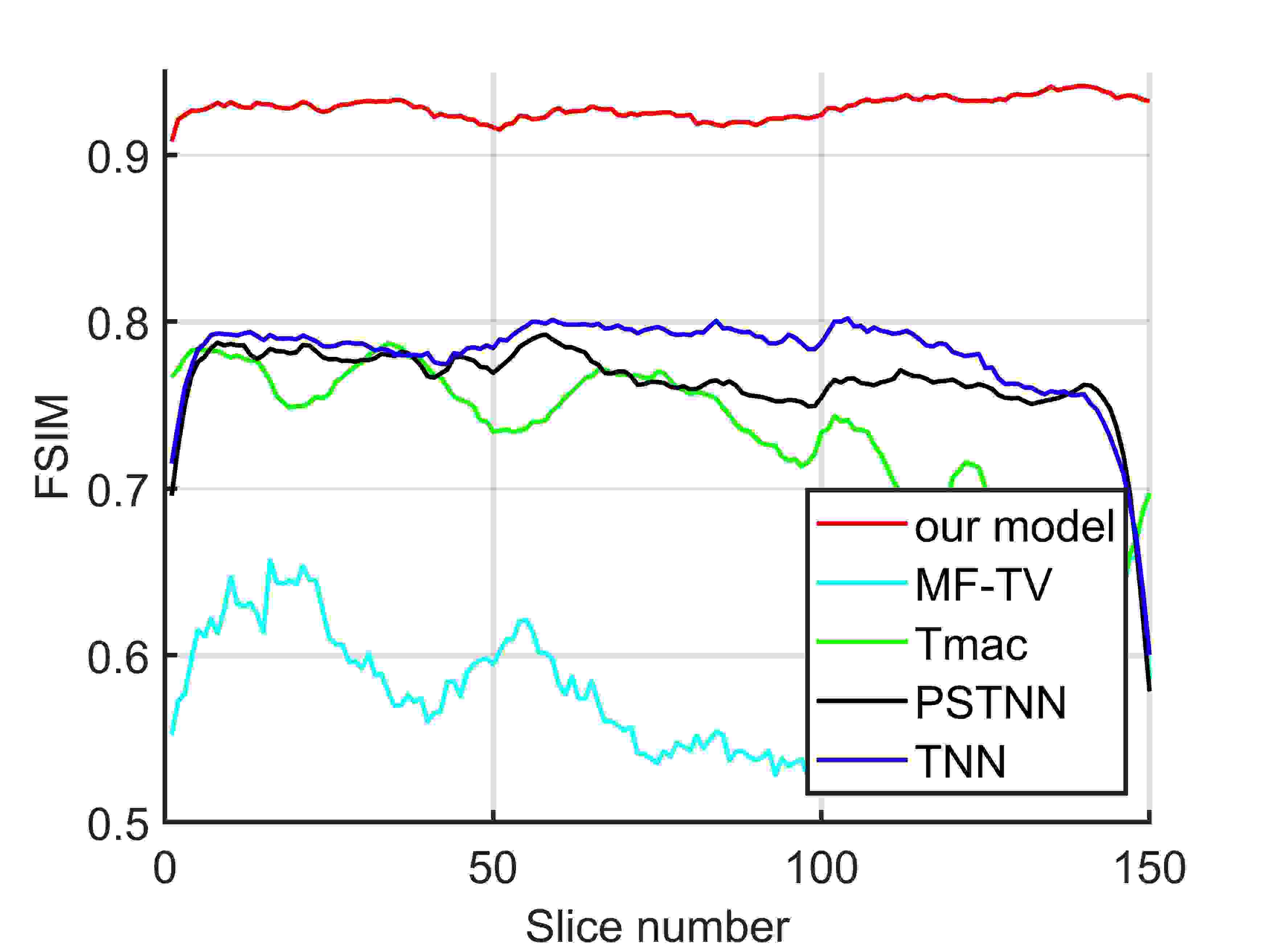}}%
	\hfil
	\subfloat[]{\includegraphics[width=0.2\linewidth]{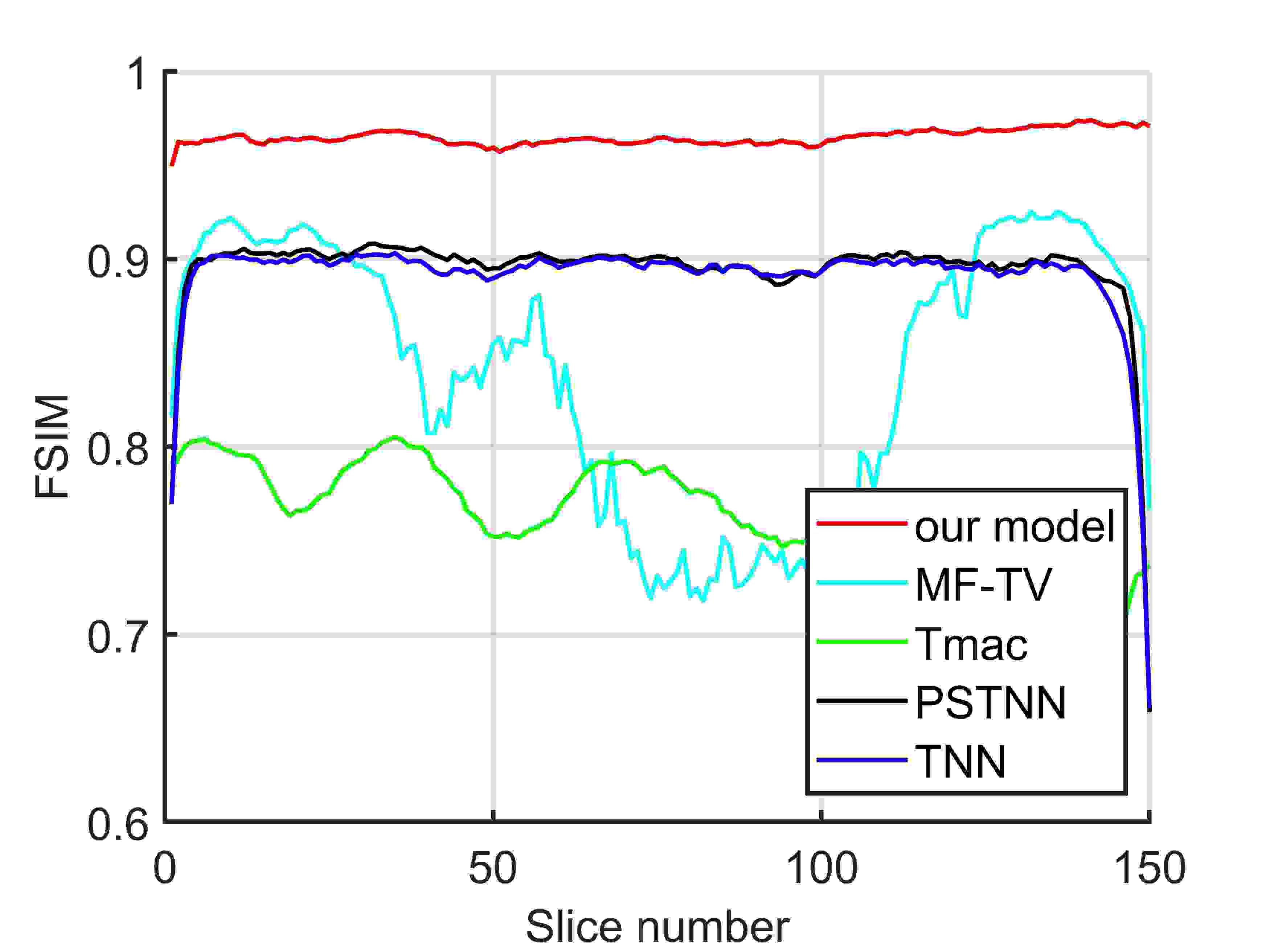}}%
	\caption{The PSNR, SSIM and FSIM of the recovered MRI by MF-TV, Tmac, TNN, PSTNN and our model for all slices, respectively.}
	\label{PSNR and SSIM of MRI}
\end{figure*}

To further verify the versatility of our model for different datasets,
in this subsection we compare our model with MF-TV, Tmac, TNN and PSTNN on MRI dataset, i.e., the cubical MRI data\footnote{http://brainweb.bic.mni.mcgill.ca/brainweb/selection$\_$normal.html}.
Its one slice is shown in Fig. \ref{figure_MR_sr0.1_1}.
The size of the dataset is 181 $\times$ 217 $\times$ 150.
We test all five methods on a series of sampling rates: 5\%, 10\%, 20\% and 30\%.
In addition, the $n$-rank is approximated by using the number of the largest 0.5\% singular values.

For quantitative comparison, Table \ref{table_MRI} reports the PQIs of the results completed by different methods.
The best result for each PQI are marked in bold.
From Table \ref{table_MRI}, it can be found that our method obtains the highest indices among the five tested methods in all SR cases.
Further, the same advantage of our model can also be seen in Fig. \ref{PSNR and SSIM of MRI} which reports the PSNR, SSIM and FSIM of each slice.

For visual comparison,
Fig. \ref{figure_MR_sr0.1_1},
Fig. \ref{figure_MR_sr0.1_2},
Fig. \ref{figure_MR_sr0.1_3} and
Fig. \ref{figure_MR_sr0.1_4} show slices with 90\% missing values and the corresponding completed slices by all the tested methods.
From the results, we see again that our model can better retain the local details and texture information of the images,
and effectively restore the main structure of the images than other compared methods.
Therefore, one can see that the recovered data obtained by our model has the best visual evaluation.

\begin{table}[t]
	\centering
	\caption{The averaged PSNR, SSIM, FSIM, ERGA and SAM of the recovered results on MRI by Tmac, MF-TV, TNN, PSTNN and our model with different sampling rates. The best values are highlighted in bolder fonts.}
	\label{table_MRI}
	 \setlength{\tabcolsep}{1mm}{
	\begin{tabular}{ccccccccc}
		\hline \hline
		&&&SR =0.05 &&&&	\\
		PQI&	nosiy&	our model&	MF-TV&	TMac&	PSTNN&	TNN\\
		PSNR	&	10.258	&	\textbf{26.414}		&	12.332	&	20.51	&	15.859	&	18.218	\\
		SSIM	&	0.228	&	\textbf{0.722}		&	0.099	&	0.45	&	0.224	&	0.27	\\
		FSIM	&	0.473	&	\textbf{0.834}		&	0.52	&	0.711	&	0.642	&	0.646	\\
		ERGA	&	1030.203	&	\textbf{184.279}		&	814.747	&	339.385	&	545.77	&	434.774	\\
		MSAM	&	76.54	&	\textbf{20.411}		&	55.603	&	31.367	&	36.355	&	31.11	\\
		\hline
		&&&SR = 0.1&&&&\\				
		PQI	&	nosiy	&	our model	&	MF-TV	&	TMac	&	PSTNN	&	TNN	\\
		PSNR	&	10.492	&	\textbf{32.652}		&	15.406	&	21.411	&	22.061	&	22.535	\\
		SSIM	&	0.241	&	\textbf{0.912}		&	0.25	&	0.531	&	0.482	&	0.536	\\
		FSIM	&	0.511	&	\textbf{0.926}		&	0.587	&	0.732	&	0.764	&	0.78	\\
		ERGA	&	1002.8	&	\textbf{89.116}		&	584.827	&	308.655	&	275.473	&	266.753	\\
		MSAM	&	70.986	&	\textbf{14.637}		&	41.826	&	29.345	&	24.585	&	24.6	\\
		\hline
		&&&SR = 0.2&&&&\\
		PQI	&	nosiy	&	our model	&	MF-TV	&	TMac	&	PSTNN	&	TNN	\\
		PSNR	&	11.003	&	\textbf{36.529}	&	27.062	&	22.33	&	29.152	&	28.571	\\
		SSIM	&	0.271	&	\textbf{0.962}		&	0.737	&	0.586	&	0.804	&	0.802	\\
		FSIM	&	0.564	&	\textbf{0.963}		&	0.84	&	0.754	&	0.895	&	0.891	\\
		ERGA	&	945.583	&	\textbf{57.037}		&	173.636	&	276.269	&	127.133	&	136.182	\\
		MSAM	&	62.887	&	\textbf{11.559}		&	21.792	&	27.267	&	17.513	&	17.855	\\
		\hline \hline
	\end{tabular}}
\end{table}

\subsection{Hyperspectral Image Data}

\begin{figure*}[!t]	
	\centering			
	\subfloat[Original]{\includegraphics[width=0.13\linewidth]{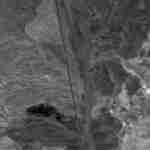}}%
	\hfil	
	\subfloat[97.5\% Masked]{\includegraphics[width=0.13\linewidth]{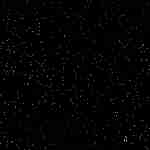}}%
	\hfil					
	\subfloat[our model]{\includegraphics[width=0.13\linewidth]{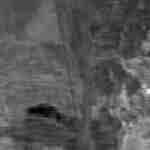}}%
	\hfil
	%	\subfloat[our model-1]{\includegraphics[width=0.23\linewidth]{myfigure/Image_result/Image_HSI_b72_sr0.05/DBNN}}%
	%	\hfil
	\subfloat[MF-TV]{\includegraphics[width=0.13\linewidth]{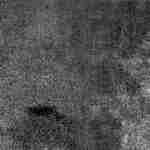}}%
	\hfil					
	\subfloat[Tmac]{\includegraphics[width=0.13\linewidth]{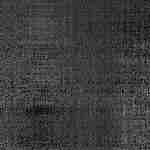}}%
	\hfil
	\subfloat[PSTNN]{\includegraphics[width=0.13\linewidth]{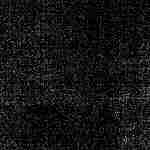}}%
	\hfil
	\subfloat[TNN]{\includegraphics[width=0.13\linewidth]{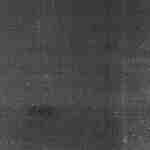}}%	
	\caption{One slice of the recovered HSI "Cuprite" by our model, MF-TV, Tmac, PSTNN and TNN.  The sampling rate is 2.5\%.}
	\label{figure_HSI_sr0.025}
\end{figure*}

\begin{figure*}[!t]	
	\centering			
	\subfloat[Original]{\includegraphics[width=0.13\linewidth]{Image_HSI_b72.png}}%
	\hfil	
	\subfloat[95\% Masked]{\includegraphics[width=0.13\linewidth]{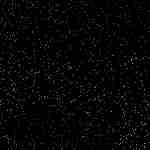}}%
	\hfil					
	\subfloat[our model]{\includegraphics[width=0.13\linewidth]{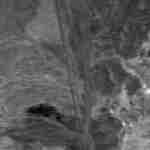}}%
	\hfil
	%	\subfloat[our model-1]{\includegraphics[width=0.23\linewidth]{myfigure/Image_result/Image_HSI_b72_sr0.05/DBNN}}%
	%	\hfil
	\subfloat[MF-TV]{\includegraphics[width=0.13\linewidth]{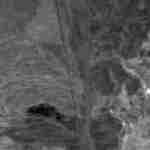}}%
	\hfil					
	\subfloat[Tmac]{\includegraphics[width=0.13\linewidth]{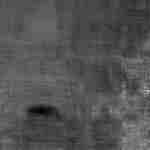}}%
	\hfil
	\subfloat[PSTNN]{\includegraphics[width=0.13\linewidth]{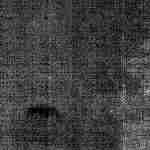}}%
	\hfil
	\subfloat[TNN]{\includegraphics[width=0.13\linewidth]{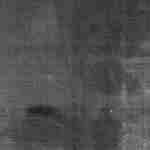}}%	
	\caption{One slice of the recovered HSI "Cuprite" by our model, MF-TV, Tmac, PSTNN and TNN.  The sampling rate is 5\%.}
	\label{figure_HSI_sr0.05}
\end{figure*}

\begin{figure*}[!t]	
	\centering
	\subfloat[PSNR]{\includegraphics[width=0.15\linewidth]{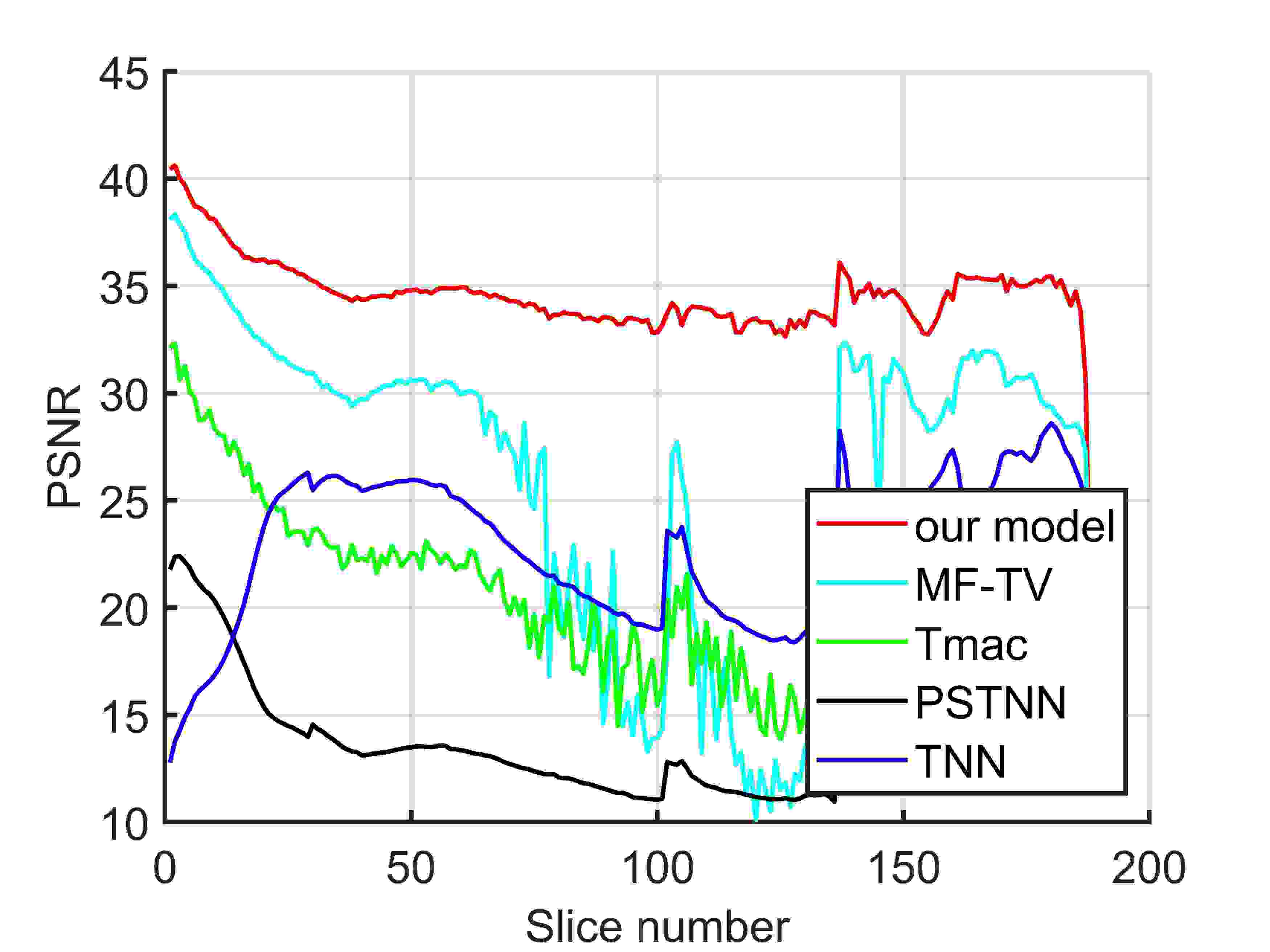}}%
	\hfil					
	\subfloat[SSIM]{\includegraphics[width=0.15\linewidth]{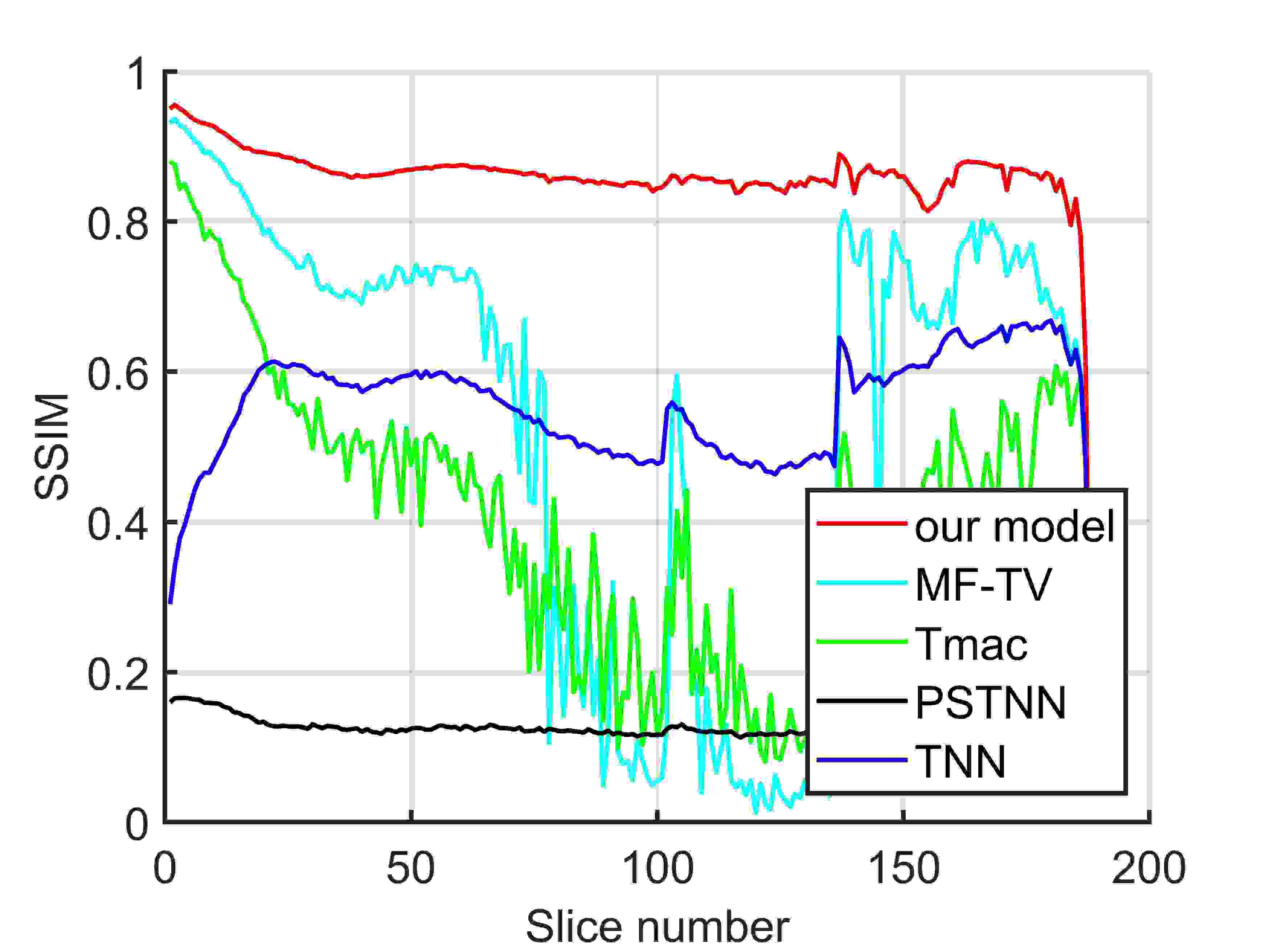}}%
	\hfil
	\subfloat[FSIM]{\includegraphics[width=0.15\linewidth]{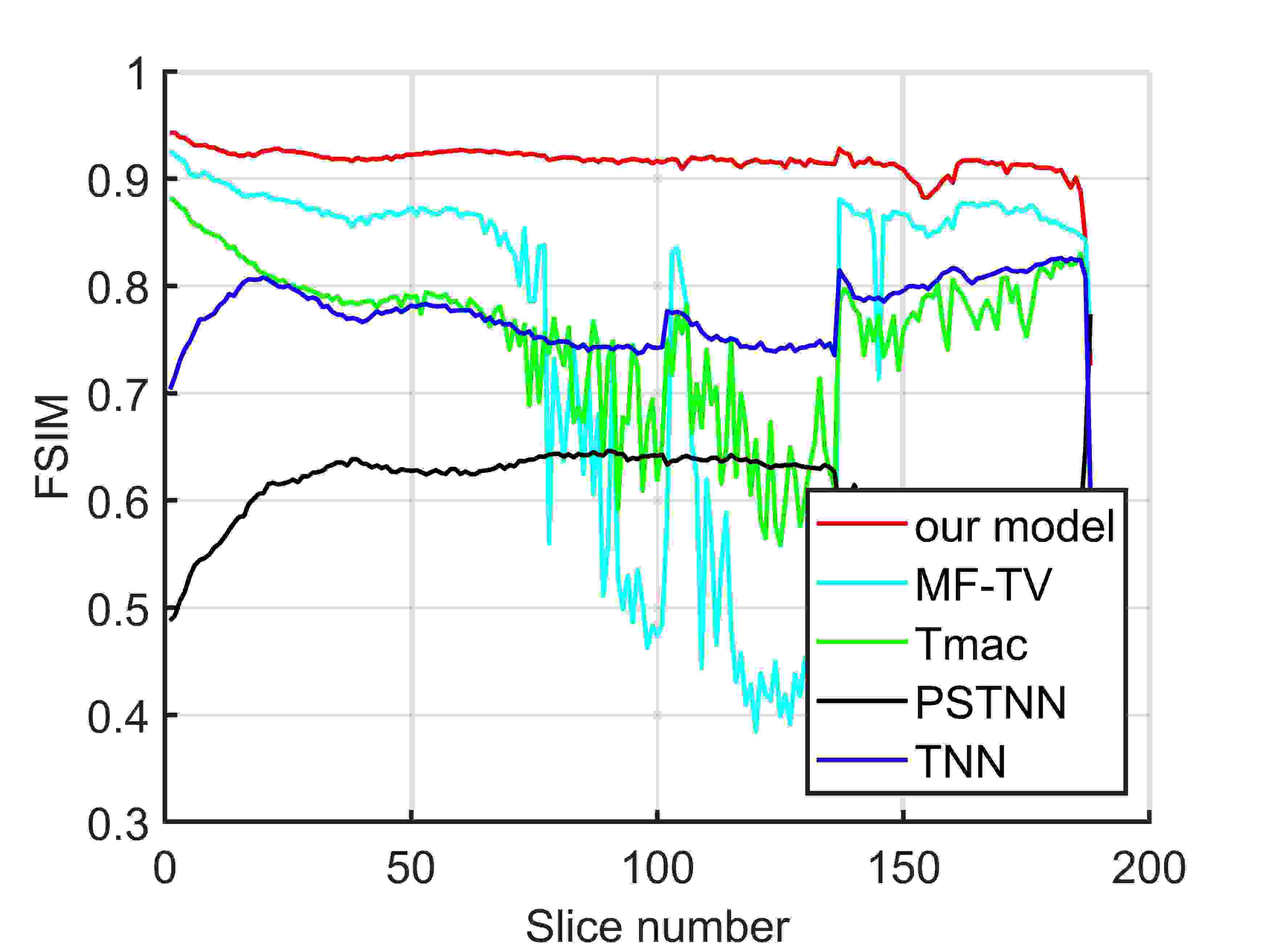}}%
	\hfil	
	\subfloat[PSNR]{\includegraphics[width=0.15\linewidth]{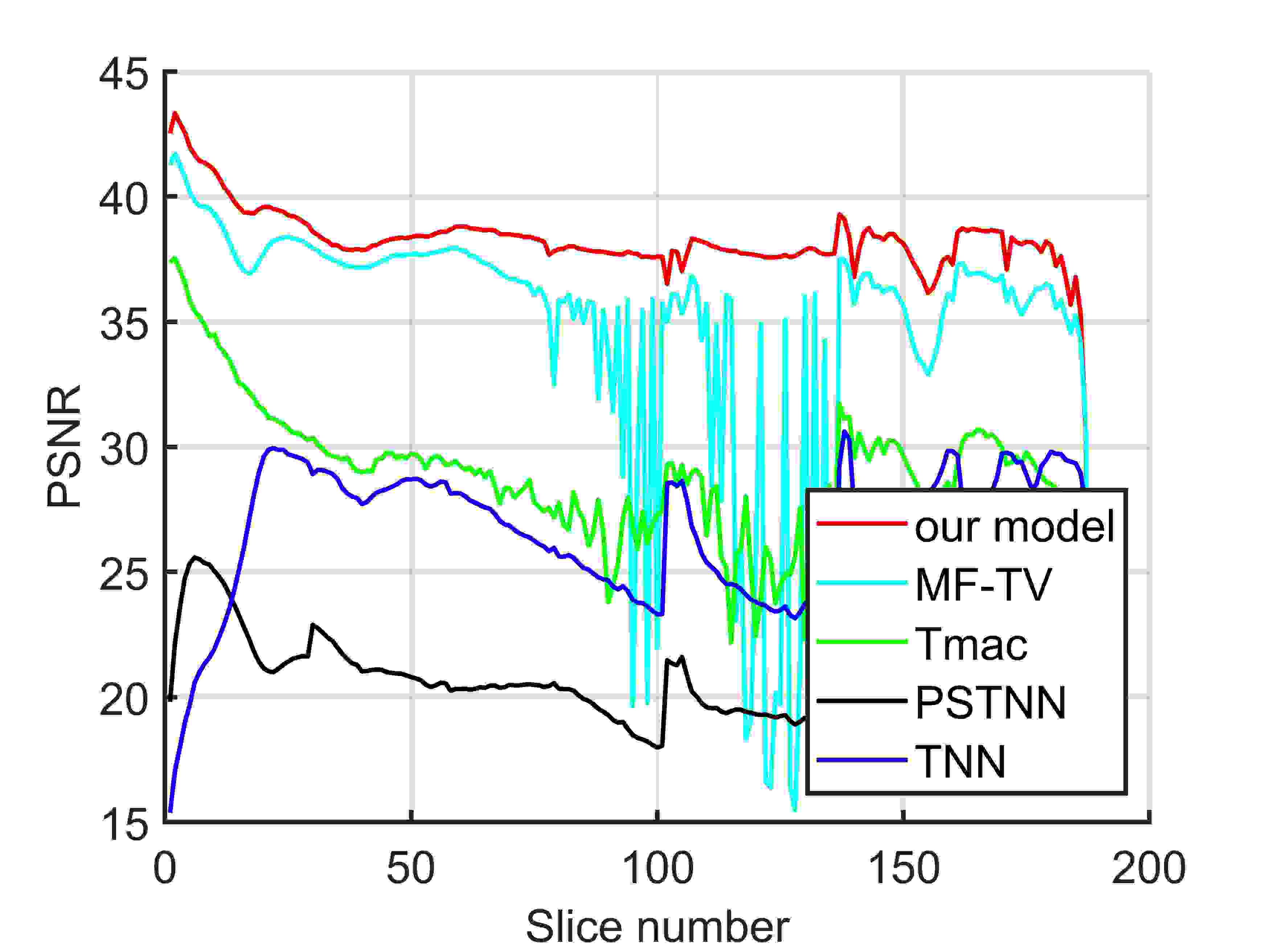}}%
	\hfil					
	\subfloat[SSIM]{\includegraphics[width=0.15\linewidth]{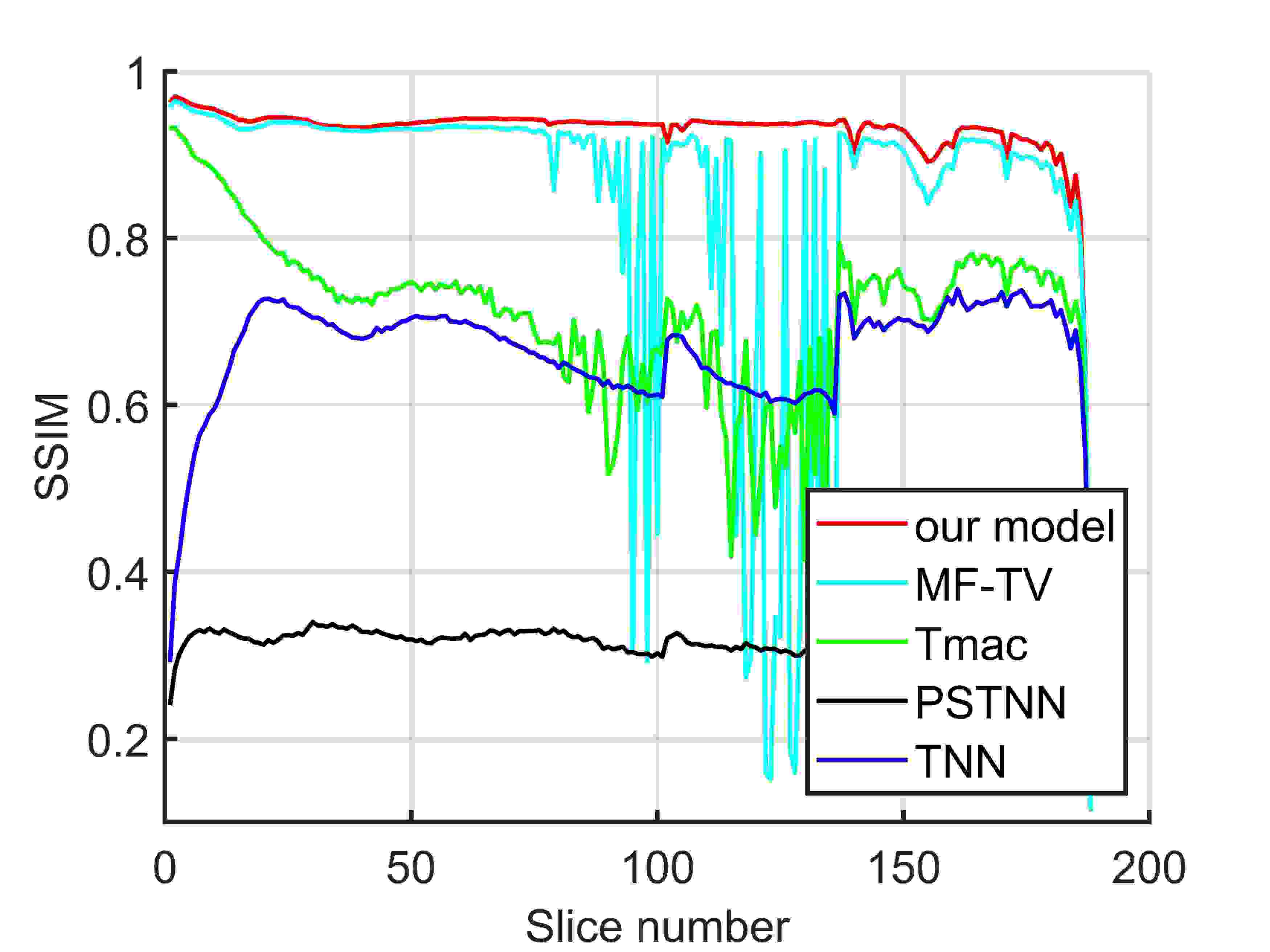}}%
	\hfil
	\subfloat[FSIM]{\includegraphics[width=0.15\linewidth]{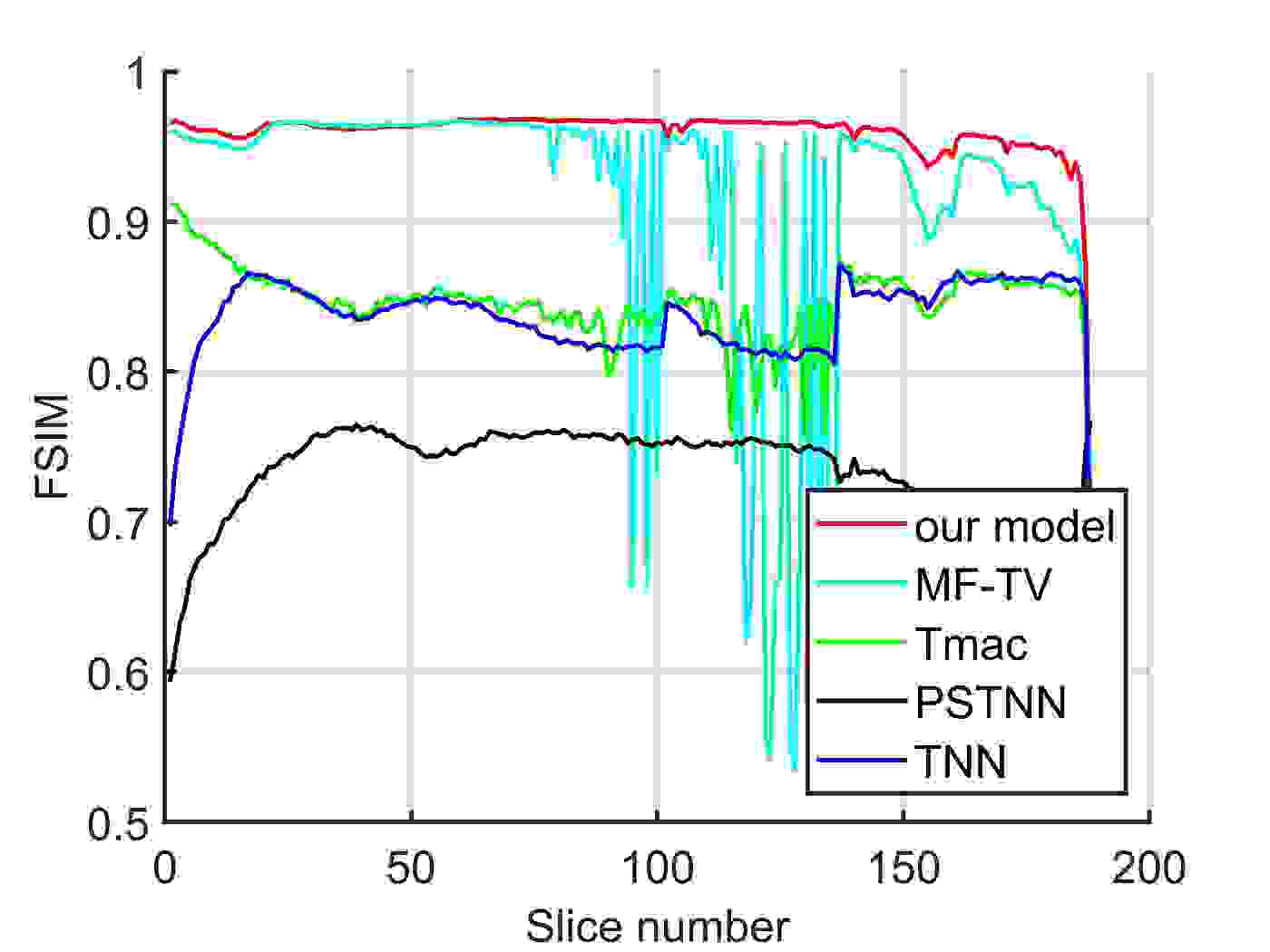}}%
	\caption{The PSNR, SSIM and FSIM of the recovered HSI "Cuprite" by MF-TV, Tmac, TNN, PSTNN and our model for all slices, respectively.(a)-(c): 97.5\% entries missing, (d)-(f): 95\% entries missing.}
	\label{PSNR and SSIM of HSI}
\end{figure*}

In this section, we compare our model with MF-TV, Tmac, TNN and PSTNN on one HSI dataset: Airborne Visible/Infrared Imaging Spectrometer (AVIRIS) Cuprite data\footnote{http://aviris.jpl.nasa.gov/html/aviris.freedata.html}.
The size of AVIRIS Cuprite data is 150 $\times$ 150 $\times$ 188.
Its one slice is shown in Fig. \ref{figure_HSI_sr0.05}.
We test all five methods on a series of sampling rates: 2.5\%, 5\% and 10\%.
In addition, the $n$-rank is approximated by using the number of the largest 0.3\% singular values.

For quantitative comparison, Table \ref{table_HSI} reports the
average PQIs of each tested method with three different sampling rates.
At sampling rates 0.025 and 0.05, Fig. \ref{PSNR and SSIM of HSI} reports the PSNR, SSIM and FSIM of each frontal slice in the completed results.
For visual comparison, Fig. \ref{figure_HSI_sr0.05} and Fig. \ref{figure_HSI_sr0.025} show
slices of the sampled data with 97.5\% and 95\% missing values and the corresponding recovered slices by the tested methods.
From the results,
we see again that our model not only obtains the highest PQIs,
but also recovers more structure information and spatial details of the images than compared methods, especially at low sampling rates.

\begin{table}
	\caption{The averaged PSNR, SSIM, FSIM, ERGA and SAM of the recovered results on hyperspectral image "Cuprite" by Tmac, MF-TV, TNN, PSTNN and our model with different sampling rates. The best values are highlighted in bolder fonts.}
	\centering
	\label{table_HSI}
	\setlength{\tabcolsep}{1mm}{
	\begin{tabular}{ccccccccc}
		\hline \hline
		&&&SR =0.025&&&&	\\
		PQI&	nosiy&	our model&	MF-TV&	TMac&	PSTNN&	TNN\\
		PSNR	&	7.666	&	\textbf{34.595}		&	26.115	&	21.25	&	13.387	&	22.783	\\
		SSIM	&	0.007	&	\textbf{0.861}		&	0.539	&	0.412	&	0.124	&	0.554	\\
		FSIM	&	0.48	&	\textbf{0.916}		&	0.765	&	0.755	&	0.613	&	0.775	\\
		ERGA	&	1043.633	&	\textbf{50.383}	&	237.074	&	235.594	&	539.574	&	245.333	\\
		MSAM	&	81.221	&	\textbf{1.662}	&	12.913	&	7.842	&	17.98	&	9.156	\\	
		\hline
		&&&SR = 0.05&&&&\\	
		PQI&	nosiy&	our model&	MF-TV&	TMac&	PSTNN&	TNN\\		
		PSNR	&	7.779	&	\textbf{38.202}	&	34.684	&	28.945	&	20.621	&	26.579	\\
		SSIM	&	0.01	&	\textbf{0.928}	&	0.845	&	0.712	&	0.31	&	0.663	\\
		FSIM	&	0.471	&	\textbf{0.960}	&	0.915	&	0.846	&	0.735	&	0.836	\\
		ERGA	&	1030.139	&	\textbf{41.898}	&	89.372	&	93.352	&	234.445	&	154.292	\\
		MSAM	&	77.268	&	\textbf{1.559}	&	4.386	&	3.278	&	7.886	&	5.413	\\
		\hline
		&&&SR = 0.1&&&&\\	
		PQI	&	nosiy	&	our model	&	MF-TV	&	TMac	&	PSTNN	&	TNN	\\
		PSNR	&	8.013	&39.056	&	\textbf{40.888}	&	35.627	&	35.51	&	35.015	\\
		SSIM	&	0.014	&0.939	&	\textbf{0.957}	&	0.885	&	0.907	&	0.897	\\
		FSIM	&	0.451	&0.966	&	\textbf{0.978}	&	0.931	&	0.951	&	0.943	\\
		ERGA	&	1002.75	&34.544	&	\textbf{34.263}	&	44.518	&	54.421	&	57.537	\\
		MSAM	&	71.695	&\textbf{1.299}	&1.46	&	1.445	&	2.072	&	2.192	\\
		\hline \hline
	\end{tabular}}
\end{table}

\begin{figure*}[!t]	
	\centering
	\subfloat[]{\includegraphics[width=0.15\linewidth]{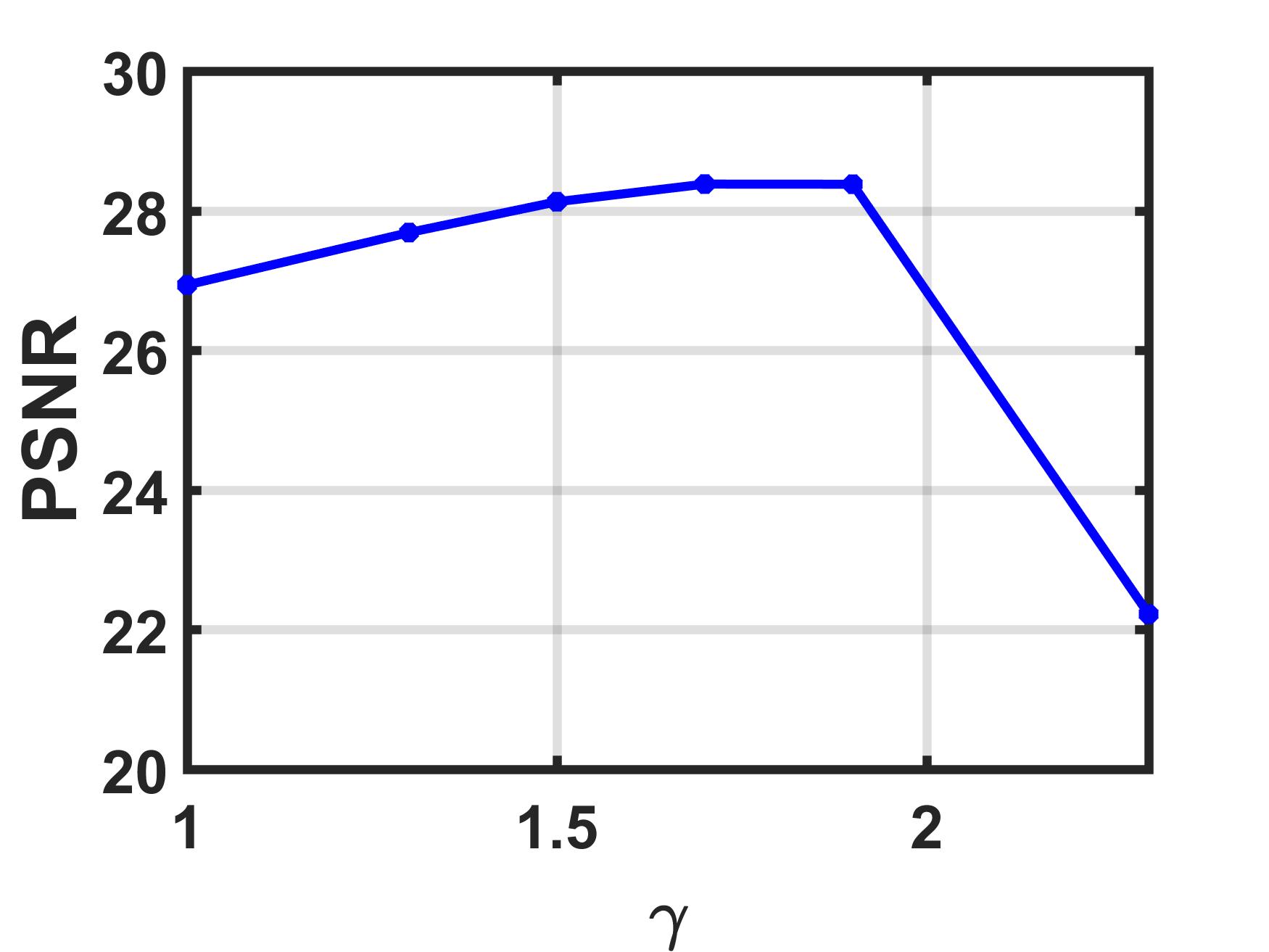}}%
	\hfil					
	\subfloat[]{\includegraphics[width=0.15\linewidth]{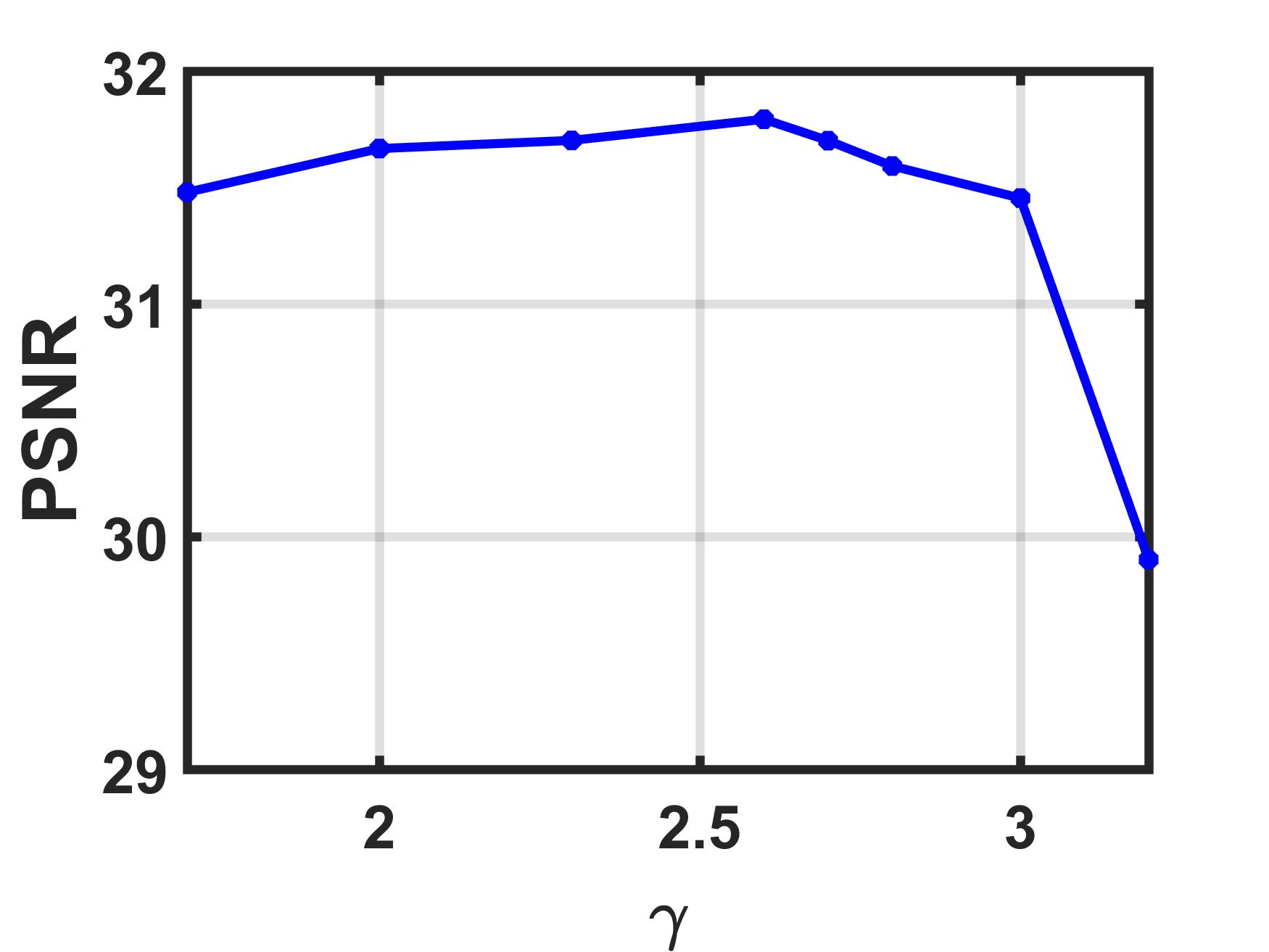}}%
	\hfil
	\subfloat[]{\includegraphics[width=0.15\linewidth]{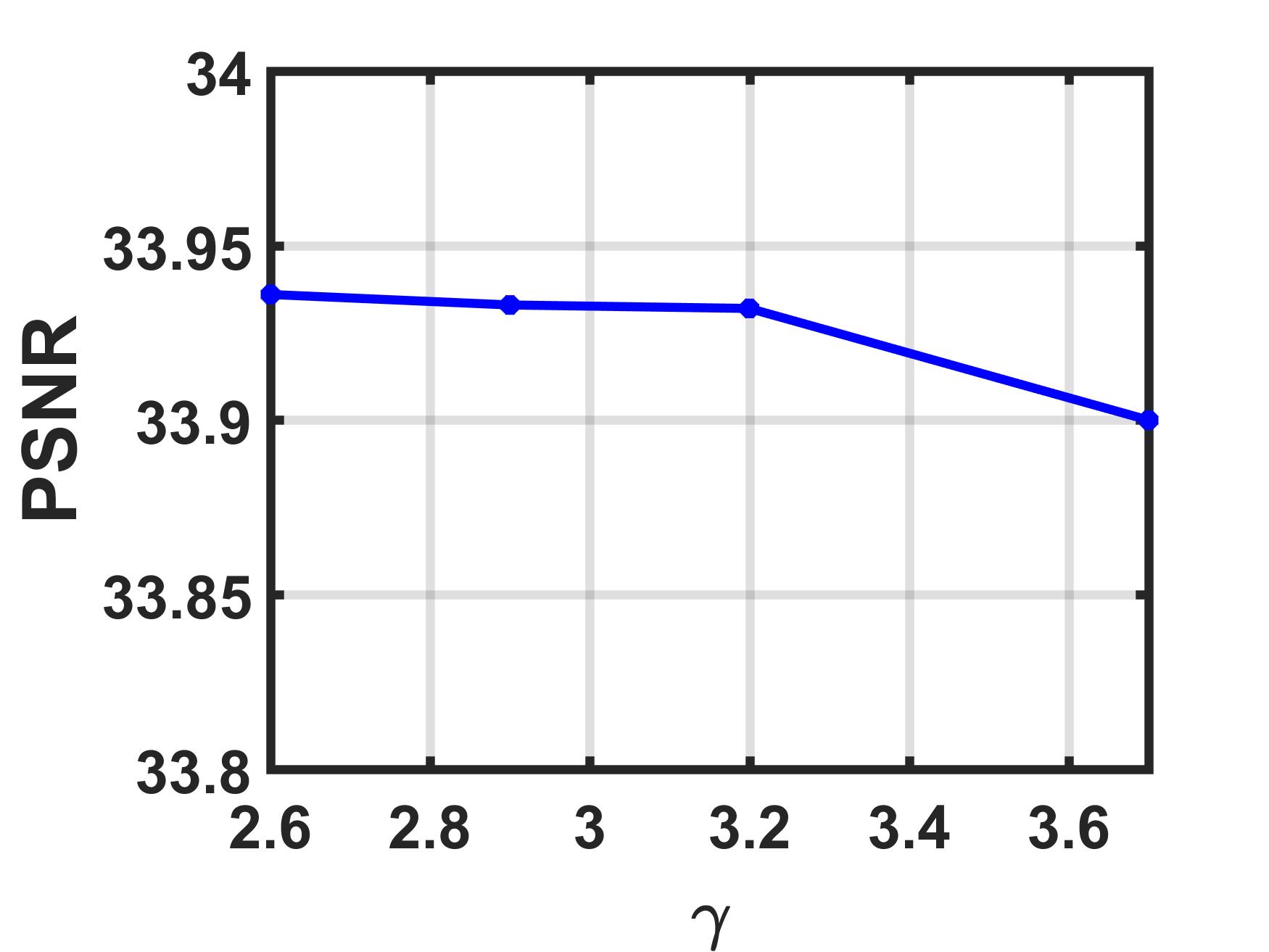}}%
	\hfil	
	\subfloat[]{\includegraphics[width=0.15\linewidth]{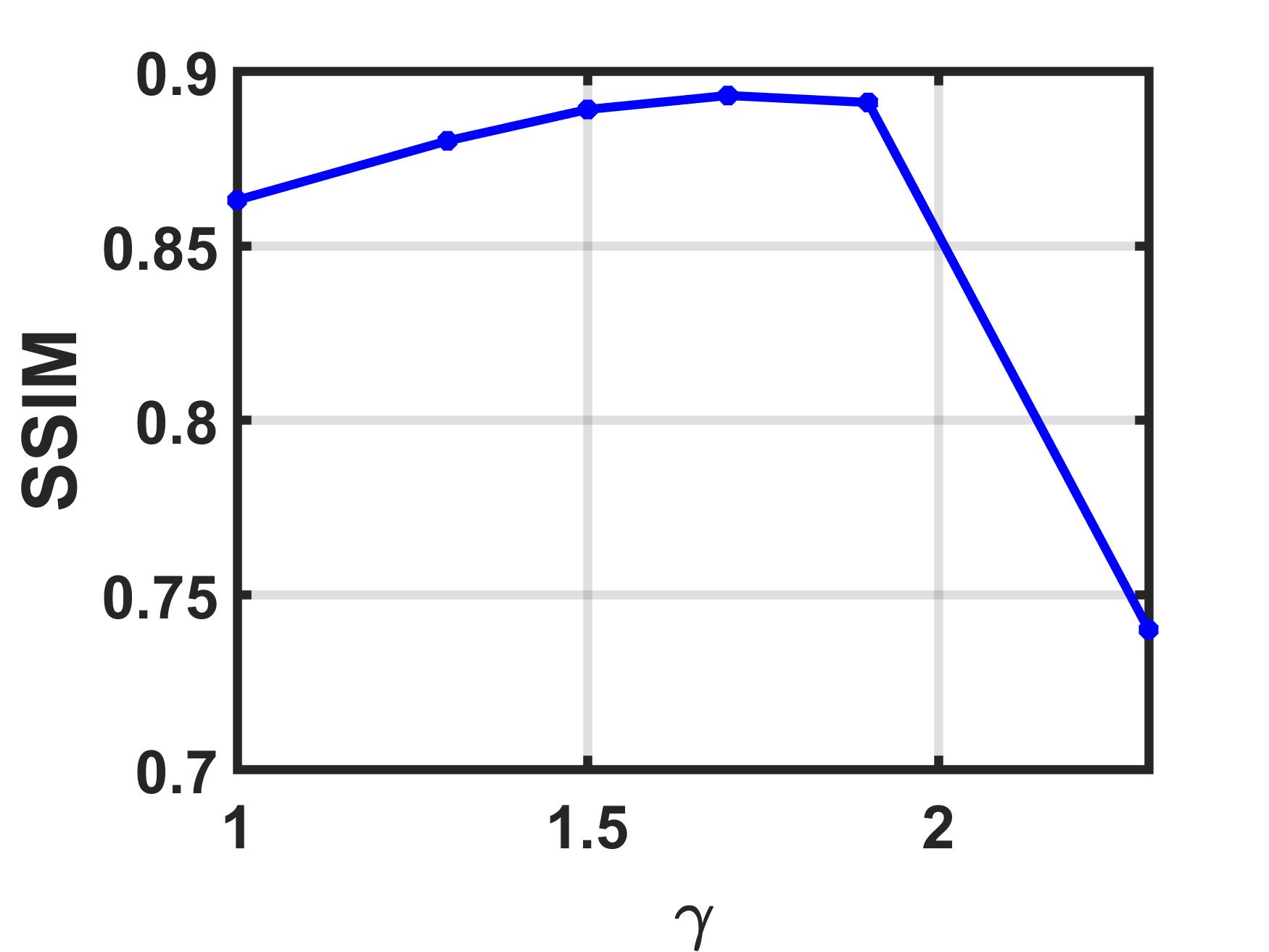}}%
	\hfil					
	\subfloat[]{\includegraphics[width=0.15\linewidth]{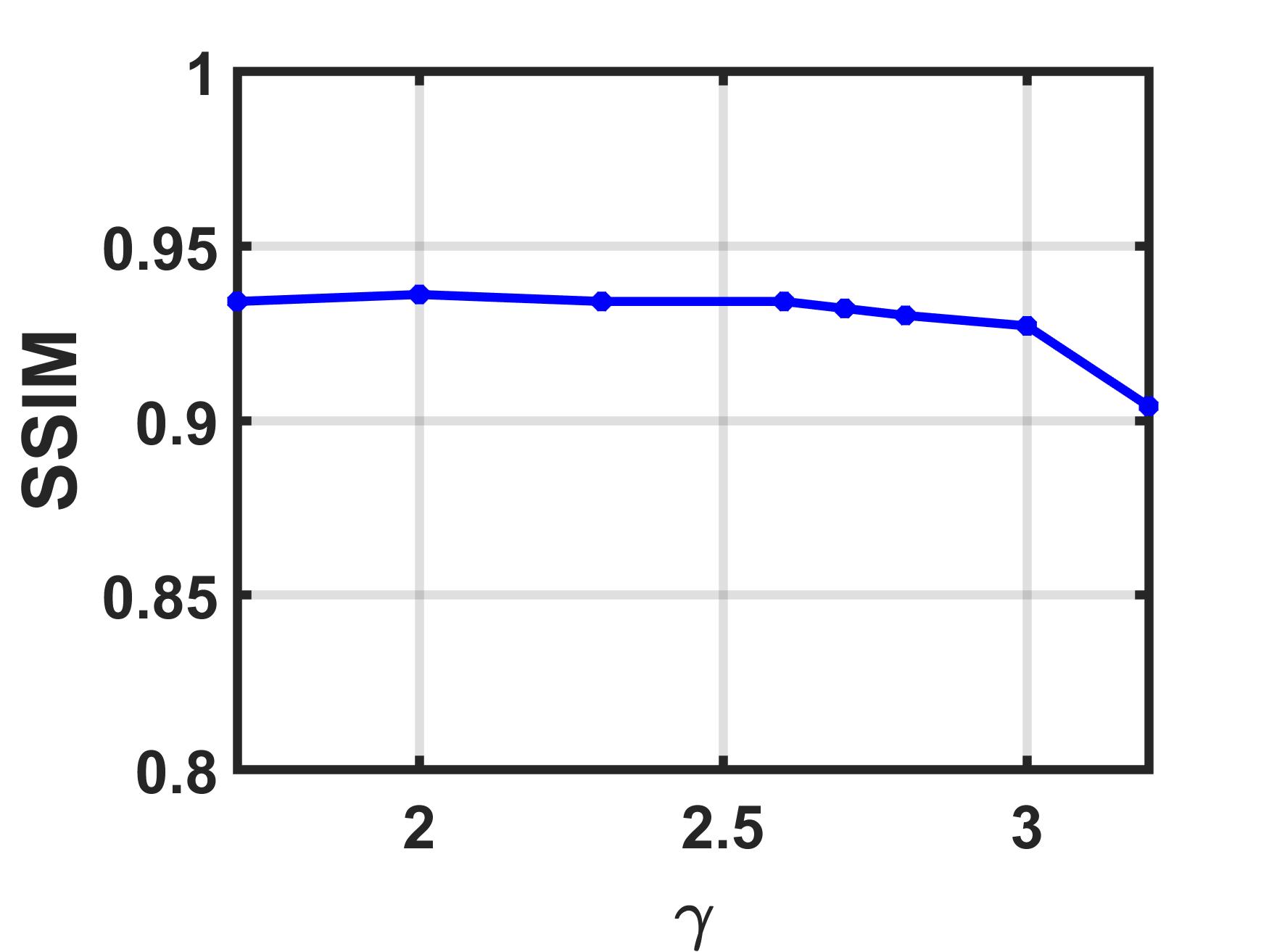}}%
	\hfil
	\subfloat[]{\includegraphics[width=0.15\linewidth]{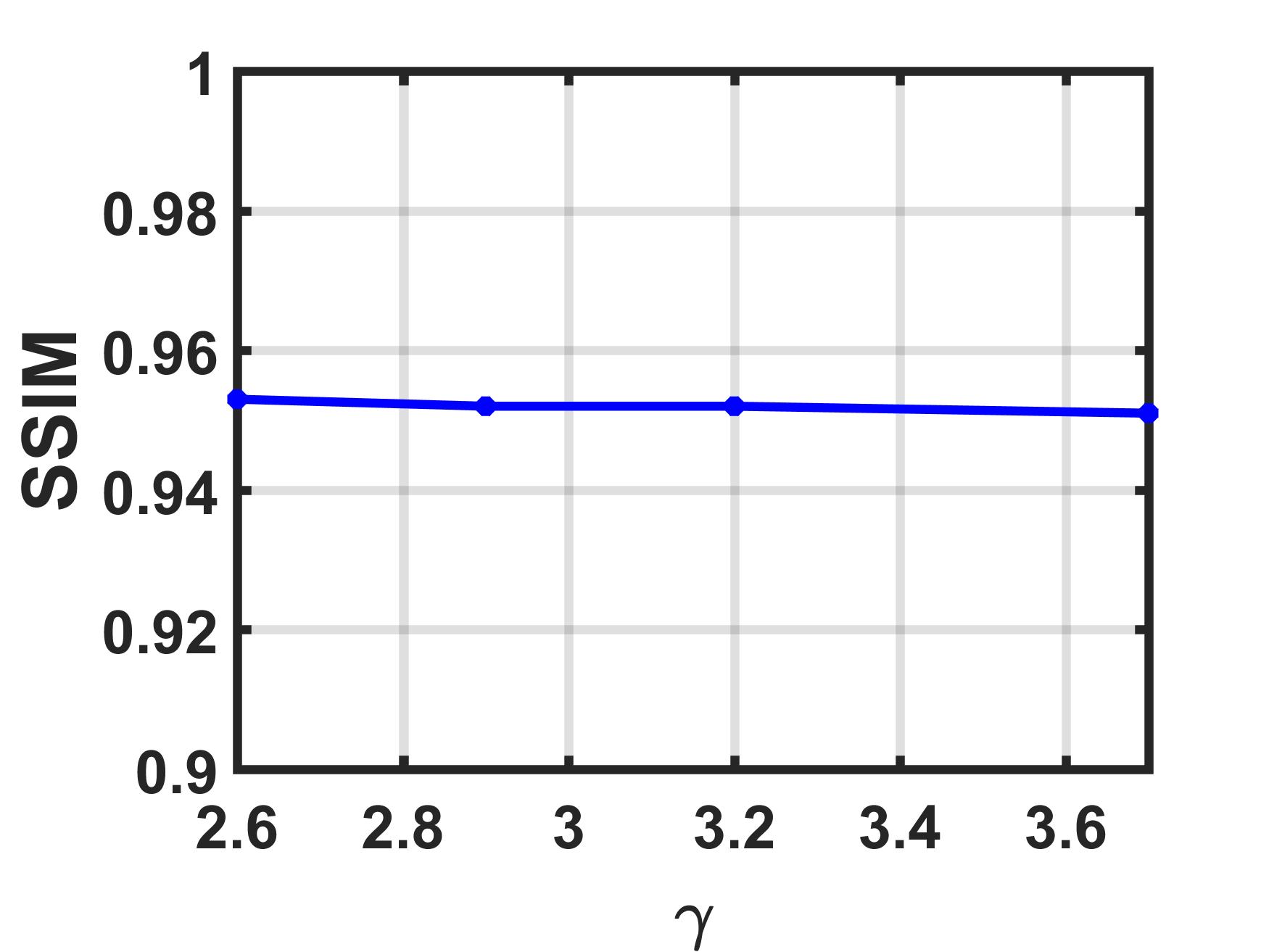}}%
	\caption{Sensitivity analysis of parameter $\gamma$ in video "hall" dataset.
		(a-c) Change in the MPSNR value: SR= 0.05, 0.1 and 0.2. (e-f) Change in the MSSIM value: 0.05, 0.1 and 0.2, respectively.}
	\label{parameter_analysis_hall}
\end{figure*}

\begin{figure*}[!t]	
	\centering
	\subfloat[]{\includegraphics[width=0.15\linewidth]{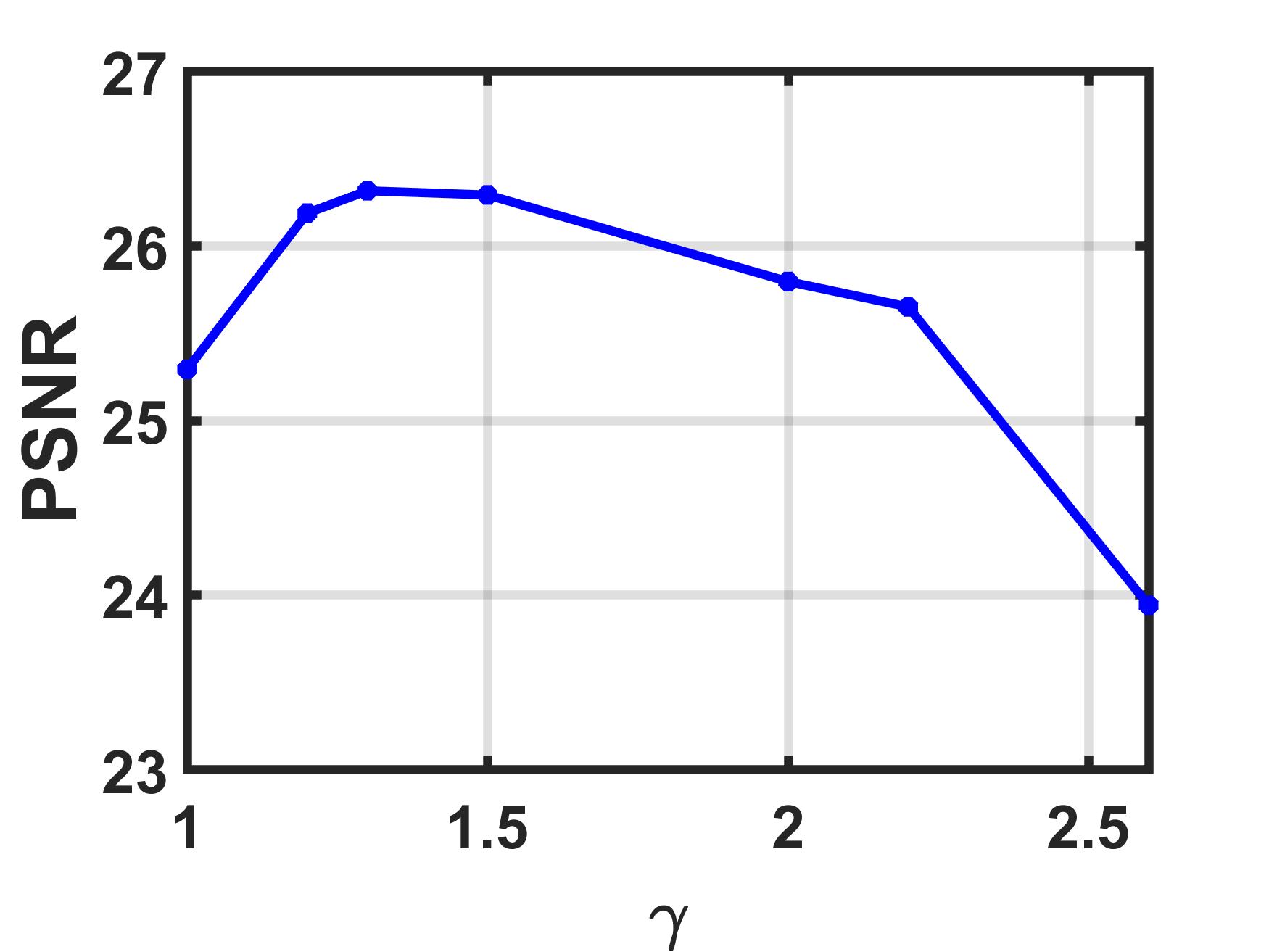}}%
	\hfil					
	\subfloat[]{\includegraphics[width=0.15\linewidth]{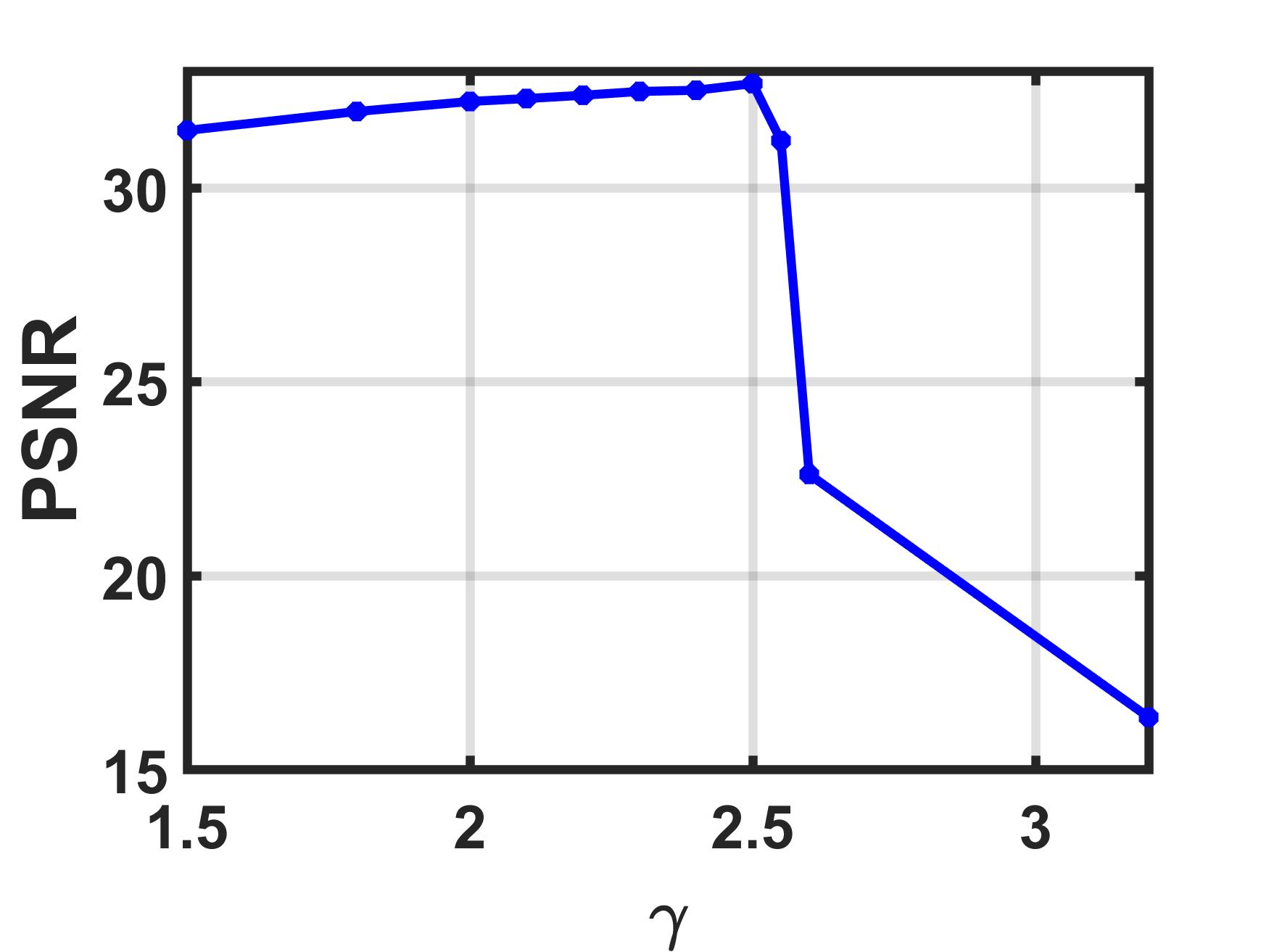}}%
	\hfil
	\subfloat[]{\includegraphics[width=0.15\linewidth]{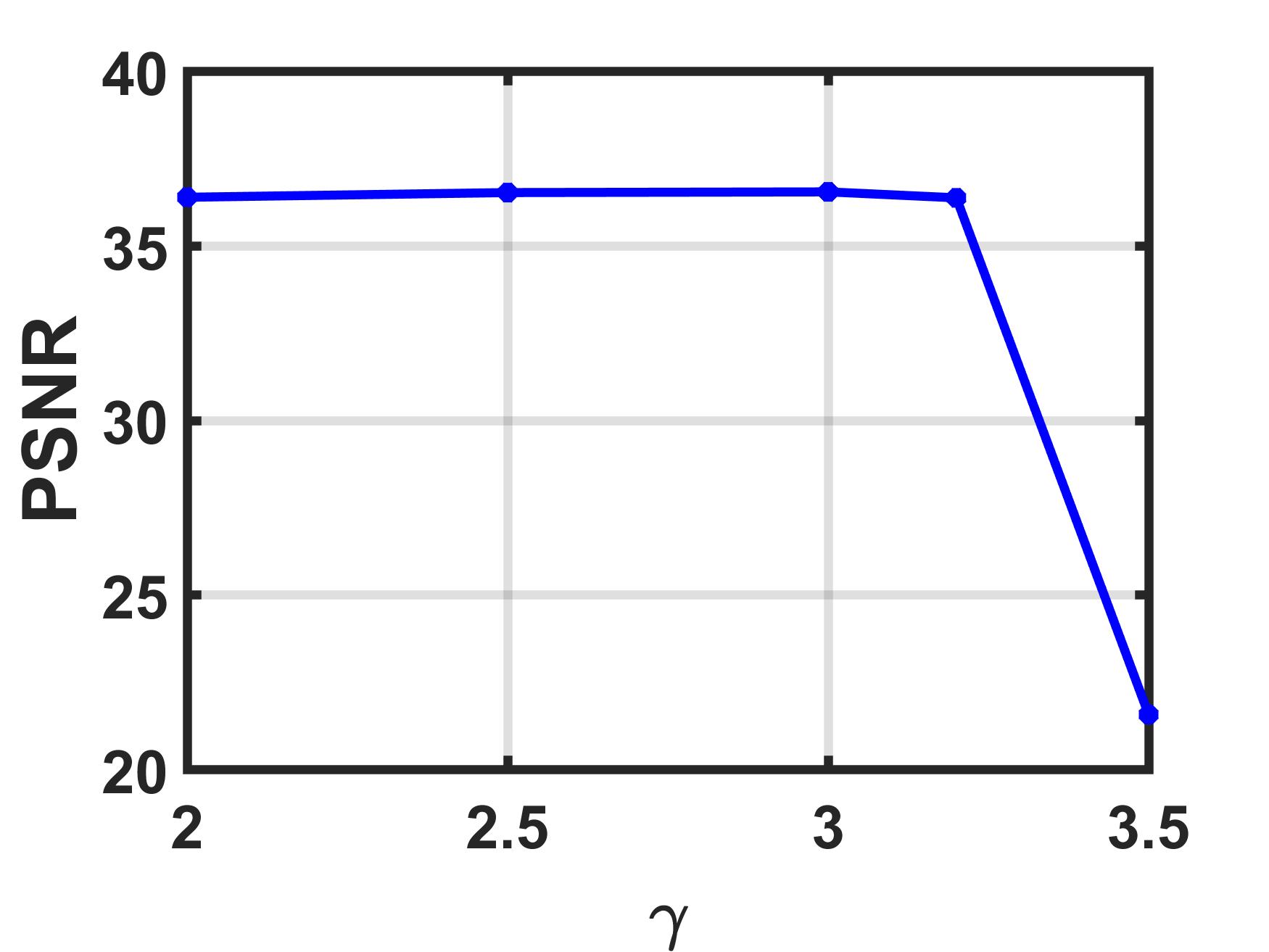}}%
	\hfil	
	\subfloat[]{\includegraphics[width=0.15\linewidth]{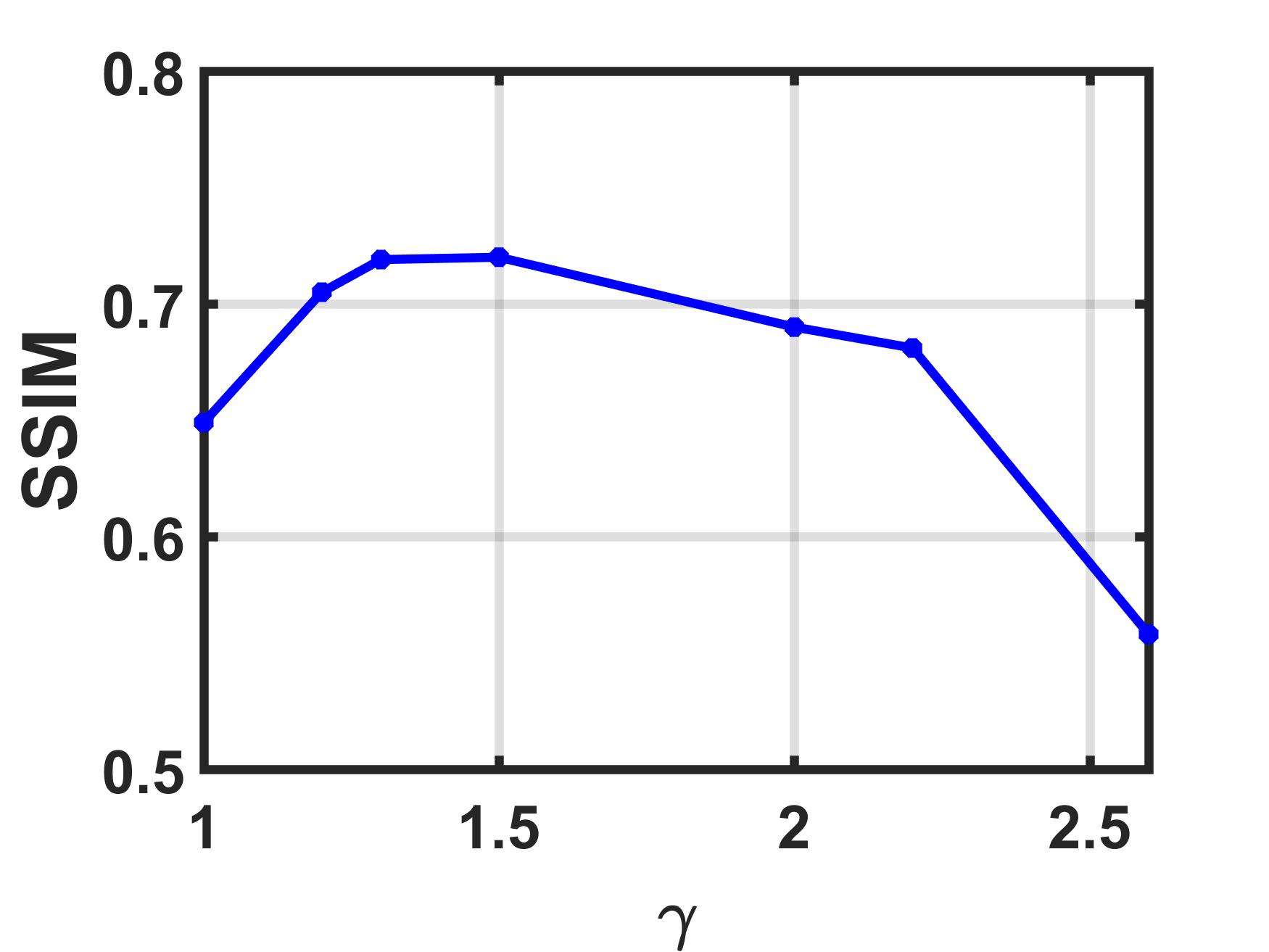}}%
	\hfil					
	\subfloat[]{\includegraphics[width=0.15\linewidth]{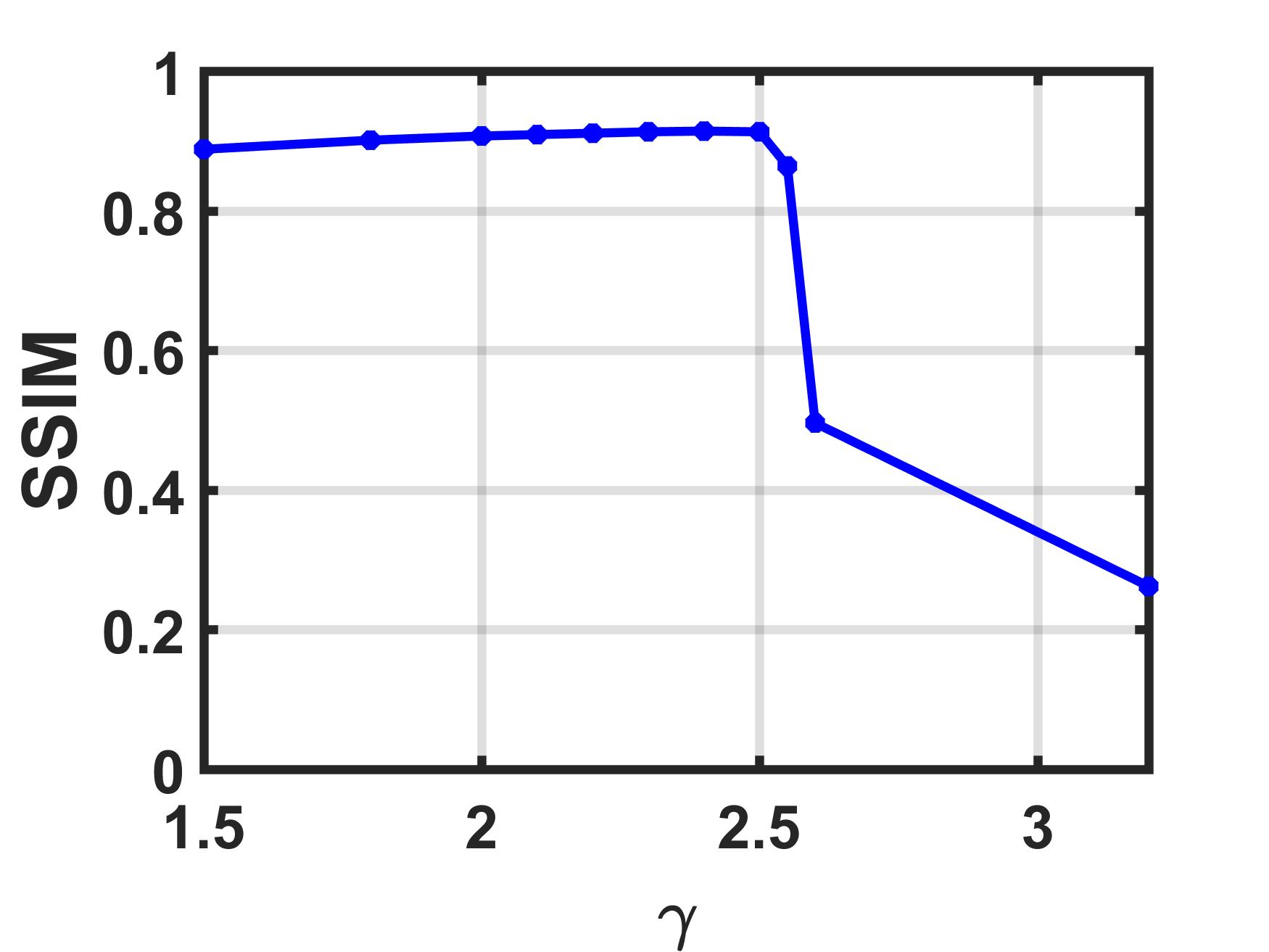}}%
	\hfil
	\subfloat[]{\includegraphics[width=0.15\linewidth]{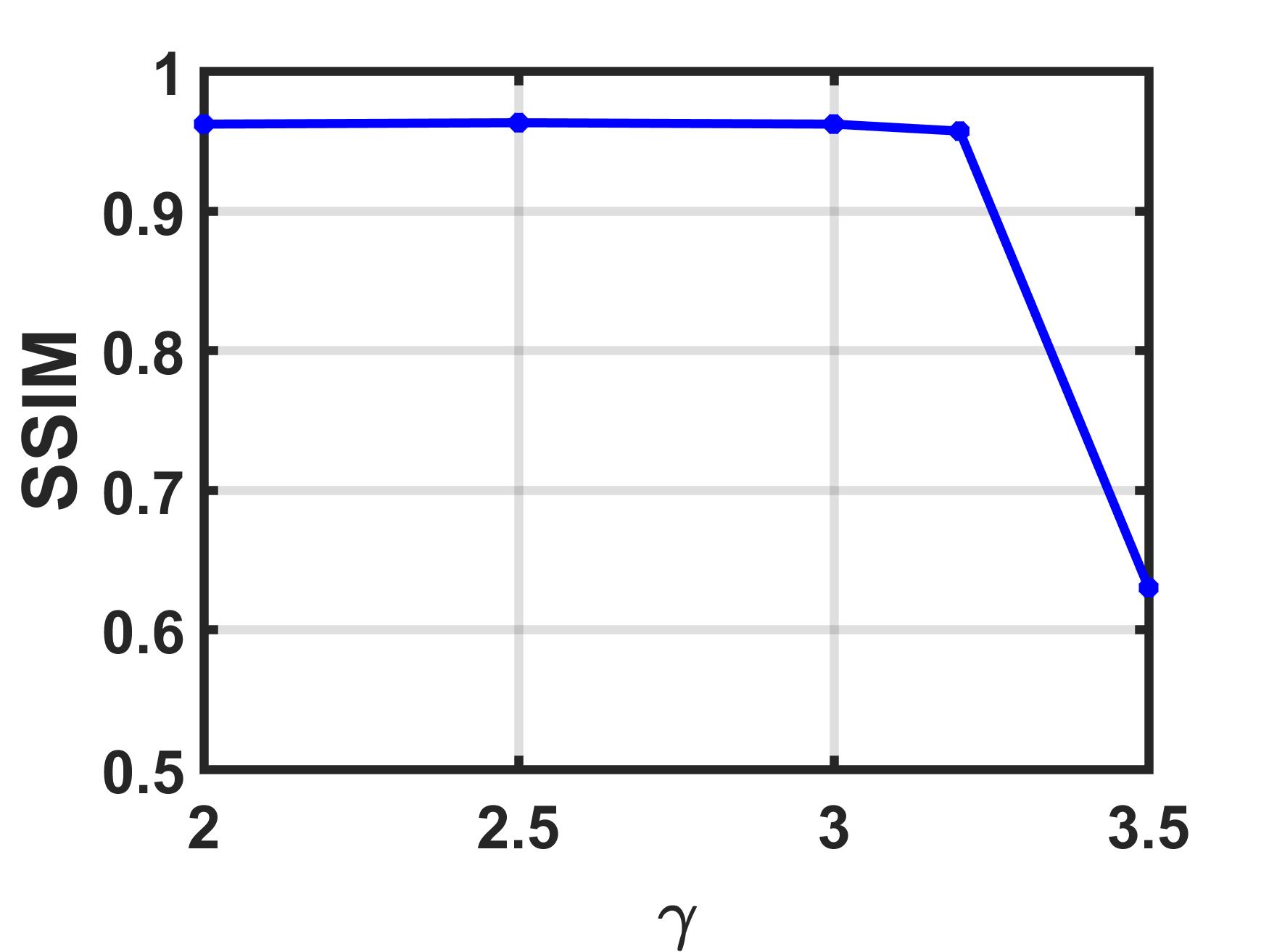}}%
	\caption{Sensitivity analysis of parameter $\gamma$ in MRI dataset.
		(a-c) Change in the MPSNR value: SR= 0.05, 0.1 and 0.2. (e-f) Change in the MSSIM value: 0.05, 0.1 and 0.2, respectively.}
	\label{parameter_analysis_MRI}
\end{figure*}

\begin{figure*}[!t]	
	\centering
	\subfloat[SR=0.025]{\includegraphics[width=0.15\linewidth]{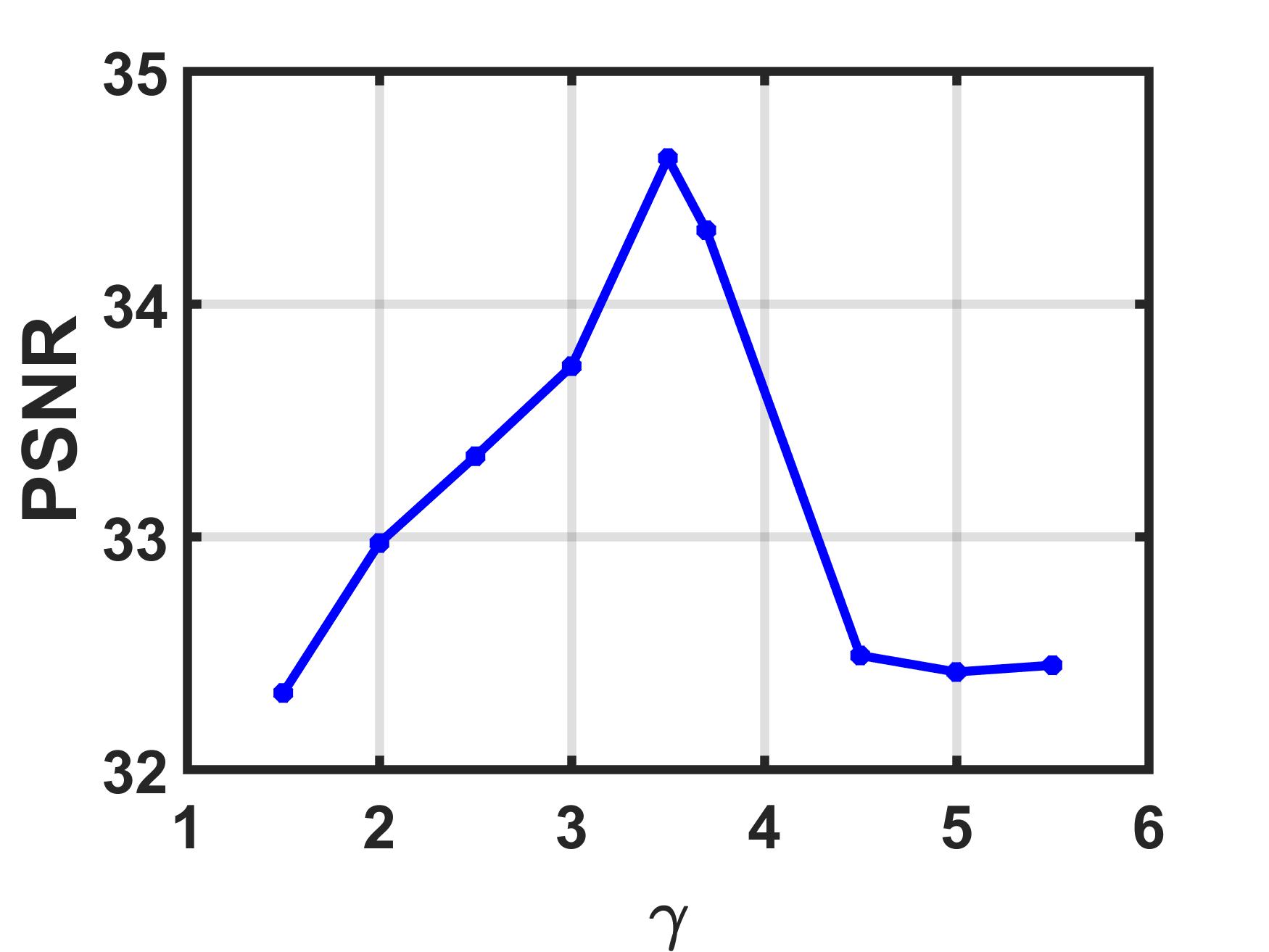}}%
	\hfil					
	\subfloat[SR=0.05]{\includegraphics[width=0.15\linewidth]{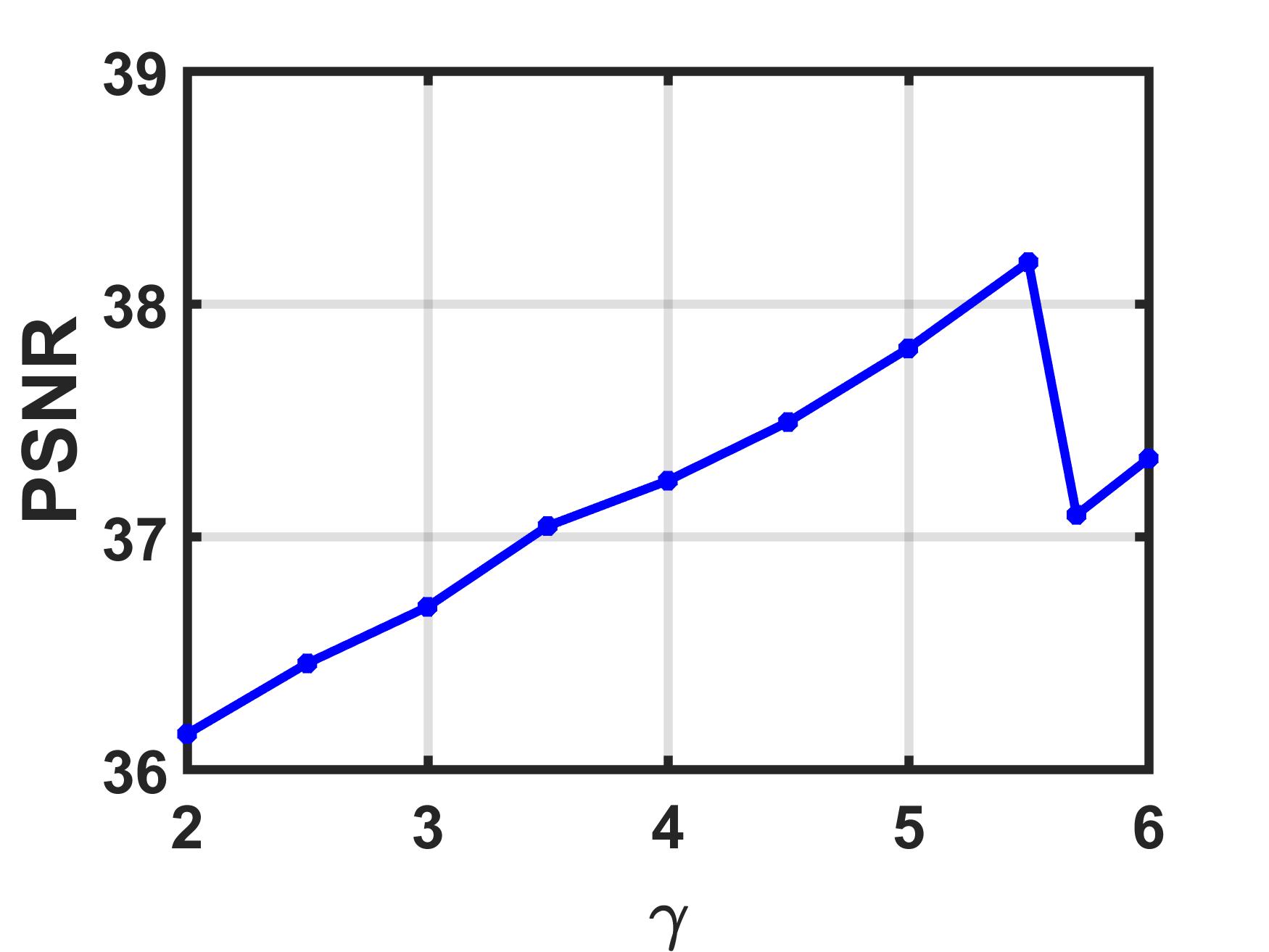}}%
	\hfil
	\subfloat[SR=0.1]{\includegraphics[width=0.15\linewidth]{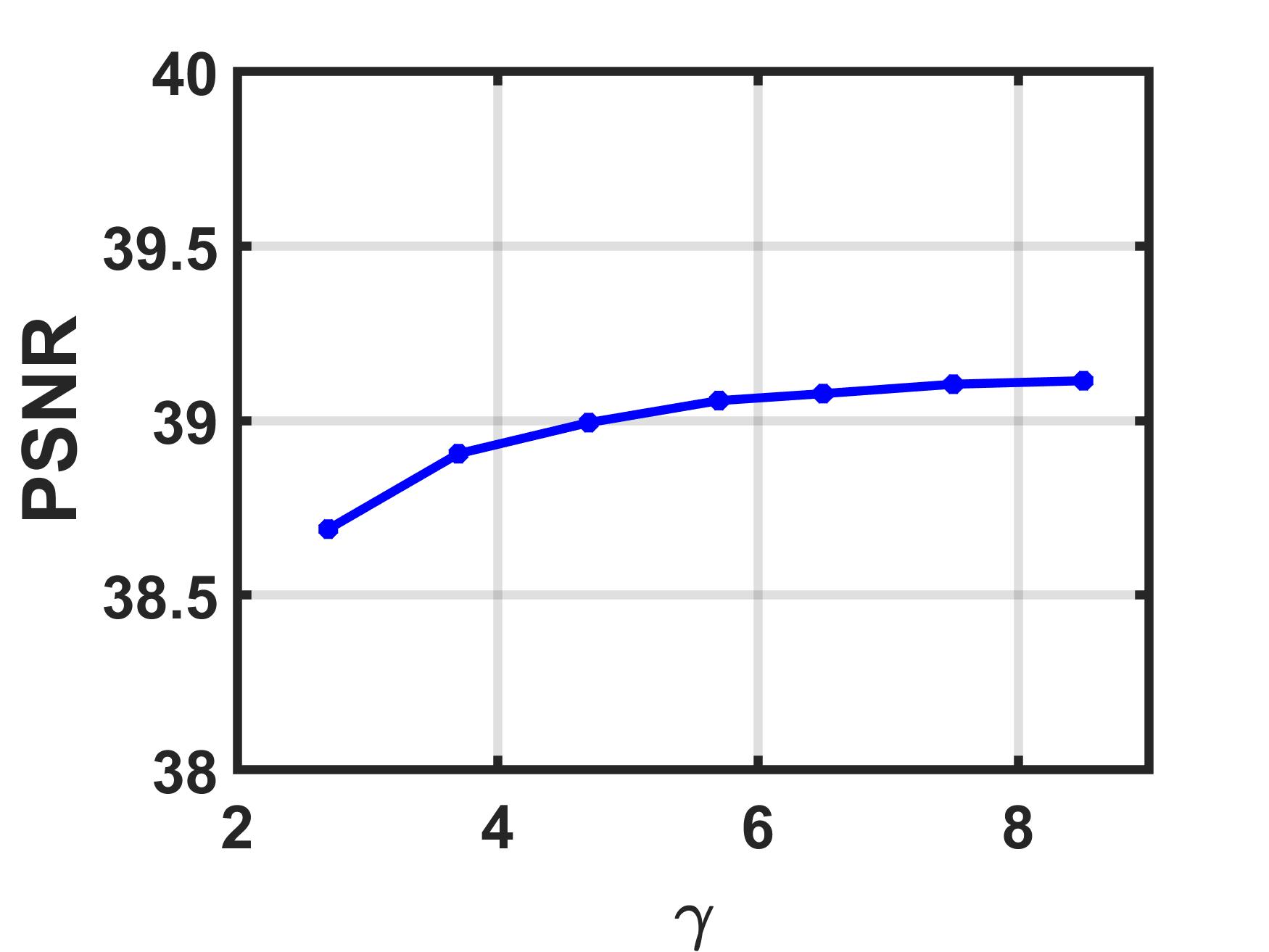}}%
	\hfil	
	\subfloat[SR=0.025]{\includegraphics[width=0.15\linewidth]{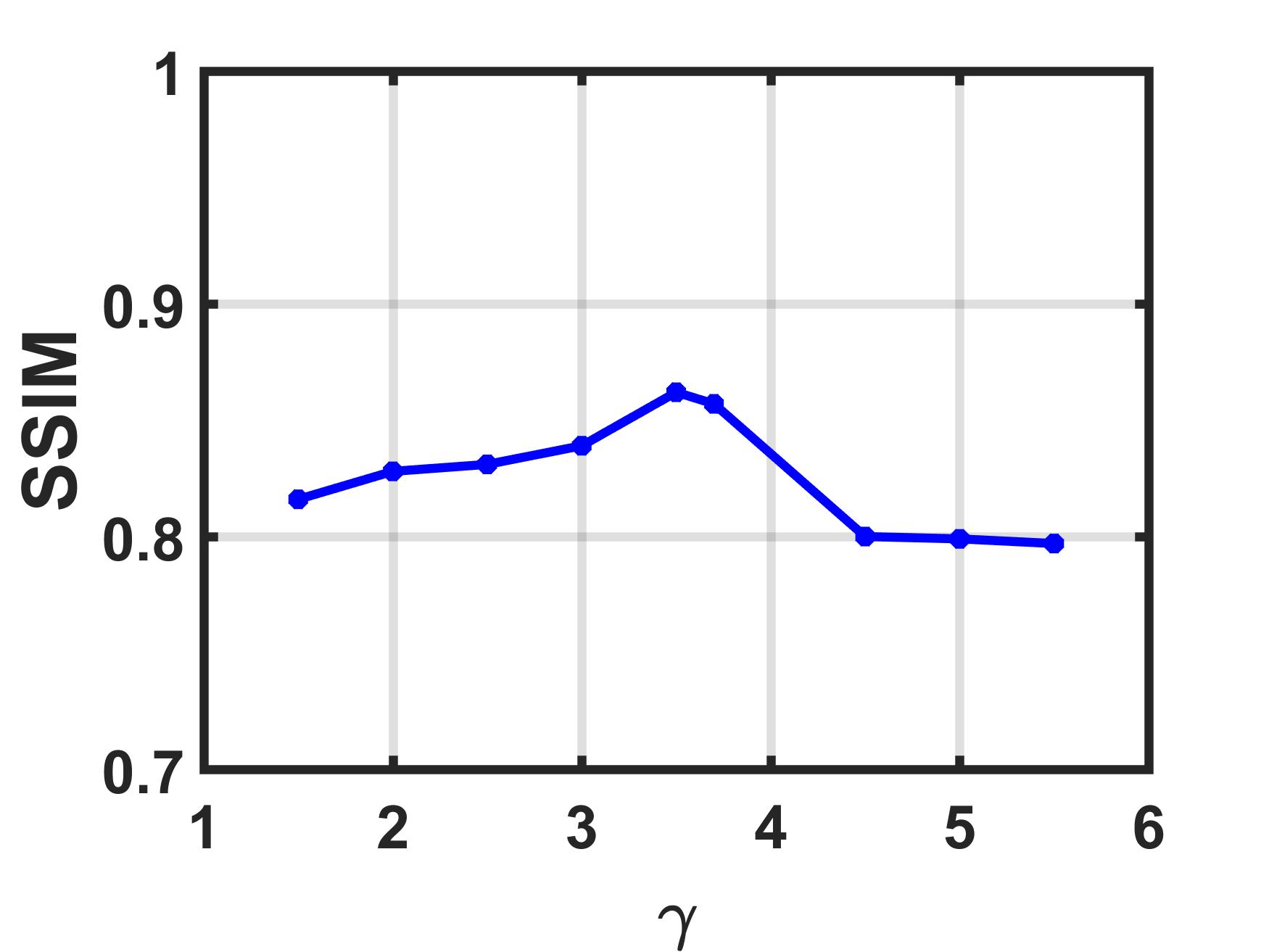}}%
	\hfil					
	\subfloat[SR=0.05]{\includegraphics[width=0.15\linewidth]{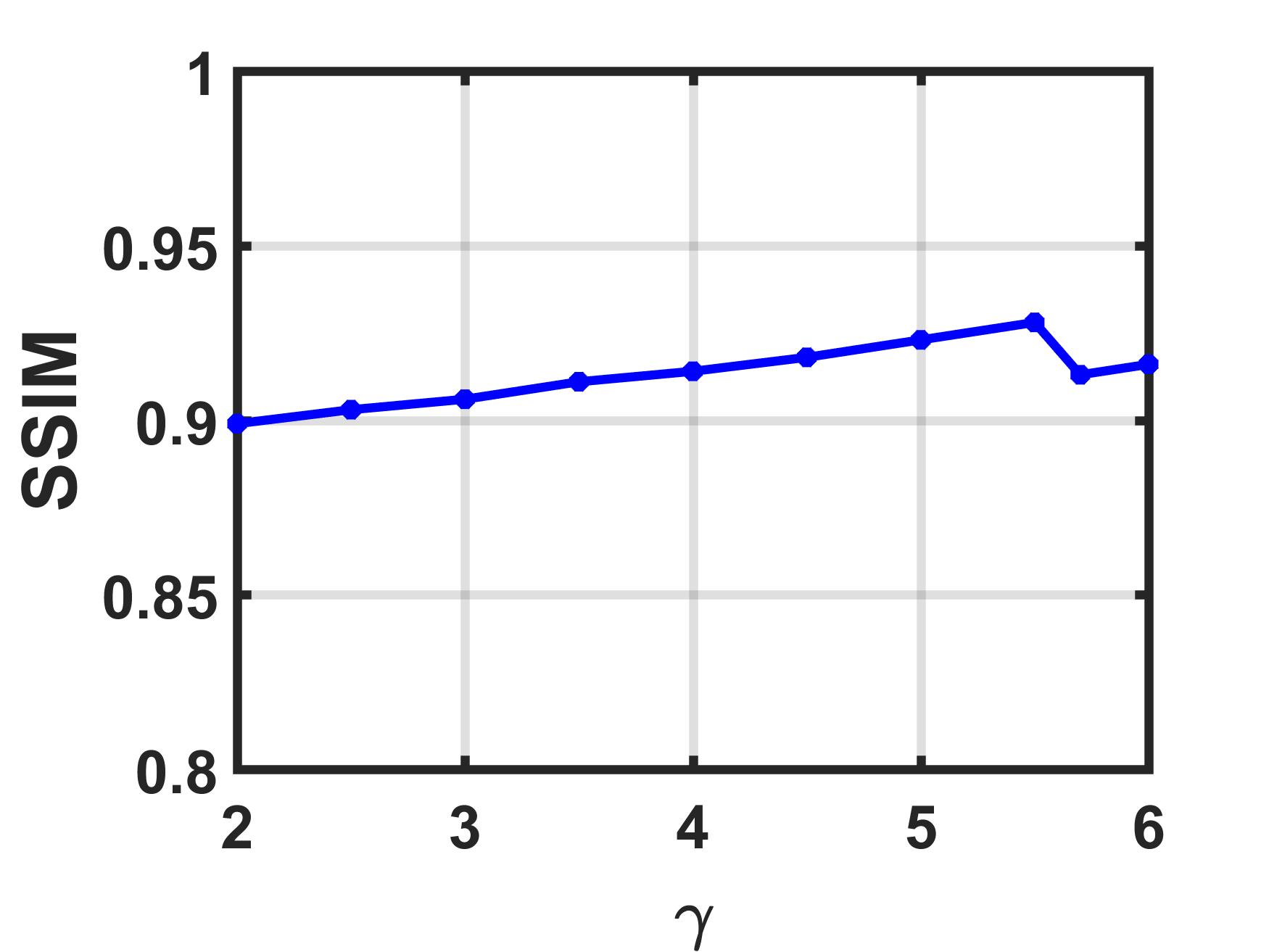}}%
	\hfil
	\subfloat[SR=0.1]{\includegraphics[width=0.15\linewidth]{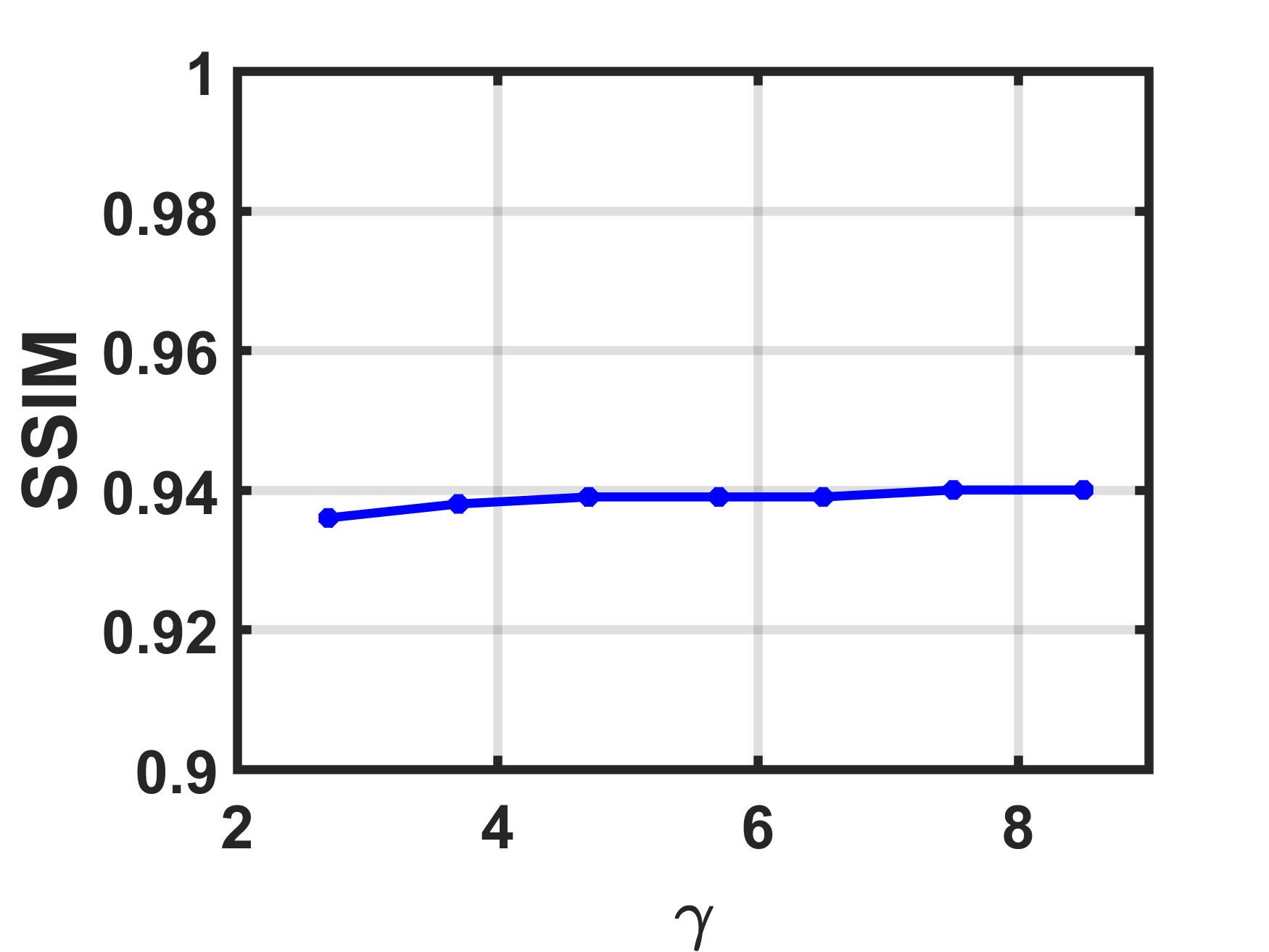}}%
	\caption{Sensitivity analysis of parameter $\gamma$ in HSI dataset.
		(a-c) Change in the MPSNR value: SR= 0.025, 0.05 and 01. (e-f) Change in the MSSIM value: 0.025, 0.05 and 0.1, respectively.}
	\label{parameter_analysis_HSI}
\end{figure*}

%\begin{table}
%	\centering
%	\caption{Parameters setting in the proposed algorithm}
%	\label{table:parameter_setting}
%	\begin{tabular}{ccc}
%		\hline
%		Dataset&SR&$\gamma$   \\
%		\hline
%		&0.05&2.3   \\
%		Video-suzie&0.1&2.5  \\
%		&0.2&2.7  \\
%		
%		
%		
%		&0.05&1.7 \\
%		Video-hall&0.1&2.6 \\
%		&0.2&2.6  \\
%		
%		
%		&0.05&1.3  \\
%		Cubical MRI dataset&0.1&2.5 \\
%		&0.2&3.0  \\
%
%		&0.025& 3.5  \\
%		AVIRIS Cuprite HSI dataset &0.05&5.5 \\
%		&0.1&5.7  \\
%		\hline 	
%	\end{tabular} 	
%\end{table}

\begin{table*}
	\centering
	\caption{Parameters setting in the proposed algorithm}
	\label{table:parameter_setting}
	\begin{tabular}{ccccc ccccc ccc}
		\hline
		&  &Vdieo-suzie&   &  &Video-hall&   &  &MRI dataset& & &HSI dataset& \\
		\hline
		SR&	0.05& 0.1&0.2      &0.05& 0.1&0.2      &0.05& 0.1&0.2    &0.025& 0.05&0.1\\
		$\gamma$&	2.3&2.5&2.7&1.7&2.6&2.6&1.3&2.5&3.0&3.5&5.5&5.7\\
		\hline 	
	\end{tabular} 	
\end{table*}

\subsection{Discussion}

Considering the fact that the $\gamma$ in $\mathbf{X}_n$ and $\mathbf{A}_n$ have certain proportional relationship,
we simply set the $\gamma$ in $\mathbf{X}_n$ as 0.1, and tune the $\gamma$ in $\mathbf{A}_n$ carefully.
For experiments on video datasets , MRI and HSI dataset,
Fig. \ref{parameter_analysis_hall}, Fig. \ref{parameter_analysis_MRI} and Fig. \ref{parameter_analysis_HSI} show the effect of parameter $\gamma$ on PSNR and SSIM at different sampling rates, respectively.
To enhance the repeatability of our model,
we list the optimal $\gamma$ of various datasets at different sampling rates in Table \ref{table:parameter_setting}.
In addition, we manually set both $\tau_n$ and $\lambda_n$ as 0.01.
%In addition for $\mu$, we first initialize it as $\mu=10^{-2}$ and then update it via $\mu=\min \left(\rho \mu, \mu_{\max }\right)$, which has been widely used in the ALM-based algorithms \cite{ALM_1,ALM_2}.

\section{Conclusions}
\label{conclusion}

In this paper, we propose a novel double low-rank tensor model based on multi-mode matrix decomposition for tensor completion.
Instead of using the traditional single nuclear norm or its relaxation to represent the low-rank prior of underlying tensor directly,
we first apply parallel matrix factorization to all modes of underlying tensor,
then, a novel double nonconvex $L_{\gamma}$ norm is designed to represent the underlying joint-manifold drawn from the mode factorization factors.
An BSUM based algorithm is designed to efficiently solve the proposed model,
and it can be demonstrated that our numerical scheme converges to the coordinatewise minimizers.
The proposed model has been evaluated on three types of public tensor datasets,
which show that our algorithm can complete a variety of low-rank tensors with significantly fewer samples than the compared methods.

\section*{Acknowledgment}
The authors would like to express their thanks to Dr. C. Lu, Dr. Y. Xu, Dr. T. Ji and Dr. T. Jiang for sharing their codes for the tested methods.
In addition, this research is supported by the Fundamental Research Funds for the Central Universities under Grant No. 2452019073 and the National
Natural Science Foundation of China under Grant No. 61876153.

%

%%%%%%%%%\appendices
%%%%%%%%%\section{Proof of the First Zonklar Equation}
%%%%%%%%%Appendix one text goes here.
%%%%%%%%%
%%%%%%%%%% you can choose not to have a title for an appendix
%%%%%%%%%% if you want by leaving the argument blank
%%%%%%%%%\section{}
%%%%%%%%%Appendix two text goes here.
%%%%%%%%%
%%%%%%%%%
%%%%%%%%%% use section* for acknowledgment
%%%%%%%%%\section*{Acknowledgment}
%%%%%%%%%
%%%%%%%%%
%%%%%%%%%The authors would like to thank...

% Can use something like this to put references on a page
% by themselves when using endfloat and the captionsoff option.
\ifCLASSOPTIONcaptionsoff
  \newpage
\fi

\bibliographystyle{IEEEtran}% our
\bibliography{IEEEabrv,mybibfile}% our

% Generated by IEEEtran.bst, version: 1.14 (2015/08/26)
\begin{thebibliography}{10}
\providecommand{\url}[1]{#1}
\csname url@samestyle\endcsname
\providecommand{\newblock}{\relax}
\providecommand{\bibinfo}[2]{#2}
\providecommand{\BIBentrySTDinterwordspacing}{\spaceskip=0pt\relax}
\providecommand{\BIBentryALTinterwordstretchfactor}{4}
\providecommand{\BIBentryALTinterwordspacing}{\spaceskip=\fontdimen2\font plus
\BIBentryALTinterwordstretchfactor\fontdimen3\font minus
  \fontdimen4\font\relax}
\providecommand{\BIBforeignlanguage}[2]{{%
\expandafter\ifx\csname l@#1\endcsname\relax
\typeout{** WARNING: IEEEtran.bst: No hyphenation pattern has been}%
\typeout{** loaded for the language `#1'. Using the pattern for}%
\typeout{** the default language instead.}%
\else
\language=\csname l@#1\endcsname
\fi
#2}}
\providecommand{\BIBdecl}{\relax}
\BIBdecl

\bibitem{tensor_video}
K.~A. Patwardhan, G.~Sapiro, and M.~Bertalm{\'\i}o, ``Video inpainting under
  constrained camera motion,'' \emph{IEEE Transactions on Image Processing},
  vol.~16, no.~2, pp. 545--553, 2007.

\bibitem{tensor_MRI}
J.~Yuan, ``Mri denoising via sparse tensors with reweighted regularization,''
  \emph{Applied Mathematical Modelling}, vol.~69, pp. 552--562, 2019.

\bibitem{MRI_tensor}
B.~Yaman, S.~Weing{\"a}rtner, N.~Kargas, N.~D. Sidiropoulos, and
  M.~Ak{\c{c}}akaya, ``Low-rank tensor models for improved multi-dimensional
  mri: Application to dynamic cardiac t1 mapping,'' \emph{IEEE Transactions on
  Computational Imaging}, 2019.

\bibitem{tensor_3dimage}
A.~C. Sauve, A.~O. Hero, W.~L. Rogers, S.~J. Wilderman, and N.~H. Clinthorne,
  ``3d image reconstruction for a compton spect camera model,'' \emph{IEEE
  Transactions on Nuclear Science}, vol.~46, no.~6, pp. 2075--2084, 1999.

\bibitem{tensor_3DReconstruction}
J.~{Tachella}, Y.~{Altmann}, M.~{Márquez}, H.~{Arguello-Fuentes},
  J.~{Tourneret}, and S.~{McLaughlin}, ``Bayesian 3d reconstruction of
  subsampled multispectral single-photon lidar signals,'' \emph{IEEE
  Transactions on Computational Imaging}, vol.~6, pp. 208--220, 2020.

\bibitem{tensor_HSI}
H.-J. Zeng, X.-Z. Xie, K.~Wen-Feng, S.~Cui, and J.-F. Ning, ``Hyperspectral
  image denoising via combined non-local self-similarity and local low-rank
  regularization,'' \emph{IEEE Access}, vol.~8, pp. 50\,190--50\,208, 2020.

\bibitem{chen2018tensor}
Y.~Chen, S.~Wang, and Y.~Zhou, ``Tensor nuclear norm-based low-rank
  approximation with total variation regularization,'' \emph{IEEE Journal of
  Selected Topics in Signal Processing}, vol.~12, no.~6, pp. 1364--1377, 2018.

\bibitem{tensor_web}
J.-T. Sun, H.-J. Zeng, H.~Liu, Y.~Lu, and Z.~Chen, ``Cubesvd: a novel approach
  to personalized web search,'' in \emph{Proceedings of the 14th international
  conference on World Wide Web}, 2005, pp. 382--390.

\bibitem{tensor_seismic_data}
N.~Kreimer and M.~D. Sacchi, ``A tensor higher-order singular value
  decomposition for prestack seismic data noise reduction and interpolation,''
  \emph{Geophysics}, vol.~77, no.~3, pp. V113--V122, 2012.

\bibitem{lu2019TRPCA}
C.~Lu, J.~Feng, Y.~Chen, W.~Liu, Z.~Lin, and S.~Yan, ``Tensor robust principal
  component analysis with a new tensor nuclear norm,'' \emph{IEEE Transactions
  on Pattern Analysis and Machine Intelligence}, vol.~42, no.~4, pp. 925--938,
  2019.

\bibitem{LMaFit}
Z.~Wen, W.~Yin, and Y.~Zhang, ``Solving a low-rank factorization model for
  matrix completion by a nonlinear successive over-relaxation algorithm,''
  \emph{Mathematical Programming Computation}, vol.~4, no.~4, pp. 333--361,
  2012.

\bibitem{CP}
R.~A. Harshman and M.~E. Lundy, ``Parafac: Parallel factor analysis,''
  \emph{Computational Statistics \& Data Analysis}, vol.~18, no.~1, pp. 39--72,
  1994.

\bibitem{tucker_1}
L.~R. Tucker, ``Some mathematical notes on three-mode factor analysis,''
  \emph{Psychometrika}, vol.~31, no.~3, pp. 279--311, 1966.

\bibitem{ZENG_HSI_tensor}
H.~Zeng, X.~Xie, H.~Cui, Y.~Zhao, and J.~Ning, ``Hyperspectral image
  restoration via cnn denoiser prior regularized low-rank tensor recovery,''
  \emph{Computer Vision and Image Understanding}, vol. 197-198, p. 103004,
  2020.

\bibitem{t_SVD}
M.~E. Kilmer, K.~Braman, N.~Hao, and R.~C. Hoover, ``Third-order tensors as
  operators on matrices: A theoretical and computational framework with
  applications in imaging,'' \emph{SIAM Journal on Matrix Analysis and
  Applications}, vol.~34, no.~1, pp. 148--172, 2013.

\bibitem{tensor_train_TingzhuHuang}
M.~Ding, T.-Z. Huang, T.-Y. Ji, X.-L. Zhao, and J.-H. Yang, ``Low-rank tensor
  completion using matrix factorization based on tensor train rank and total
  variation,'' \emph{Journal of Scientific Computing}, vol.~81, no.~2, pp.
  941--964, 2019.

\bibitem{tensor_ring_YipengLiu}
H.~Huang, Y.~Liu, and C.~Zhu, ``Low-rank tensor completion via tensor ring with
  balanced unfolding,'' \emph{arXiv preprint arXiv:1903.03315}, 2019.

\bibitem{HaLRTC}
J.~Liu, P.~Musialski, P.~Wonka, and J.~Ye, ``Tensor completion for estimating
  missing values in visual data,'' \emph{IEEE Transactions on Pattern Analysis
  and Machine Intelligence}, vol.~35, no.~1, pp. 208--220, 2012.

\bibitem{tensor_Qrank}
H.~Kong and Z.~Lin, ``Tensor q-rank: A new data dependent tensor rank,''
  \emph{arXiv preprint arXiv:1910.12016}, 2019.

\bibitem{list_nonconvex}
C.~Lu, J.~Tang, S.~Yan, and Z.~Lin, ``Generalized nonconvex nonsmooth low-rank
  minimization,'' in \emph{Proceedings of the IEEE Conference on Computer
  Vision and Pattern Recognition}, 2014, pp. 4130--4137.

\bibitem{Original_BFMN}
F.~Shang, J.~Cheng, Y.~Liu, Z.-Q. Luo, and Z.~Lin, ``Bilinear factor matrix
  norm minimization for robust pca: Algorithms and applications,'' \emph{IEEE
  Transactions on Pattern Analysis and Machine Intelligence}, vol.~40, no.~9,
  pp. 2066--2080, 2017.

\bibitem{PSTNN}
T.-X. Jiang, T.-Z. Huang, X.-L. Zhao, and L.-J. Deng, ``Multi-dimensional
  imaging data recovery via minimizing the partial sum of tubal nuclear norm,''
  \emph{Journal of Computational and Applied Mathematics}, vol. 372, p. 112680,
  2020.

\bibitem{xue2019nonconvex}
J.~Xue, Y.~Zhao, W.~Liao, and J.~C.-W. Chan, ``Nonconvex tensor rank
  minimization and its applications to tensor recovery,'' \emph{Information
  Sciences}, vol. 503, pp. 109--128, 2019.

\bibitem{Folded_concave}
W.~Cao, Y.~Wang, C.~Yang, X.~Chang, Z.~Han, and Z.~Xu, ``Folded-concave
  penalization approaches to tensor completion,'' \emph{Neurocomputing}, vol.
  152, pp. 261--273, 2015.

\bibitem{logsum_penalty}
E.~J. Candes, M.~B. Wakin, and S.~P. Boyd, ``Enhancing sparsity by reweighted
  l1 minimization,'' \emph{Journal of Fourier Analysis and Applications},
  vol.~14, no. 5-6, pp. 877--905, 2008.

\bibitem{ji_nonlogDet}
T.-Y. Ji, T.-Z. Huang, X.-L. Zhao, T.-H. Ma, and L.-J. Deng, ``A non-convex
  tensor rank approximation for tensor completion,'' \emph{Applied Mathematical
  Modelling}, vol.~48, pp. 410--422, 2017.

\bibitem{lysaker_noise_2004_10_27}
M.~Lysaker, S.~Osher, and X.-C. Tai, ``Noise removal using smoothed normals and
  surface fitting,'' \emph{IEEE Transactions on Image Processing}, vol.~13,
  no.~10, p. 1345, 2004.

\bibitem{burger_nonlinear_2006_10_28}
M.~Burger, G.~Gilboa, S.~Osher, and J.~Xu, ``Nonlinear inverse scale space
  methods,'' \emph{Communications in Mathematical Sciences}, vol.~4, no.~1, pp.
  179--212, 2006.

\bibitem{lou_L1L2}
Y.~Lou, P.~Yin, Q.~He, and J.~Xin, ``Computing sparse representation in a
  highly coherent dictionary based on difference of $ l_1 $ and $ l_2 $,''
  \emph{Journal of Scientific Computing}, vol.~64, no.~1, pp. 178--196, 2015.

\bibitem{Non_LRMA}
Y.~Chen, Y.~Guo, Y.~Wang, D.~Wang, C.~Peng, and G.~He, ``Denoising of
  hyperspectral images using nonconvex low rank matrix approximation,''
  \emph{IEEE Transactions on Geoscience and Remote Sensing}, vol.~55, no.~9,
  pp. 5366--5380, 2017.

\bibitem{Tmac}
Y.~Xu, R.~Hao, W.~Yin, and Z.~Su, ``Parallel matrix factorization for low-rank
  tensor completion,'' \emph{arXiv preprint arXiv:1312.1254}, 2013.

\bibitem{MFTV}
T.-Y. Ji, T.-Z. Huang, X.-L. Zhao, T.-H. Ma, and G.~Liu, ``Tensor completion
  using total variation and low-rank matrix factorization,'' \emph{Information
  Sciences}, vol. 326, pp. 243--257, 2016.

\bibitem{TNN}
Z.~Zhang and S.~Aeron, ``Exact tensor completion using t-svd,'' \emph{IEEE
  Transactions on Signal Processing}, vol.~65, no.~6, pp. 1511--1526, 2016.

\bibitem{T_Sp}
H.~Zhang, J.~Yang, F.~Shang, C.~Gong, and Z.~Zhang, ``Lrr for subspace
  segmentation via tractable schatten-$ p $ norm minimization and
  factorization,'' \emph{IEEE Transactions on Cybernetics}, vol.~49, no.~5, pp.
  1722--1734, 2018.

\bibitem{HSI_unmixing}
J.~M. Bioucas-Dias, A.~Plaza, N.~Dobigeon, M.~Parente, Q.~Du, P.~Gader, and
  J.~Chanussot, ``Hyperspectral unmixing overview: Geometrical, statistical,
  and sparse regression-based approaches,'' \emph{IEEE Journal of Selected
  Topics in Applied Earth Observations and Remote Sensing}, vol.~5, no.~2, pp.
  354--379, 2012.

\bibitem{BSUM}
M.~Razaviyayn, M.~Hong, and Z.-Q. Luo, ``A unified convergence analysis of
  block successive minimization methods for nonsmooth optimization,''
  \emph{SIAM Journal on Optimization}, vol.~23, no.~2, pp. 1126--1153, 2013.

\bibitem{lu_nonconvex}
C.~Lu, J.~Tang, S.~Yan, and Z.~Lin, ``Nonconvex nonsmooth low rank minimization
  via iteratively reweighted nuclear norm,'' \emph{IEEE Transactions on Image
  Processing}, vol.~25, no.~2, pp. 829--839, 2015.

\bibitem{WSVT}
S.~Ga{\"\i}ffas and G.~Lecu{\'e}, ``Weighted algorithms for compressed sensing
  and matrix completion,'' \emph{arXiv preprint arXiv:1107.1638}, 2011.

\bibitem{PSNR}
Q.~Huynh-Thu and M.~Ghanbari, ``Scope of validity of psnr in image/video
  quality assessment,'' \emph{Electronics Letters}, vol.~44, no.~13, pp.
  800--801, 2008.

\bibitem{SSIM}
Z.~Wang, A.~C. Bovik, H.~R. Sheikh, E.~P. Simoncelli \emph{et~al.}, ``Image
  quality assessment: from error visibility to structural similarity,''
  \emph{IEEE Transactions on Image Processing}, vol.~13, no.~4, pp. 600--612,
  2004.

\bibitem{FSIM}
L.~Zhang, L.~Zhang, X.~Mou, and D.~Zhang, ``{FSIM}: {A} feature similarity
  index for image quality assessment,'' \emph{IEEE Transactions on Image
  Processing}, vol.~20, no.~8, pp. 2378--2386, 2011.

\bibitem{EGRAS}
L.~Wald, \emph{Data fusion: definitions and architectures: fusion of images of
  different spatial resolutions}.\hskip 1em plus 0.5em minus 0.4em\relax
  Presses des MINES, 2002.

\bibitem{SAM}
F.~Kruse, A.~Lefkoff, and J.~Dietz, ``Expert system-based mineral mapping in
  northern death valley, california/nevada, using the airborne visible/infrared
  imaging spectrometer (aviris),'' \emph{Remote Sensing of Environment},
  vol.~44, no. 2-3, pp. 309--336, 1993.

\end{thebibliography}

\end{document}